%% file: 00_CERBERUS_FR_SubT_PhaseI_II.tex
\documentclass{article}
\usepackage{apalike}
\usepackage{graphicx}
\usepackage{jfrExamplee}
\usepackage{setspace}
\usepackage[binary-units]{siunitx}
\usepackage{subcaption}
\usepackage{url}
\usepackage{booktabs}
\usepackage{multirow}

\usepackage{comment}
\usepackage{graphics} %
\usepackage{epsfig}
\usepackage{amsmath} %
\usepackage{amssymb}  %

\usepackage{algorithm}
\usepackage{algpseudocode}
\algnewcommand\AAND{\textbf{ and }}
\algnewcommand\Or{\textbf{ or }}
\usepackage{color}
\usepackage[table,xcdraw]{xcolor}

\input{acronyms}

\input{math_notations}

\definecolor{revision}{RGB}{0, 0, 0}
\newcommand{\revisedtext}[1]{{\color{revision} #1}}

\DeclareSIUnit[per-mode=symbol,per-symbol=p]{\MBps}{\mega\byte\per\second}
\DeclareSIUnit[per-mode=symbol,per-symbol=p]{\MB}{\mega\byte}

\title{CERBERUS: Autonomous Legged and Aerial Robotic Exploration in the Tunnel and Urban Circuits of the DARPA Subterranean Challenge}

\author{
Marco Tranzatto$^{1}$
\thanks{$^{1}$ Robotic Systems Lab, ETH Zurich, $^{2}$ Autonomous Robots Lab, University of Nevada, Reno | Norwegian University of Science and Technology, $^{3}$ Autonomous Systems Lab, ETH Zurich, $^{4}$ Flyability, ${5}$ Oxford Robotics Institute, University of Oxford $^{6}$ University of California, Berkeley, $^{7}$ Sierra Nevada Corporation, Direct correspondence to Marco Tranzatto \texttt{marcot@ethz.ch} }
\And

Frank Mascarich$^{2}$\And
Lukas Bernreiter$^{3}$\And
Carolina Godinho$^{4}$\And
Marco Camurri$^{5}$\And
Shehryar Khattak$^{1}$\And
Tung Dang$^{2}$\And
Victor Reijgwart$^{3}$\And
Johannes L\"{o}je$^{4}$\And
David Wisth$^{5}$\And
Samuel Zimmermann$^{1}$\And
Huan Nguyen$^{2}$\And
Marius Fehr$^{3}$\And
Lukas Solanka$^{4}$\And
Russell Buchanan$^{5}$\And
Marko Bjelonic$^{1}$\And
Nikhil Khedekar$^{2}$\And
Mathieu Valceschini$^{4}$\And
Fabian Jenelten$^{1}$\And
Mihir Dharmadhikari$^{2}$\And
Timon Homberger$^{1}$\And
Paolo De Petris$^{2}$\And
Lorenz Wellhausen$^{1}$\And
Mihir Kulkarni$^{2}$\And
Takahiro Miki$^{1}$\And
Satchel Hirsch$^{2}$\And
Markus Montenegro$^{1}$\And
Christos Papachristos$^{2}$\And
Fabian Tresoldi$^{1}$\And
Jan Carius$^{1}$\And
Giorgio Valsecchi$^{1}$\And
Joonho Lee$^{1}$\And
Konrad Meyer$^{1}$\And
Xiangyu Wu$^{6}$\And
Juan Nieto$^{3}$\And
Andy Smith$^{7}$\And
Marco Hutter$^{1}$\And
Roland Siegwart$^{3}$\And
Mark Mueller$^{6}$\And
Maurice Fallon$^{5}$\And
Kostas Alexis$^{2}$
 \\
}

\begin{document}

\maketitle

\begin{abstract}
Autonomous exploration of subterranean environments constitutes a major frontier for robotic systems as underground settings present key challenges that can render robot autonomy hard to achieve. This has motivated the DARPA Subterranean Challenge, where teams of robots search \revisedtext{for} objects of interest in various underground environments. In response, the CERBERUS system-of-systems is presented as a unified strategy towards subterranean exploration using legged and flying robots.
As primary robots, ANYmal quadruped systems are deployed considering their endurance and potential to traverse challenging \revisedtext{terrain}.
For aerial robots, both conventional and collision-tolerant multirotors are utilized to explore \revisedtext{spaces too narrow or otherwise unreachable} by ground systems.
Anticipating degraded sensing conditions, a complementary multi-modal sensor fusion approach utilizing camera, LiDAR, and inertial data for resilient robot pose estimation is proposed. Individual robot pose estimates are refined by a centralized multi-robot map optimization approach to improve the reported location accuracy of detected objects of interest in the DARPA-defined coordinate frame.
Furthermore, a unified exploration path planning policy is presented to facilitate the autonomous operation of both legged and aerial robots in complex underground networks.
Finally, to enable communication between the robots and the base station, CERBERUS utilizes a ground rover with a high-gain antenna and an optical fiber connection to the base station, alongside breadcrumbing of wireless nodes by our legged robots.
We report results from the CERBERUS system-of-systems deployment at the DARPA Subterranean Challenge Tunnel and Urban Circuits, along with the current limitations and the lessons learned for the benefit of the community.

\end{abstract}

\section{Introduction}\label{sec:introduction}
\input{01_introduction}

\section{Related Work}\label{sec:related_work}
\input{02_related_work}

\section{CERBERUS System-of-Systems}\label{sec:system_of_systems}
\input{03_system_of_systems}

\section{CERBERUS Robots} \label{sec:cerberus_robots}

This section outlines the CERBERUS robots, their system design and specifics with respect to subterranean exploration functionality.

\subsection{Subterranean Walking Robots - ANYmal B SubT} \label{sec:anymal_b_subt}
\input{04_subterranean_walking_robots}

\subsection{Subterranean Aerial Robots}\label{sec:aerial_robots}
\input{05_subterranean_aerial_robots}

\subsection{Roving Robots} %
\input{06_roving_robots} %

\section{Exploration Autonomy} \label{sec:path_planner}
\input{07_expl_path_planning}

\section{Multi-Modal Perception for Multi-Robot Mapping} \label{sec:multi_model_perception}
\input{08_distributed_multi_modal_perception}

\section{Artifacts}\label{sec:artifacts}
\input{09_artifacts}

\section{Networking}\label{sec:networking}
\input{10_networking}

\section{Operator Interface}\label{sec:operator}
\input{11_operator_interface}

\section{Experimental Evaluation}\label{sec:experimental_studies}

\input{12_experimental_evaluation}

\section{Main Lessons Learned and Discussion}\label{sec:lessons_learned}
\input{13_lessons_learnt_and_discussion}

\section{Priorities for Future Work}\label{sec:future_work}
\input{14_future_work}

\section{Conclusion}\label{sec:concl}
This paper presents a comprehensive report of the technological progress made by team ``CERBERUS'' participating in the DARPA Subterranean Challenge up to the Urban Circuit event. A detailed explanation of the CERBERUS vision and motivation for the team's overall ``System-of-Systems'' methodology is given, followed by an extensive presentation of the employed robotic systems. \textcolor{revision}{At the core of the CERBERUS robotic system is a team of legged and flying robots (including collision-tolerant designs), further enhanced with a roving platform.} The implemented autonomy components, including multi-modal perception and path planning\textcolor{revision}{,} as well as the automated artifact detection and scoring systems, are detailed alongside the robot-deployable communications solution and the operator interfaces for exerting high-level control. \textcolor{revision}{The contributed research allowed to facilitate certain levels of resilience in the CERBERUS robots operating inside visually-degraded, large-scale, narrow and broadly complex underground settings.} Finally, the work presents the team's performance in both the Tunnel and Urban Circuits of the challenge, followed by the critical lessons learned from these experiences and a brief overview of future work to be performed leading up to the final Subterranean Challenge \textcolor{revision}{event}. \textcolor{revision}{We hope that the presented work, released open-source code, and future activities can contribute towards the overall community goal of enabling resilient robotic autonomy in extreme underground environments.}

\subsubsection*{Acknowledgments}
This material is based upon work supported by the Defense Advanced Research Projects Agency (DARPA) under Agreement No. HR00111820045. The presented content and ideas are solely those of the authors.

The authors would also like to thank ANYmal Bear, ANYmal Badger, SMB Armadillo, Aerial Scouts Alpha, Bravo, Charlie, and Gagarins who were not harmed during the DARPA SubT Challenge Events, even though the same might not be said for the prior field deployments. In any case, they made it home safely after each circuit and were able to keep up their hard working attitude and support to the robotics communities of their research labs. Even more importantly, we want to thank and express our appreciation for all members of the other teams participating in the SubT Challenge, the organization committee and the overall community involved in this exciting competition.

\appendix
\section{List of Open Source Packages}\label{sec:appendixA}
\input{15_open_source_list.tex}

\section{Videos from the Tunnel and Urban Circuits}\label{sec:appendixB}
\input{16_videos.tex}

\bibliographystyle{apalike}

\end{document}

%% file: acronyms.tex
\usepackage[printonlyused]{acronym}

\acrodef{ARL}{Autonomous Robots Lab}
\acrodef{ARI}{Artifacts Reporting Interface}
\acrodef{ASL}{Autonomous Systems Lab}
\acrodef{GUI}{Graphical User Interface}
\acrodef{MCI}{Mission Control Interface}
\acrodef{NTP}{Network Time Protocol}
\acrodef{OPC}{Operator's PC}
\acrodef{ROS}{Robot Operating System}
\acrodef{RSSI}{Received Signal Strength Indication}
\acrodef{RSL}{Robotic Systems Lab}
\acrodef{STIX}{SubT Integration Exercise}
\acrodef{ICP}{Iterative Closest Point}
\acrodef{UGV}{Unmanned Ground Vehicle}
\acrodef{CNN}{Convolutional Neural Network}

%% file: math_notations.tex
\DeclareMathAlphabet{\pazocal}{OMS}{zplm}{m}{n}

%% file: 01_introduction.tex
This paper reports the technological progress made by team ``CERBERUS'' participating in the DARPA Subterranean Challenge (SubT Challenge). The SubT Challenge \textcolor{revision}{is} an international robotics competition organized and coordinated by the Defense Advanced Research Projects Agency (DARPA) and calls on teams to compete with respect to exploring, mapping and searching large-scale underground environments such as tunnels or mines, urban subterranean infrastructure and cave networks. Responding to the requirements of this challenge, this paper outlines the vision and technological developments of team CERBERUS, short for ``CollaborativE walking \& flying RoBots for autonomous ExploRation in Underground Settings''. Team CERBERUS corresponds to a combined system of legged and aerial robots equipped with multi-modal perception capabilities, localization and mapping autonomy, as well as robot-deployable communications that enables resilient navigation, exploration, mapping and object search inside complex, large-scale, sensor-degraded and communications-denied subterranean environments. CERBERUS is developed based on an international partnership, which up to the time of writing \textcolor{revision}{involves} the Autonomous Robots Lab (University of Nevada, Reno), Robotic Systems Lab (ETH Zurich), Autonomous Systems Lab (ETH Zurich), HiPeR Lab (University of California, Berkeley), the Dynamic Robot Systems Group (University of Oxford), Flyability and Sierra Nevada Corporation. This paper presents both the technological elements of CERBERUS and detailed results of our team's performance in the Tunnel and Urban Circuits, which were held in an underground mine and a never-commissioned nuclear facility, respectively.

\begin{figure}[h!]
\centering
    \includegraphics[width=0.99\columnwidth]{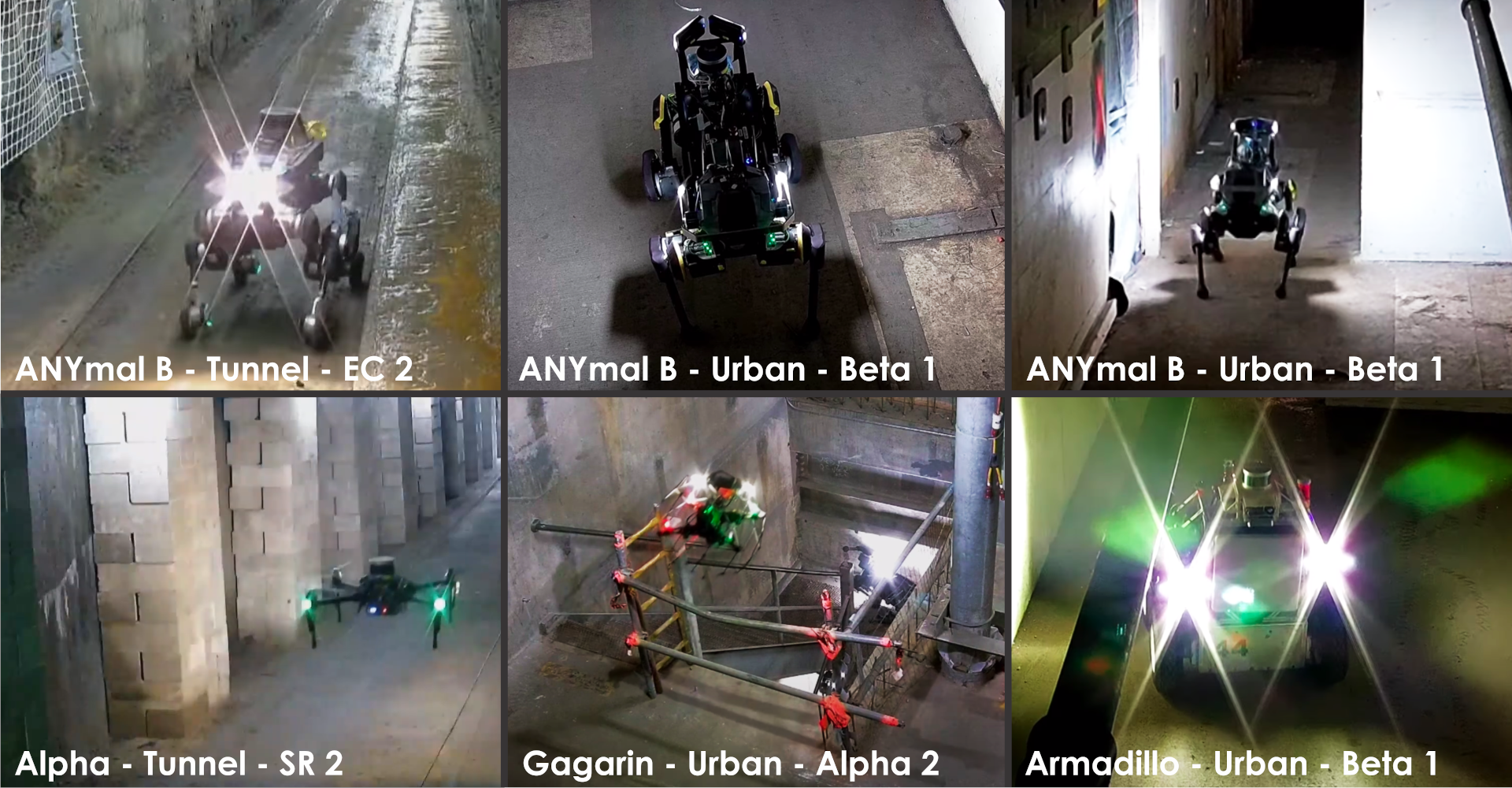}
\caption{Instances of deployments of the CERBERUS robotic system-of-systems in the DARPA Subterranean Challenge Tunnel and Urban Circuits. Upper row: ANYmal B on wheels and two standard ANYmal B robots exploring sections of the Tunnel Circuit's ``Experimental'' Course and Urban Circuit's ``Alpha'' and ``Beta'' Courses. Lower row: the Alpha and Gagarin Aerial Scout\textcolor{revision}{s} in the Tunnel Circuit's ``Safety Research'' Course and the Urban Circuit's ``Alpha'' Course respectively, alongside the Armadillo rover in the Urban Circuit's ``Beta'' Course. Relevant video results are available at~\url{http://bit.ly/cerberus2020} }\label{fig:cerberus_intro}
\end{figure}

During the Tunnel and Urban Circuits, CERBERUS' core robots were the ``ANYbotics ANYmal B'' quadruped, and two classes of ``Aerial Scouts'', namely multirotors with conventional airframes, as well as the ``Gagarin'' class of collision-tolerant aerial robots. We have further added a roving robot called ``Armadillo''. ANYmal has the most extended endurance compared to all our robots and integrates our multi-modal perception and autonomy solutions. At the same time, ANYmal can also deploy wireless communication nodes. The Aerial Scout, with conventional airframe, is a traditional quadrotor design and acts as a rapid explorer. The Gagarin collision-tolerant Aerial Scout is of smaller size and its mechanical structure enables it to survive contact with surfaces and to make its way through challenging, narrow, multi-branched, and possibly multi-level underground environments. Both aerial platforms integrate the CERBERUS multi-modal perception and autonomy pipelines and operate without any human supervision after take-off. Lastly, the Armadillo roving platform primarily deploys a high-gain wireless communications antenna deeper in the underground environment, while maintaining an optical fiber connection to the Base Station. As a fallback role, this roving robot also integrates multi-modal perception capabilities to allow it to detect artifacts.
Figure~\ref{fig:cerberus_intro} presents the CERBERUS robots deployed at the Tunnel and Urban Circuits of the SubT Challenge. 

To enable autonomous exploration of subterranean environments, team CERBERUS focuses on multi-modal and multi-robot perception, as well as path planning for exploration and area search. A unified complementary onboard perception solution is developed that fuses LiDAR, cameras, and inertial sensor cues. This enables resilient localization and situational awareness even in conditions of perceptual degradation such as geometric self-similarity, darkness and the presence of obscurants. The individual multi-modal perception capabilities are further exploited at the level of multi-robot collaboration with map sharing and alignment across the robotic systems of CERBERUS using centralized map optimization at the Base Station. In brief, each robot can detect objects of interest in the field, and with map optimization, refine their position estimates back at the Base Station. 
Building on top of this core capacity, CERBERUS uses a unified exploration path planning solution on both legged and aerial robotic systems. The method relies on a graph-based search and an information gain formulation related to volumetric exploration, avoids collisions, and accounts for traversability limitations. While it enables efficient local search for optimized paths, it also enables the ability for automated homing and to travel to frontiers of the exploration space detected at earlier steps of the robot's mission. It is thus tailored to the large-scale, multi-level, and often narrow character of underground environments.

As CERBERUS develops autonomous exploration and situational awareness through a team of robotic systems overseen by a Human Supervisor located outside of the underground environment, the final major development priority relates to a robot-deployable communications solution. Maintaining a connection with sufficient bandwidth as the robots proceed deeper into the subterranean environment is challenging due to the size of the environments, along with their geometric and material properties. In response to this challenge, CERBERUS utilizes customized $5.8\textrm{GHz}$ WiFi modules with each robot integrating its own antennas, the ANYmal robots ferrying and being able to deploy communication breadcrumbs, and the Armadillo rover integrating a high-gain antenna connected to the Base Station. This network allows for bidirectional communication between the robots and to the Base Station, enabling both high-level control of the robot behavior by the supervisor, as well as the reporting of detected objects of interest.

Alongside detailing the technological elements, we present how the robots operated in the underground mine environments of the Tunnel Circuit and the abandoned nuclear facility of the Urban Circuit and discuss technological adjustments in our solution. We compare our mapping results with ground-truth provided by DARPA, specifically present the results of both the onboard and the multi-robot SLAM solutions, detail and discuss the performance of our autonomy pipeline and when the single-operator exerted control, and finally report the detected and missed objects of interest. Finally, we present an assessment of our team performance both at the system level and for each technological component and focus on the challenges faced and the lessons learned with the hope of informing other researchers contributing to this exciting field. As significant components of our research have been released as open-source contributions, a selected list of relevant software packages is also provided.

The remainder of this paper is structured as follows: Section~\ref{sec:related_work} presents related work, while a high-level overview of the CERBERUS vision is outlined in Section~\ref{sec:system_of_systems}. The details about the individual robotic systems are presented in Section~\ref{sec:cerberus_robots}. The two autonomy-enabling functionalities for exploration (path planning and multi-modal multi-robot perception) are presented in Sections~\ref{sec:path_planner} and~\ref{sec:multi_model_perception} respectively, followed by our object detection and reporting solution in Section~\ref{sec:artifacts}. The robot-deployable communications solution is outlined in Section~\ref{sec:networking}, while the Human Supervisor interface is discussed in Section~\ref{sec:operator}. Experimental results primarily from the DARPA Subterranean Challenge field deployments are presented in Section~\ref{sec:experimental_studies}, followed by lessons learned in Section~\ref{sec:lessons_learned}. Finally, we present our plans for future work in Section~\ref{sec:future_work}, followed by conclusions in Section~\ref{sec:concl}. An appendix is also provided containing links to selected open-source contributions of our team, alongside links from deployment results at the Tunnel and Urban Circuit events.

%% file: 02_related_work.tex
The domain of subterranean exploration has attracted the interest of a niche community. The works in~\cite{silver2006topological,baker2004campaign} present methods for topological exploration of underground environments based on edge exploration and the detection of intersections. This work has been evaluated using ground platforms with the Groundhog system demonstrating pioneering levels of exploration capacity. The contribution in~\cite{distributed_subt_exploration} presents methods for underground exploration using a team of UAVs. The works in~\cite{nikolakopoulos2019autonomous,mansouri2019visual} outline methods for underground robotic exploration - using aerial robots - via contour and junction detection. Similarly, the work in ~\cite{kanellakis2019open} utilizes methods based on open space attraction. The paper in~\cite{jacobsonlocalizes} focuses on the localization challenges in subterranean settings. From a systems perspective, the set of works in~\cite{morris2006recent,novak2015exploration,maity2013amphibious} overview innovations in the domain of ground and submersible subterranean robotics. An overview of the utilization of ground robotic systems in underground environments is provided in~\cite{tardioli2019ground}. Furthermore, the methods in~\cite{bakambu2007autonomous,losch2018design} outline navigation solutions for robotic rovers underground. It should be noted that the research challenges faced in such robotic missions relate to a breadth of research in the domains of perception, planning, control, communications and more. This is reflected in the relevance of the challenges faced in other competitions, including - but not limited to - the earlier ``Multi Autonomous Ground-robotic International Challenge (MAGIC)'' as reported in papers from several of the participating teams~\cite{olson2012progress,lacaze2012reconnaissance,boeing2012wambot,butzke2012university}. %

Accelerated by the kick-off of the DARPA SubT Challenge in September $2018$, the domain of underground robotic autonomy is currently experiencing rapid growth. The works in~\cite{palieri2020locus} and~\cite{Tabib2019SimultaneousLA} present accurate multi-sensor techniques for subterranean localization and mapping, while the work in~\cite{Bouman2020jpl} presents progress in the ability of legged robots to search such environments autonomously. At the same time, the contribution in~\cite{miller2020mine} uses multiple legged systems, while a relevant dataset of artifact classification is released by the same team~\cite{shivakumar2019pst900}. The work in~\cite{steindl2020bruce} focuses on a hexapod design that has benefits in negotiating the challenging terrain often encountered in subterranean applications. The work in~\cite{hines2020virtual} takes a different approach using virtual surfaces to propose a cohesive mapping and planning solution for ground robots. Aiming to explore scenes of interest better, the contribution in~\cite{wang2020visual} helps in identifying areas of visual interest through online learning, while the work in  ~\cite{carlbaum2020feature} proposes a deep \ac{CNN} architecture to establish a novel feature tracking scheme for robust localization. Focusing on the multi-robot operations aspect of underground autonomy, the authors in~\cite{rouvcek2019darpa,zoulabuilding,goel2020rapid,corah2019CorOMeGoe} present methods for multi-robot exploration, map sharing and communications. Contributing a largely different idea, the work in~\cite{huang2019duckiefloat} proposes the use of blimp system for long-endurance in underground exploration missions. Our team contributions are aligned with these community efforts and aim to support the development of subterranean robotic autonomy. %

%% file: 03_system_of_systems.tex
The Defense Advanced Research Projects Agency (DARPA) created the DARPA Subterranean Challenge (SubT Challenge) to incentivize research on novel approaches to explore, navigate, map, and search underground environments with teams of robots. The underground domain features key challenges for mobile autonomous systems including dynamic and rough terrain, limited visibility, poor communication, and hard access. Extensive research in the fields of autonomy, perception, network, and mobility is therefore needed to allow robots to explore these settings autonomously.

The SubT Challenge is structured in two parallel tracks, namely the ``Systems'' and the ``Virtual'' track. In the former, teams choose to develop their own hardware and software to compete in a physical course. In contrast, in the latter, teams choose to develop software-only solutions that are evaluated in simulation. The proposed underground scenarios for the System\textcolor{revision}{s} track include man-made tunnel networks extending for several kilometers and presenting vertical openings (``Tunnel Circuit'' - August $2019$), multi-level urban underground structures with complex layouts (``Urban Circuit'' - February $2020$), and natural cave environments with complex morphology and constrained passages (``Cave Circuit'' - August $2020$, \textcolor{revision}{cancelled event}). Finally, the challenges of the previous circuits are combined in a comprehensive benchmark mission (``Final Event'' - September $2021$).

The primary goal is to provide rapid situational awareness to small teams preparing to enter unknown and dynamic subterranean environments. Therefore the competing teams are requested to explore and map unknown underground scenarios, with the additional requirement that only one team member, the Human Supervisor, can manage and interact with the deployed systems. Teams are scored based on their ability to localize some specified objects, called ``artifacts'', that represent features or items of interest that could be reasonably found in subterranean contexts. The classes of the artifacts are known prior to each Circuit: \textit{Survivor}, \textit{Cell Phone}, and \textit{Backpacks} are common objects for all the Events; \textit{Drill} and \textit{Fire Extinguisher} are specific to the Tunnel Circuit whereas \textit{Gas} and \textit{Vent} can be found at the Urban Circuit. A total of $20$ such objects are distributed along the Competition Course. Their positions are initially unknown to the teams. If an artifact is localized and its position is reported within \SI{5}{\meter} of its surveyed position, the team scores one point. The artifacts' surveyed positions are expressed with respect to a DARPA\textcolor{revision}{-}defined coordinate frame. 

The Tunnel and the Urban Circuit - which relate to the focus of this paper - each featured two different mission courses in their respective environments. All teams were given two ``scored runs'' per course, scheduled on different days. The total final score for each Circuit was computed as the sum of a team's best run on each course. Before the beginning of each run, the team's Pit Crew (composed of up to nine people) had $30$ minutes to complete the necessary setup of their Base Station and of the robotic systems before the actual mission would start. The Base Station is a collection of one or more PCs used by the Human Supervisor to interact with the deployed robotics agents. Moreover, it serves as an interface for forwarding artifact reports and map updates to the DARPA Command Post, which evaluates the validity of the reports and provides score updates.
Each scored run lasted for $60$ minutes and was the only time available for the teams to explore and report the artifacts' positions.

\begin{figure}[H]
    \centering
     \includegraphics[width=0.9\columnwidth]{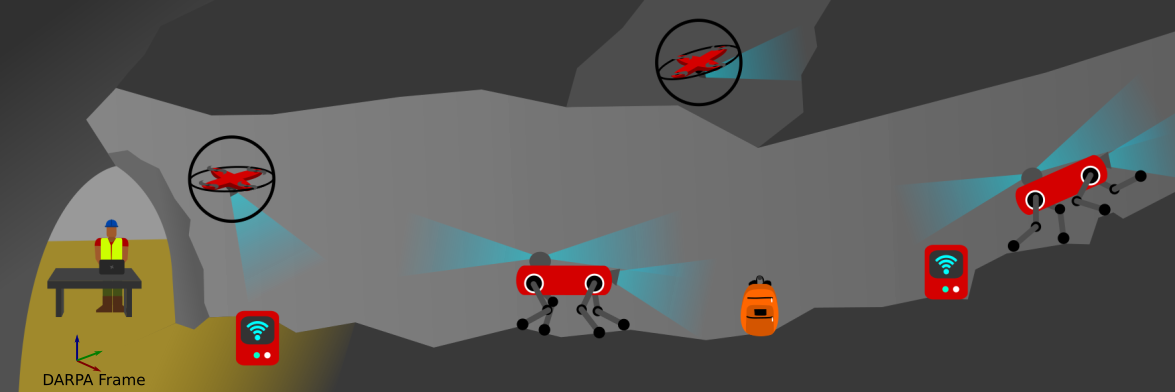}
    \caption{Graphical overview of the CERBERUS system: A single Human Supervisor outside the facility \textcolor{revision}{orchestrates the deployment of} a heterogeneous team of quadrupeds and aerial robots. Our ANYmal quadrupeds can traverse mobility hazards, while flying systems and especially the collision-tolerant robots can act as scouts or used to detect artifacts in chimneys and shafts. ANYmal deploys communication nodes to extend communication into the tunnel. The robots collaborate to reconstruct a consistent map, while artifacts detected are reported in a global frame defined by DARPA. }
    \label{fig:cerberus}
\end{figure}

Motivated by and responding to these challenges, the CERBERUS collaborative walking and flying robotic system-of-systems envisions the autonomous exploration of subterranean environments through the synergistic operation of diverse systems with unique navigation capabilities. The principal idea behind CERBERUS' vision is that the combination of legged and aerial systems offers unique advantages tailored to the needs of complex underground settings. Walking robots present potential benefits for overcoming complex terrain, both in terms of the size of obstacles that can be negotiated for a certain overall robot size, and robust traversal of dynamic terrain. Aerial robots, especially when equipped with collision-tolerance capabilities, offer the advantage of seamless navigation that is not bound to the extreme terrain challenges often present in environments such as caves. CERBERUS focuses on the co-deployment of such legged and aerial robots that are further equipped with resilient multi-modal and multi-robot localization and mapping, artifact detection and universal exploration path planning capabilities. A robot-deployable communications network solution is being realized which allows the CERBERUS system-of-systems to set up its own communications network underground. Figure~\ref{fig:cerberus} depicts an illustration of a CERBERUS deployment. At the time of writing, a subset of the technological components, methods and solutions have been developed and deployed incrementally at the Tunnel and Urban Circuit events of the competition. The highlights of these developments are presented in the rest of the paper. An appendix is provided that lists the open-source repositories containing our research to facilitate its reproducibility and verifiability, which in turn can help accelerate the developments of the community as a whole. 

%% file: 04_subterranean_walking_robots.tex
Team CERBERUS developed and deployed a team of legged robots specialized to operate in underground environments. These robots are based on the ANYbotics ANYmal B300~\cite{hutter2016iros} series quadruped, and feature several hardware and software modifications to create a specialized subterranean version: the ANYmal B ``SubT'' platform.

The ANYbotics ANYmal B300 series quadrupedal robot is a robust and agile legged system that is tailored for autonomous, long endurance tasks under challenging environments. Each leg has three compliant joints with integrated electronics which provide torque controllability and resilience against impact. This series of quadrupeds have been developed for over $10$ years and have achieved significant robustness, dexterity and mobility, while maintaining a small size and weight with an endurance of $1$ to $2$ hours. In contrast to wheeled or tracked systems, legged robots have the potential to adapt to terrain challenges by stepping over obstacles~\cite{jenelten2020} or adjusting their posture to crawl under obstacles~\cite{buchanan2020planning}. 
They also allow for omnidirectional locomotion, which is advantageous inside narrow environments.
Additionally, legged machines can be combined with wheels to achieve high speed and energy efficiency, combining the best of both worlds~\cite{Bjelonic2020}.
For this reason, legged robots are increasingly used in industrial inspection applications ~\cite{bellicoso2018advances} and in the underground domain~\cite{kolvenbach2020towards}.
Consequently, other DARPA Subterranean Challenge competitors also deployed quadrupedal robots during the Tunnel and Urban Circuits. 
For example, team \emph{PLUTO} deployed four \emph{Ghost Robotics Vision 60} during the Tunnel Circuit~\cite{miller2020mine}, and team \emph{CoSTAR}, which used wheeled systems during the Tunnel Circuit, later deployed two~\emph{Boston Dynamics Spot} robots~\cite{Bouman2020jpl}.

\begin{figure}
    \begin{subfigure}{.48\textwidth}
      \centering
      \includegraphics[scale=0.23]{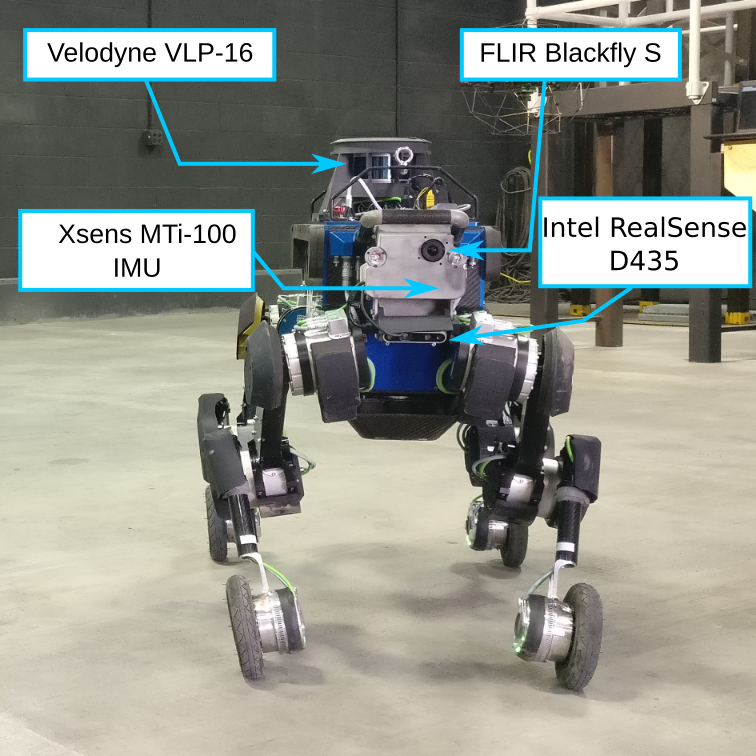}
      \caption{Tunnel Circuit}
      \label{fig:anymal_tunnel}
    \end{subfigure}%
    \begin{subfigure}{.48\textwidth}
      \centering
      \includegraphics[scale=0.23]{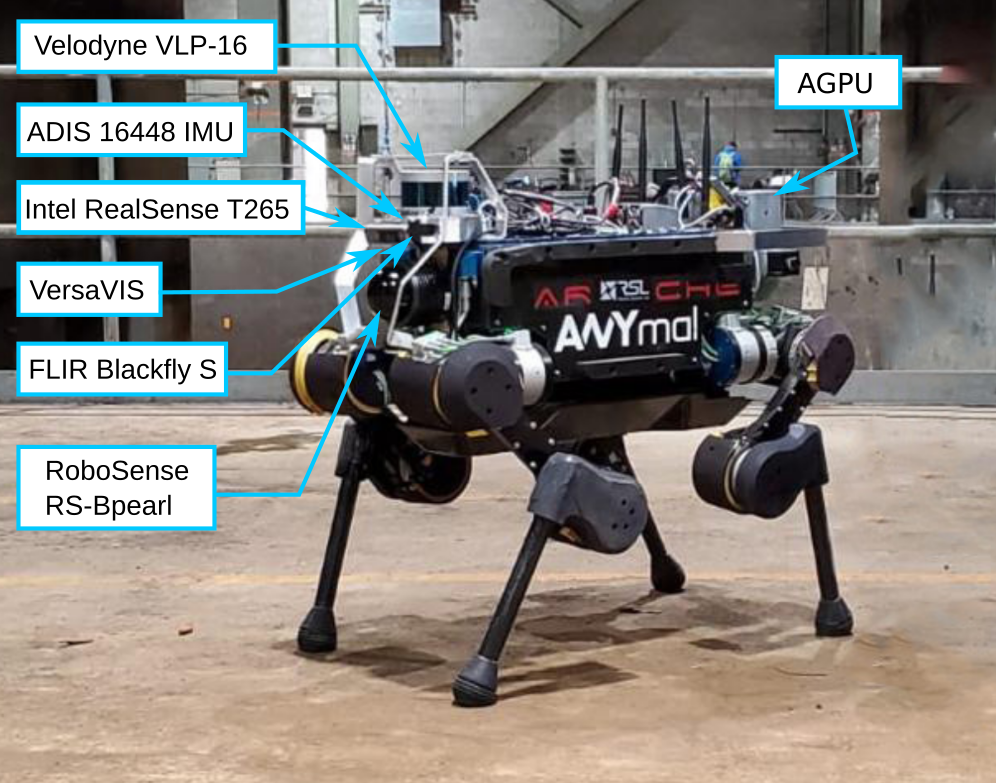}
      \caption{Urban Circuit}
      \label{fig:anymal_urban}
    \end{subfigure}
    \caption{ANYmal B SubT robot with specialized hardware and computation payload for deployment in underground settings.}
    \label{fig:sensors_front}
\end{figure}

\begin{figure}
    \centering
    \includegraphics[scale=0.25]{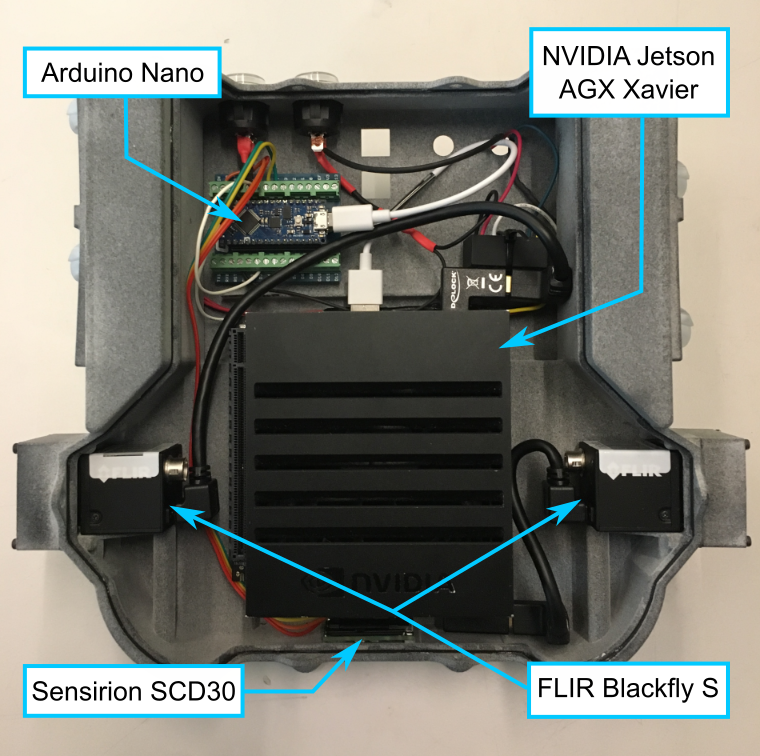}
    \caption{Auxiliary Graphics Processing Unit (AGPU) package enclosing a Jetson AGX Xavier, an Arduino Nano microcontroller, a Sensirion SDC30 gas sensor and two FLIR Blackfly S cameras.}
    \label{fig:jetson_case_second_iteration}
\end{figure}

\textbf{Hardware Modifications}\label{sec:anymal_modifications}\\
\emph{Sensors} - For the Urban Circuit, the onboard sensors were placed in two main locations atop the robot. A front facing aluminum case included an Intel RealSense T265 Camera, a FLIR Blackfly S color camera, an IMU, a VersaVIS synchronization board~\cite{Tschopp2020}, a Velodyne Puck VLP 16 LiDAR, and a Robosense BPearl LiDAR. A rear 3D-printed housing enclosed a Sensirion SDC30 gas sensor and two FLIR Blackfly S cameras. The location of each component on the robot can be seen in Figures~\ref{fig:sensors_front} and Figure~\ref{fig:jetson_case_second_iteration}. All components in the front aluminum case are protected by a sheet metal structure. In order to have \textcolor{revision}{a minimally obstructed field of view}, the metal around both LiDARs is parallel to the pulsed laser beams of the respective LiDAR. This setup is an improved version of the one deployed for the Tunnel Circuit, where an Intel RealSense D435 Camera was used instead of the Robosense BPearl LiDAR.

\emph{Terrain Mapping Sensors} - It is essential to ensure accurate terrain perception for safe navigation and locomotion during autonomous missions. Therefore terrain mapping sensors should be accurate and reliable in challenging environmental features such as darkness or reflective surfaces. Three different sensors were compared: the Intel RealSense D435, the Pico Monstar and the Robosense BPearl. The RealSense is an active stereo sensor that uses stereo matching with two IR cameras and complements it with active depth sensing based on the projection of an IR pattern. The Pico Monstar is a time-of-flight camera that is devised for long ranges and a wide Field-Of-View (FOV). The Bpearl is a dome-LiDAR that measures the reflection times of emitted laser pulses using a hemispherical scanning pattern. An overview of the comparison results is given in Table \ref{tab:depth_sensors}. 
The Pico Monstar and the RealSense were found to be prone to reflections (showing missing data on wet ground) and other conditions of the surrounding environments (puddles, absorbent material). Therefore the Bpearl was chosen as the terrain mapping sensor unit for the Urban Circuit, given its accuracy and robustness in various conditions and its large $90\si{\degree}$ FOV.

\begin{table}[b]
   \centering
   \begin{tabular}{p{0.13\linewidth} | p{0.13\linewidth} | p{0.37\linewidth} | p{0.12\linewidth} | p{0.12\linewidth}} 
   \toprule
   \textbf{Sensor} & \textbf{Type} & \textbf{Note} & \textbf{Size (\si{\milli\meter})} & \textbf{Weight (\si{\gram})} \\
   \midrule
   RealSense & Active Stereo & Less accurate but functional under sunlight & $90$x$25$x$25$ & $72$ \\
   \midrule
   Pico Monstar & Time-of-Flight & Accurate but weak to sunlight or absorbent material & $66$x$62$x$29$ & $142$ \\
   \midrule
   BPearl & LiDAR & Accurate, robust against sunlight or absorbent material but sparse & $100$x$111$x$100$ & $920$ \\
   \bottomrule
   \end{tabular}
   \caption{Comparison of sensors for terrain mapping.}
        \label{tab:depth_sensors}
\end{table}
    
\emph{Parallel Computing} - The terrain mapping algorithms can be efficiently parallelized using a graphics processing unit (GPU), while visual object detection also benefits from running neural-networks on a GPU. A Jetson AGX Xavier was installed as an Auxiliary Graphics Processing Unit (AGPU). The unit is enclosed in a 3D-printed PA-6 housing for protection against dust and water. Figure~\ref{fig:jetson_case_second_iteration} shows the content of the AGPU.%

\emph{Foot variants} - To augment terrain specific locomotion capabilities, the legged robots were equipped with three different types of feet. The standard version of ANYmal comes with point feet which adapt to any inclination angle but are prone to slip. The flat feet developed in~\cite{Valsecchi2020} have a larger contact surface and therefore provide more robust traction. To clear a considerable distance in a short time, ANYmal was also equipped with actuated wheels~\cite{Bjelonic2020}. 
    
\begin{figure}
    \begin{subfigure}{.48\textwidth}
      \centering
      \includegraphics[scale=0.21]{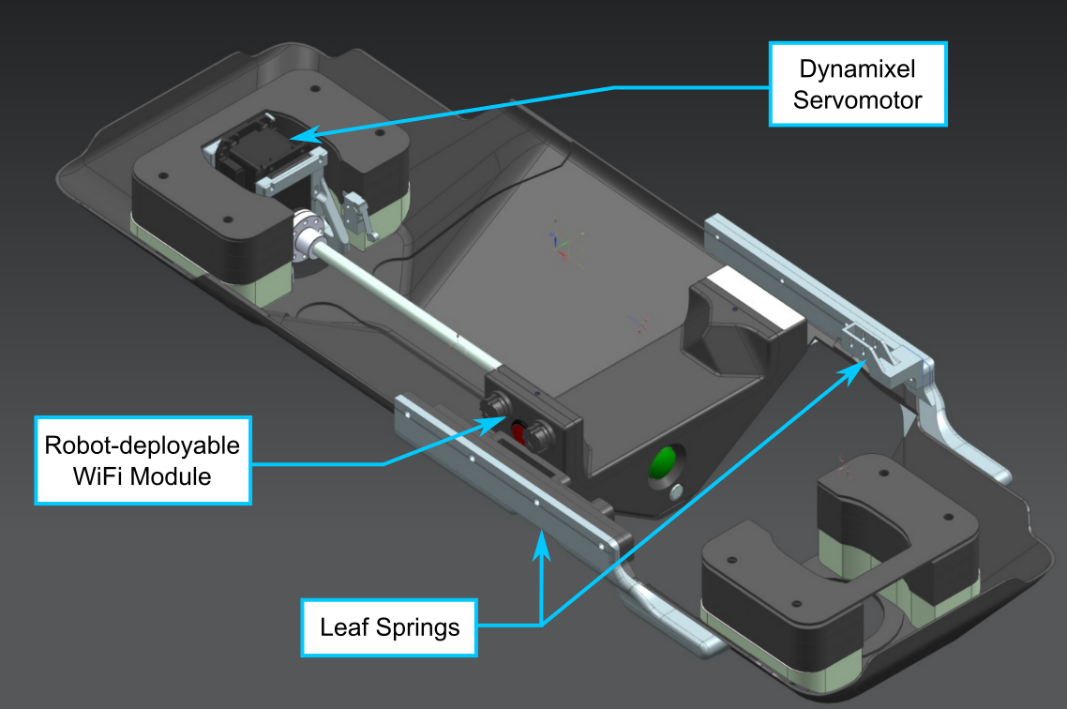}
      \caption{Release mechanism.}
      \label{fig:belly_plate}
    \end{subfigure}%
    \begin{subfigure}{.48\textwidth}
      \centering
      \includegraphics[scale=0.21]{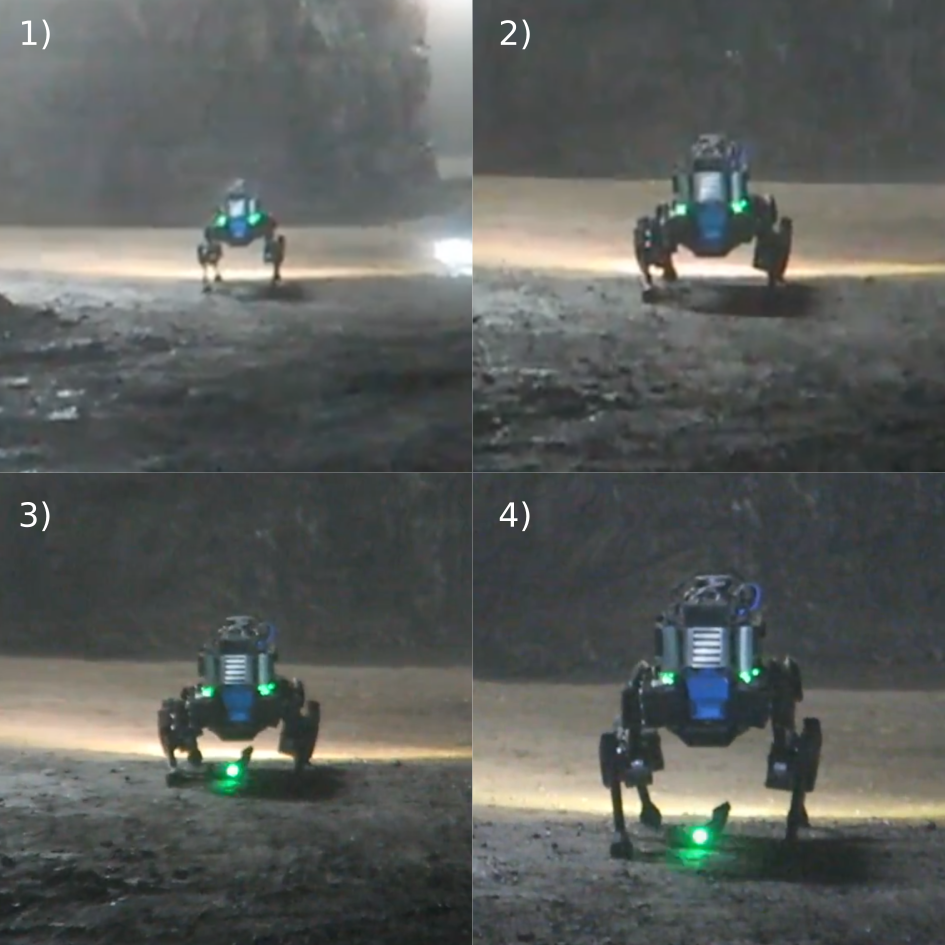} 
      \caption{WiFi beacon deployment.}
      \label{fig:deploy_wifi}
    \end{subfigure}
    \caption{ANYmal's payload release mechanism and one deployable module (\subref{fig:belly_plate}). ANYmal deploying a WiFi beacon on the ground (\subref{fig:deploy_wifi}).}
    \label{fig:wifi_beacon}
\end{figure}

\emph{Payload Release Mechanism} - The original belly plate of the robot, which protects the main body from impacts, was modified to carry and release specialized WiFi Breadcrumbs (described in Sec.~\ref{sec:mesh_node}), as depicted in Figure \ref{fig:wifi_beacon}. A Dynamixel servomotor drives a slide screw which converts the rotary motion into linear movement of a pusher plate. The modules are pushed towards an opening in the belly plate where leaf springs prevent accidental release.
    
\textbf{Software Modifications}\\
\emph{Locomotion}\label{sec:anymal_locomotion} -  Two different locomotion controllers were deployed on the ANYmal robots. During the Tunnel Circuit, a model-based perceptive controller was utilized~\cite{jenelten2020}. The footholds and  base motion are sequentially optimized using a simplified description of the robot dynamics \textcolor{revision}{and while accounting} for the local terrain properties in a batch search. Additionally, a slip reflex~\cite{Jenelten2019} stabilizes the robot in case of major slip events. In the Urban Circuit, a neural-network controller was deployed. It was trained in simulation using reinforcement learning~\cite{lee2020learning}.
The controller relies solely on proprioceptive sensor information and uses a history of joint and IMU measurements to implicitly infer information about the terrain and external disturbances without depending on the local terrain map.
While the model-based controller performs motion planning with anticipation of terrain changes, the reinforcement learning-based controller has significantly increased locomotion robustness in various challenging environments\textcolor{revision}{. During the Tunnel Circuit, ANYmal experienced corrupted local terrain representations leading to incorrect locomotion controller actions
that prevented the robot from continuing the mission. Therefore at the Urban Circuit the learning-based locomotion controller was employed due to its robust performance, as well as its characteristic ability to plan footholds without relying on an explicit representation of the local terrain.}

\begin{figure}
    \centering
    \includegraphics[width=0.6\linewidth]{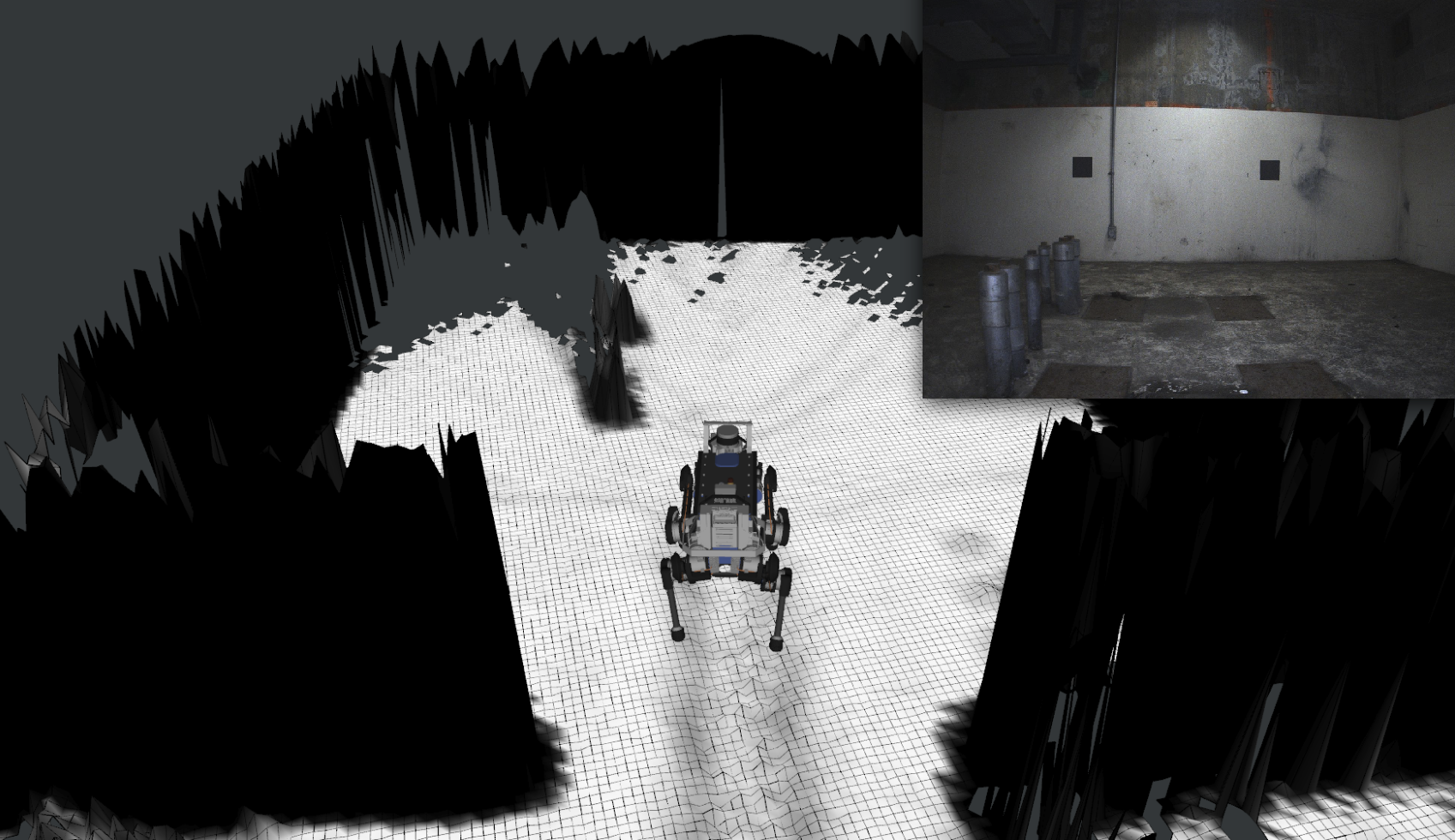}
    \caption{Elevation map built by ANYmal during the Urban Circuit. Black cells indicate non-traversable regions as classified by the learning-based traversability estimation.}
    \label{fig:elevation_map}
\end{figure}

\emph{Terrain Perception} -   As legged robots are constrained to move on the ground plane, a terrain representation in the form of a $2.5$D elevation map is commonly used. It is built in a robot-centric fashion from depth sensor readings and robot odometry~\cite{fankhauser2014robot}. \textcolor{revision}{The map has dimensions equal to \SI{12}{}x\SI{12}{\meter}, with a resolution of \SI{4}{\centi \meter}. These settings reflect the sensor limitations and represent a trade-off between capturing the surface of the terrain accurately and keeping the computation cost to a reasonable level for onboard processing.}
In addition to the height value, a traversability value is computed for each grid cell. The onboard state machine uses this information to \textcolor{revision}{prevent the robot walking into ``non-traversable'' areas.}
Traversability values are computed with a two-layer \ac{CNN} where the first layer applies multiple learnable traversability templates to the height map and the second layer computes a weighted sum of template activations~\cite{wellhausen2021rough}. 
Since this network has only $120$ learnable parameters, it can be trained using only a dozen hand-labeled elevation maps and allows for rapid inference as it operates directly on the elevation maps in GPU memory. A visual representation of the elevation map with associated traversability estimation is shown in Figure~\ref{fig:elevation_map}.

\emph{Path Tracking} - The path tracking module accepts paths from the autonomy module that are created either by the the high-level exploration planner or by the Human Supervisor. This module is organized in a hierarchical manner: a path manager converts the received paths (defined in an inertial frame) into a drifting frame in which the robot's locomotion controller operates. Subsequently it triggers the path following module which computes twist commands for the robot to follow the given path.

\begin{figure}
    \centering
    \includegraphics[scale=0.3]{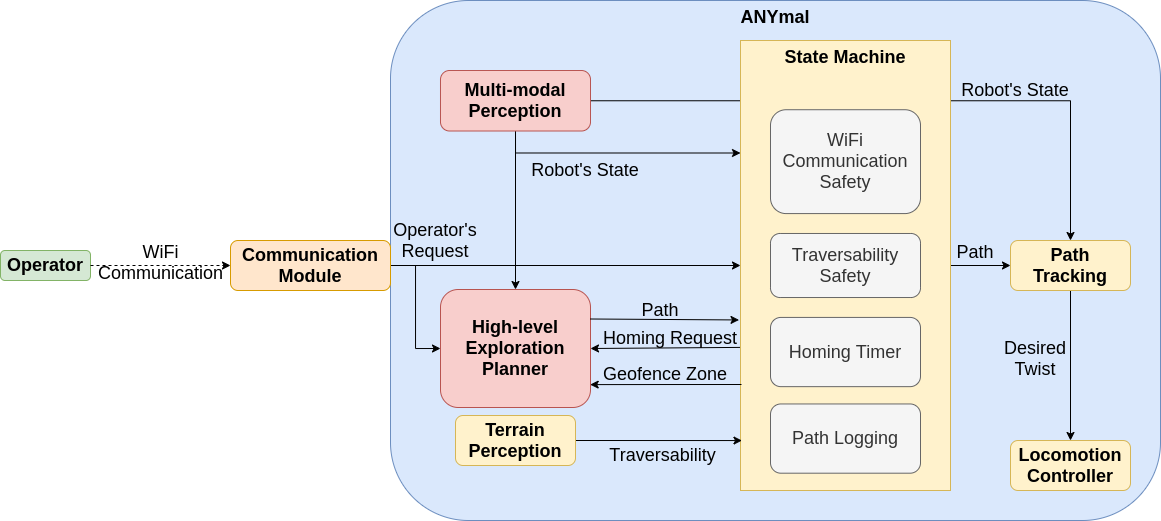}
    \caption{Schematic system overview of the main software modules on ANYmal, specifically tailored for underground exploration missions. The components in the blue shaded area are executed on the onboard computer. The yellow rectangles are ANYmal specific modules, whereas red rectangles indicate software modules used also by the other CERBERUS' robots.}
    \label{fig:mission_autonomy}
\end{figure}

\emph{State Machine} -   The capabilities for coordinating different software modules to execute autonomous missions are provided by a multi-faceted state machine, whose interaction with the other main modules is depicted in Figure~\ref{fig:mission_autonomy}.
Its primary purpose is to minimize Human Supervisor interactions by automatically coordinating between a high-level exploration planner, providing way-points where the robot should go, and a path tracking controller that steers the agent to the desired location. The state machine features a safety layer that monitors and responds to events such as non-traversable terrain, unexpected obstacles or loss of communication. This software module operates in two different modes depending on the operator's requirements: (a) supervised-autonomous and (b) fully-autonomous mode. The former case is used when the operator has a data connection to the robot, can check the high-level exploration planner's goals, and can decide to overrule the given way-points. During this mode, if communication to the Base Station is unexpectedly lost, the robot is commanded to trace back its path until a connection is re-established and subsequently wait for the supervisor to make a decision. The latter mode is used to explore regions without WiFi coverage. In this case, the state machine does not take any action in case communication to the Base Station is lost. This mode can be activated by the operator, alongside setting a time limit for autonomous exploration.
Once this time has elapsed, a homing request to the high-level exploration planner is issued, which will compute a trajectory back to the Base Station.
It is critical to get the robot back into the region with WiFi coverage because the generated map and the identified artifacts need to be streamed back to the operator in order to score points.

Despite the current operating mode of the state machine, if obstacles are \textcolor{revision}{detected along the robot's direction of motion}, a traversability safety module is triggered. This \textcolor{revision}{module ensures that the robot reverses a fixed distance and informs the high-level exploration planner about the non-traversable area (called \textcolor{revision}{a} ``geofence zone''). The exploration planner will then re-plan a new path that avoids the obstacle.} \textcolor{revision}{The state machine uses the current robot’s direction to check if the agent is moving  towards a non-traversable area. The direction of motion is identified by performing a look-ahead projection of the robot's future poses (up to \SI{30}{\centi \meter} ahead) using its estimated velocity.} 
\textcolor{revision}{Non-traversable} obstacles are identified with the help of the traversability estimation layer of the elevation map. \textcolor{revision}{The estimated look-ahead poses are used to} anticipate obstacles \textcolor{revision}{and to stop the robot} sufficiently early. \textcolor{revision}{The state machine checks for the current robot’s direction of motion instead of the high-level exploration planner’s path for a set of reasons. Firstly, the robot can react to previously undetected non-traversable regions. Secondly, the portion of the elevation map closer to the agent has higher accuracy. Lastly, checking the traversability of a path of several meters in a high resolution map is computationally expensive. Therefore, checking for a relatively short range in front of the robot is preferred.}

%% file: 05_subterranean_aerial_robots.tex
The SubT Challenge provides several unique challenges to autonomous robotic operation. Mines, caves, and tunnels frequently contain many challenging terrain features and multi-level settings. Terrain obstacles, including stairs, shafts both vertical and horizontal, boulders, cliffs, and \textcolor{revision}{bodies of water} all impose significant hazards to ground robots. Simultaneously, the elevated perspective inherent to aerial platforms provides additional capabilities to observe, localize, and report artifacts. Despite limited endurance, aerial platforms can cover more area in less time in many environments. Furthermore, they may correspond to the only option for certain subsets of subterranean settings. Therefore, CERBERUS developed and deployed a team of ``Aerial Scouts'' to provide rapid exploration capabilities, especially in regions of the course which were inaccessible by ground robots.

\begin{table}[b]
\centering
\begin{tabular}{c|c|c|c}  
\toprule
& \textcolor{revision}{Alpha / Bravo} & Charlie & Gagarin \\
\midrule
     Platform   & DJI M100      & DJI M100       & Collision-tolerant platform  1-3  \\
     Camera     & FLIR Blackfly S & FLIR Blackfly S & Intel RealSense T265   \\
     Thermal    &  N/A      & FLIR Tau2      &      N/A    \\
     LiDAR      & \textcolor{revision}{Velodyne Puck LITE}         & Ouster OS1           & Ouster OS0  / OS1    \\
     IMU        & VectorNav VN-100        & VectorNav VN-100         & Integrated in autopilot   \\
\bottomrule
\end{tabular}
\caption{Sensor suites used in different Aerial Scouts.} \label{tab:aerial}
\end{table}

\textbf{System Design:} Three robots, named Alpha, Bravo, and Charlie, utilize the DJI Matrice 100 quadrotor as their airframe. For high-level processing, including sensor data collection, localization, mapping, and planning, an Intel NUC Core-i7 computer (NUC7i7BNH) is mounted on each robot, while an Intel Movidius Neural Compute Stick is used for artifact detection. As shown in Table~\ref{tab:aerial}, all three robots fuse LiDAR, visual and inertial data, while Charlie further integrates thermal vision and thus provides unique localization capabilities in visually-degraded environments, as well as thermal object detection (e.g., for the survivor artifact class). To enable perception in low-light environments, all three M100 Aerial Scouts are equipped with Cree LEDs, which are synchronized with their visible light camera's shutter producing a power-efficient flashing behavior. 

\begin{figure}[h!]
\centering
    \includegraphics[width=0.7\columnwidth]{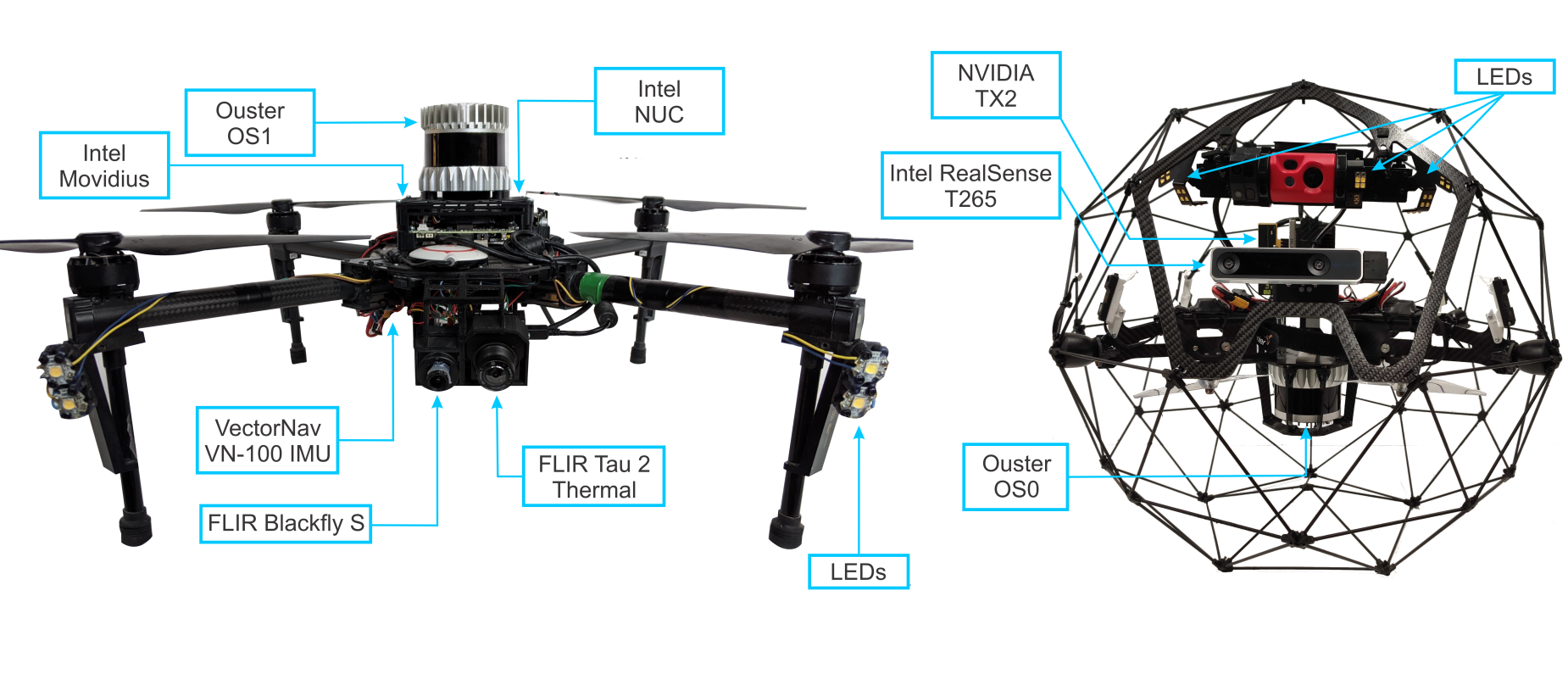}
\caption{The Charlie (left) and Gagarin (right) Aerial Scouts.}\label{fig:aerial_robot_design}
\end{figure}

A second - and particularly important for our team - class of Aerial Scouts, named Gagarin, also provides a number of unique capabilities. This robot is designed to traverse narrow and visually degraded environments.
Gagarin is a drone surrounded by a collision tolerant cage that allows it to sustain collisions with obstacles at speeds exceeding $2\textrm{m/s}$. In order to thrive in narrow spaces, this drone has a compact design with a tightly integrated power-efficient NVIDIA TX2 board with $6$ ARM CPU-cores and a $256$-core NVIDIA Pascal GPU, an Ouster OS0 or OS1 (depending on the version of the Gagarin robot) LiDAR having a wide vertical FOV, an IMU and two fisheye cameras, rendering it suitable for going through staircases and vertical shafts. Additionally, to traverse visually-degraded environments, Gagarin has a custom lighting system involving LEDs at different angles designed to reduce the effect of dust on the image, increasing the visual range and enhancing the environment texture.

All Aerial Scouts communicate with the Base Station using their onboard $5.8\textrm{GHz}$ WiFi radios to connect directly to the Base Station antenna or through the breadcrumbs deployed by our legged robots. Importantly, as reliable high-bandwidth communication becomes increasingly infeasible in most underground environments, the Aerial Scouts are able to operate completely autonomously with little to no input from the Human Supervisor, and return to their home position before their batteries become depleted. \textcolor{revision}{The M100 Aerial Scouts are capable of flight times of $8$ to $10$ minutes, while the Gagarin collision-tolerant aerial scouts present an endurance of $5$ to $6$ minutes.}

\begin{figure}[h!]
\centering
    \includegraphics[width=0.92\columnwidth]{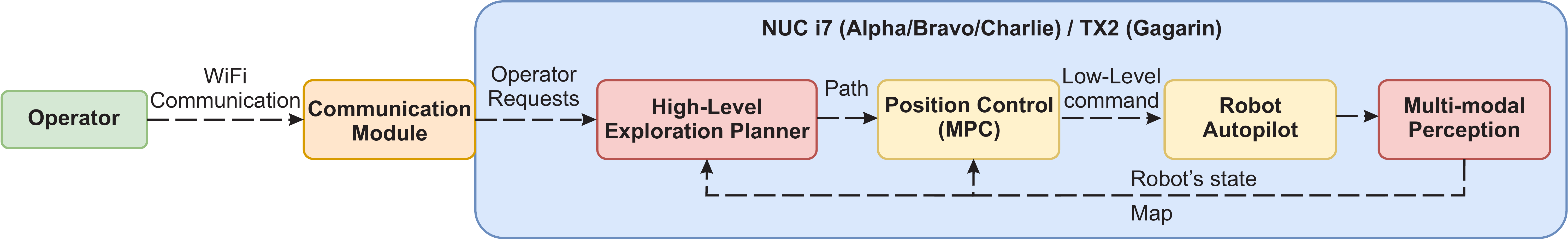}
\caption{Schematic system overview of the core functional modules of the aerial scouts. All the components inside the blue shaded area are executed onboard the robot. Yellow color depicts robot-specific functionalities, while red is used for software components that are unified across the CERBERUS robots. }\label{fig:blockdiagram_aerial}
\end{figure}

\textbf{Software Architecture:} The Aerial Scouts all share the same software architecture, whose basic functionalities are illustrated in Figure~\ref{fig:blockdiagram_aerial}. Similar to other robots in the CERBERUS team, the Aerial Scouts use a high-level exploration path planner, described in Section~\ref{sec:path_planner}, to plan feasible paths for the robots. The path is then tracked by a linear model predictive controller as in~\cite{mpc_rosbookchapter} and the low-level command is executed by the onboard autopilot inside each robot. The odometry feedback of the robots and the map of the environment are provided by the multi-modal perception module presented in Section~\ref{sec:multi_model_perception}.  

%% file: 06_roving_robots.tex
Alongside the \textcolor{revision}{legged and} aerial platforms, team CERBERUS also brought a roving robot called Armadillo into the field. This ground vehicle is complementary to the walking and flying systems presented above.

\textbf{System Design:} Armadillo is based on the Inspectorbots Super Mega Bot (SMB) robot platform and was chosen to complement the performance of the core CERBERUS team, primarily as a means to deploy a high-gain communications antenna and due to its capability to carry large payloads and sensors. A custom-designed frame, made of lightweight aluminum extrusions, was used to house all the components on the robot. This frame was fixed atop Armadillo's chassis and an onboard computer, namely a Zotac VR GO Backpack PC, was placed inside the protected cage-like structure. A sensor-head, attached at the highest part of the frame, houses various sensors to provide an unobstructed view of the environment.

\begin{figure}[h!]
\centering
    \includegraphics[width=0.45\columnwidth]{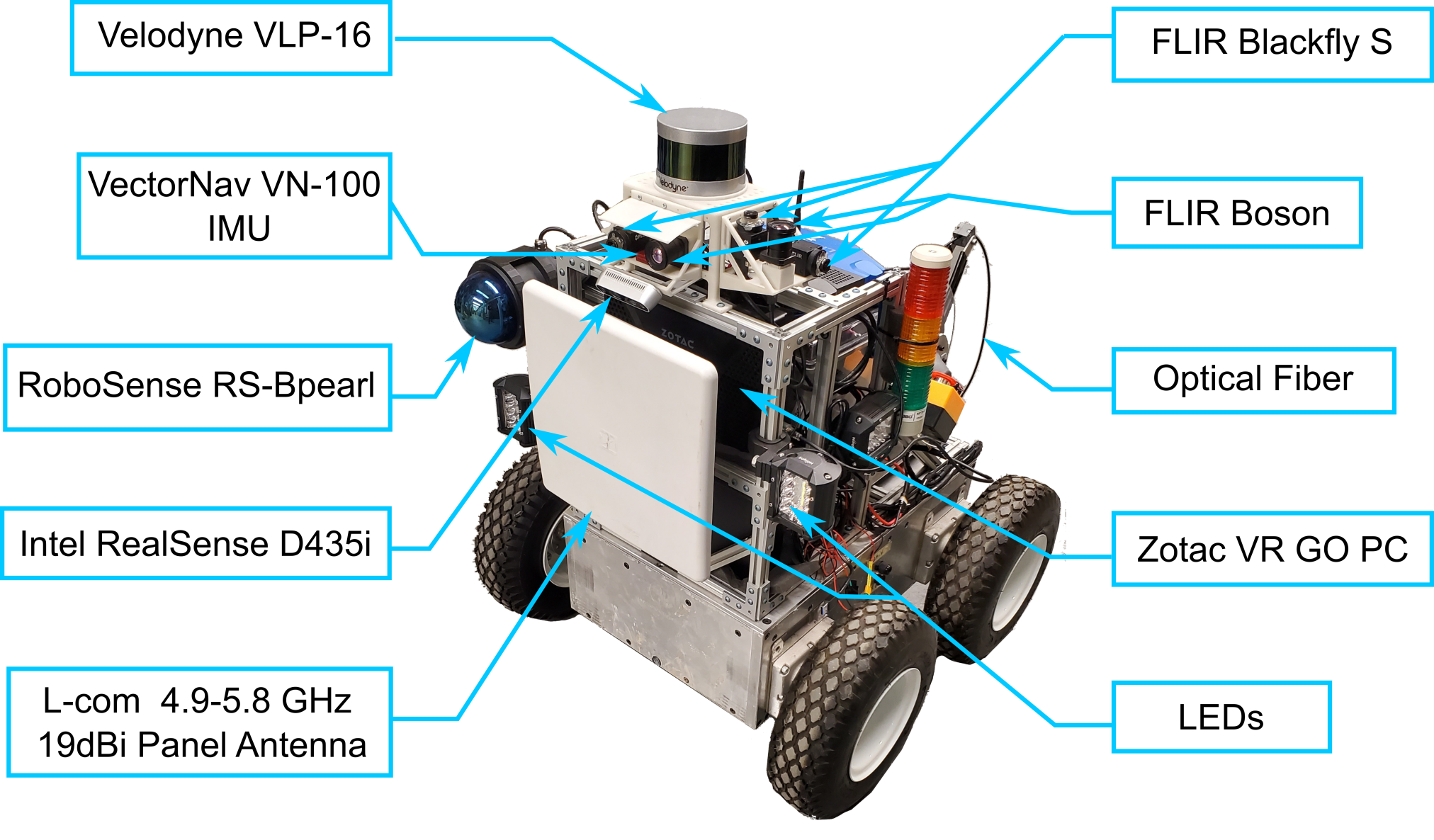}
\caption{The Armadillo roving robot - a modified Inspectorbots Super Mega Bot. }\label{fig:armadillo_description}
\end{figure}

The sensor-head on the Armadillo consists of a two level design having mounting points for all the sensors of the robot. A FLIR Boson LWIR Camera and a FLIR Blackfly Color camera are mounted in the front facing configuration alongside a VectorNav VN-100 IMU. Two identical FLIR Blackfly S cameras are mounted on the sides of the sensor looking at either side of the robot. Additional FLIR Boson and Blackfly cameras are mounted in a top-facing configuration for detecting artifacts on the ceiling.

A frame extension on the right side of the robot is made to mount a Robosense BPearl LiDAR Sensor that provides a dome-like view of the environment. At the back of the robot, a $300\textrm{m}$-long optical fiber reel is mounted with the capability of automatic unrolling while moving. This allows wired connectivity from the Base Station to the robot. An additional $5.8$GHz directional antenna is mounted in the front of Armadillo, which allows the robot to act as a network range extender, providing high-bandwidth connectivity to the Base Station.

\textbf{Software Architecture:} The Armadillo shared the same perception architecture as the flying robots. However one significant difference is that - given its constant connectivity with the base-station through the optical fiber connection and design choices of our team - it was manually controlled by the Human Supervisor. The front facing Blackfly camera, the IMU and the Velodyne LiDAR are used for the multimodal perception as described in Section~\ref{sec:multi_model_perception}. The remaining sensors provide a live feed to the operator at the Base Station.

%% file: 07_expl_path_planning.tex
The CERBERUS system-of-systems performs autonomous exploration and search for artifacts in the  subterranean environment based on a Graph-based exploration path planner (GBPlanner)~\cite{dang2020graph} that is universally designed and deployed across both legged and aerial robots. The method exploits a volumetric representation of the incrementally \textcolor{revision}{updated and} explored $3\textrm{D}$ occupancy map $\mathbb{M}$ reconstructed based on measurements from a set of onboard depth sensors $\{\mathbb{S}\}$, as well as robot poses derived from each robot localization system $\{\mathbb{O}\}$. \textcolor{revision}{In the Tunnel Circuit $\mathbb{M}$ was based on the work in~\cite{hornung13auro}, while in the Urban Circuit the method in~\cite{oleynikova2017voxblox} was employed. In both cases, $\mathbb{M}$ is using a fixed resolution, thus each voxel has an edge size $r_V$ (throughout this work this value was set to $r_V=0.2\si{\metre}$).} The map consists of voxels $m$ of three categories, namely $m \in \mathbb{M}_{free}$, $m \in \mathbb{M}_{occupied}$, or $m\in \mathbb{M}_{unknown}$ representing free, occupied, and unknown space respectively, while certain (generally disconnected) subsets of the map may correspond to ``no-go'' zones $\mathbb{M}_{NG}$ (hereafter referred to as ``geofences'') representing possible traversability constraints or other imposed limits. Let $d_{\max}$ be the effective range, and $[F_H, F_V]$ be the FOV in horizontal and vertical directions of each of the depth sensors $\mathbb{S}$. In addition, let the robot's configuration at time $t$ be defined as the $3\textrm{D}$ position and heading $\xi_t =[x_t,y_t,z_t,\psi_t]$. Notably, since for most range sensors perception stops at surfaces, hollow spaces or narrow pockets cannot always be explored fully. This leads to a residual map $\mathbb{M}_{\ast, res} \subset \mathbb{M}_{unknown}$ with volume $V_{\ast,res}$ which is infeasible to explore given the robot's constraints. As a result, given a volume $V_{\ast}$, the potential volume to be explored is $V_{\ast,explored} = V_{\ast} \setminus V_{\ast,res}$. 

\begin{figure}[h!]
    \centering
     \includegraphics[width=0.95\columnwidth]{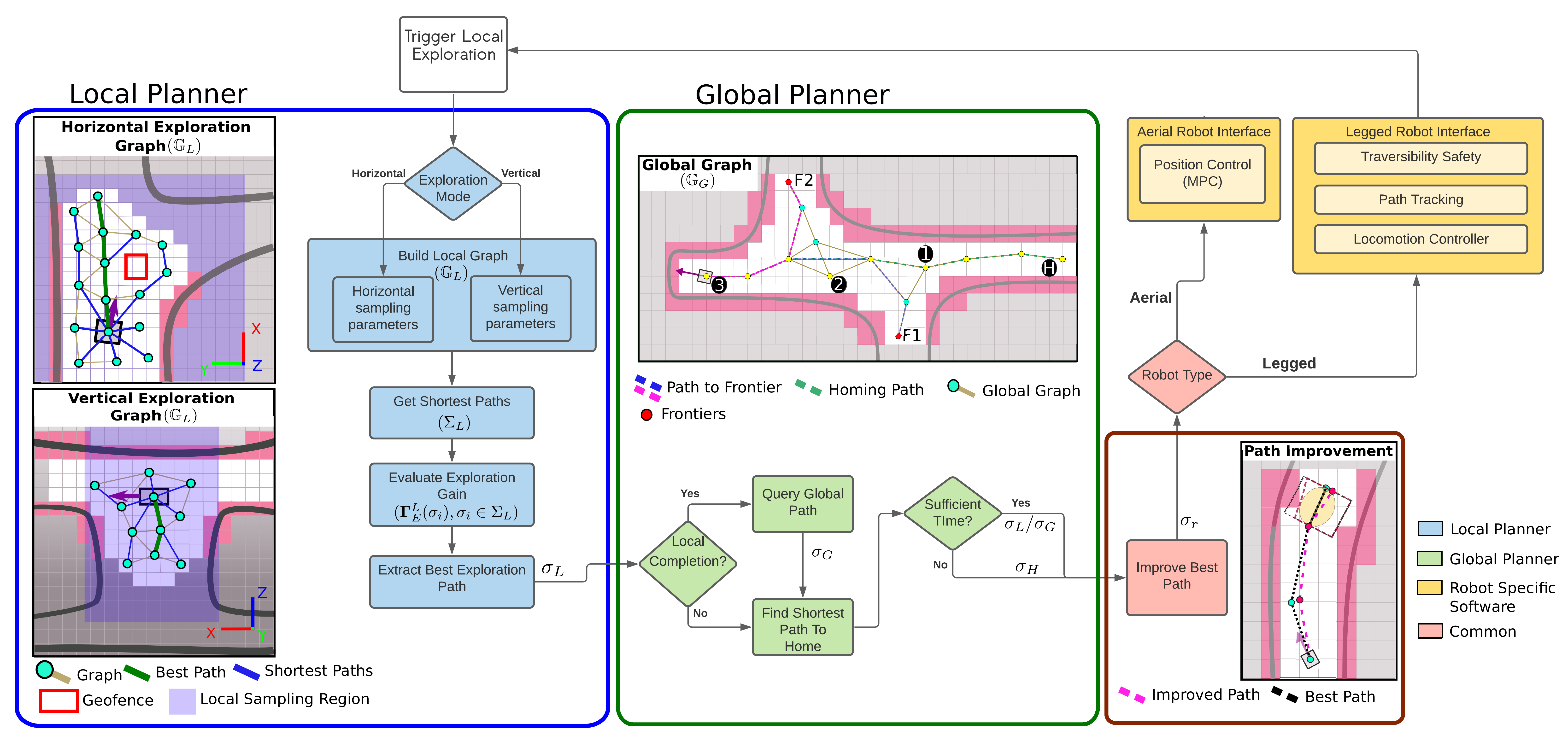}
    \caption{Functional diagram of the key functionalities of GBPlanner. This method bifurcates around a local and global planning mode with the first responsible for efficient exploration around the current robot location and the second responsible for re-position\textcolor{revision}{ing} to frontiers of the explored volume in distant areas of the map, as well as auto-homing. Special functionality is provided for multi-level exploration, while the admissible paths account for collisions with $3\textrm{D}$ obstacles as well as ``no-go'' zones (geofences) that mark the \textcolor{revision}{non-traversable} regions detected.}
    \label{fig:gbplanner_overview}
\end{figure}

Given this representation, the planning problem is organized based on a local and a global planning bifurcation. The overall planner architecture can be seen in Figure~\ref{fig:gbplanner_overview}. In the local exploration planning mode, given the occupancy map $\mathbb{M}$ and a local subset of it $\mathbb{M}_L$ centered around the current robot configuration $\xi_{0}$, the goal is to find an admissible path $\sigma_L = \{\xi_i\}$ in order to guide the robot towards unmapped areas and thus maximize an exploration gain $\mathbf{\Gamma}_E^L(\sigma_i), \sigma_i \in \Sigma_L (\textrm{set of candidate paths}),$ primarily relating to the volume expected to be mapped if the robot traverses along $\sigma_L$ with a sensor $\mathbb{S}$. A path is admissible if it is collision-free in the sense of not colliding with $3\textrm{D}$ obstacles in the map and not going through geofences. 

Given the subspace $\mathbb{M}_L$ the local planning problem may report that no admissible solution is interesting for the robot (zero-gain or below a set threshold). When this event, called ``local completion'', occurs, the global planning mode uses the latest $\mathbb{M}$ representation and the current robot configuration $\xi_{0}$ to identify a collision-free path $\sigma_{G}$ that leads the robot towards the frontiers of the unmapped areas. Feasible paths of this global problem take into account the remaining endurance of the robot. When the environment is explored completely (``global completion'') or the battery approaches its limits, the method identifies an admissible path $\sigma_{H}$ that returns the robot to its home position $\xi_{home}$. The functional specifics of these two modes are outlined below.

\vspace{-2ex}

\subsection{Multi-Level Local Exploration Path Planning}

The local planning mode of GBPlanner first iteratively builds an undirected graph $\mathbb{G}_L$ inside the local map space $\mathbb{M}_L$ around the current robot configuration $\xi_0$, which will be used to derive candidate exploration paths. \textcolor{revision}{In the implementation of the planner, up to the Urban Circuit, the local map subset $\mathbb{M}_L$ had a fixed cuboid size set at the beginning of the mission (details for each robot are provided in Section~\ref{sec:experimental_studies}). The graph $\mathbb{G}_L$ is built in the following manner.} Given the updated occupancy map $\mathbb{M}$, a bounded ``local'' volume $V_{D_L}$ with dimensions $D_L$ centered around $\xi_0$ and the respective map subset $\mathbb{M}_L$ \textcolor{revision}{are first derived. Then considering} a bounding box $D_R$ representing the robot\textcolor{revision}{'s} physical size (alongside the respective map subset $\mathbb{M}_R(\xi_{\ast})$ for a robot state $\xi_{\ast}$), the planner samples a set of random configurations $\xi_{rand}$ inside $V_{D_L}$ and after checking which local connections (within a defined radius $\delta$) are collision-free and outside of the geofences, builds the graph $\mathbb{G}_L$ \textcolor{revision}{with} its vertex and edge sets $\mathbb{V},\mathbb{E}$ respectively. Subsequently, using Dijkstra's shortest path algorithm~\cite{dijkstra1959note}, it derives paths $\Sigma_L$ in the graph starting from the root vertex at $\xi_0$ and \textcolor{revision}{leading} to all destination vertices. Given a path $\sigma_i \in \Sigma_L, i=1...\mu$ with a set of vertices $\nu_j^i \in \sigma_i, j=1...m_i$ along the path (corresponding to a set of robot $\xi$ configurations), an information gain, called $\mathbf{VolumeGain}$, relating to the new volume \textcolor{revision}{anticipated to be} explored is computed for each vertex $\nu_j^i$. \textcolor{revision}{To compute the $\mathbf{VolumeGain}$ at a vertex we count the number of unknown voxels inside a model of the sensor frustum (accounting for the sensor range and field-of-view) centered at the vertex. This is done by ray casting into the map with rays starting at the vertex. A fixed resolution of ray casting is employed.} The exploration gain $\boldsymbol{\Gamma}_E^L(\sigma_i)$  for the path $\sigma_i$ is \textcolor{revision}{then} calculated as follows: 

\vspace{-3ex}
\small
\begin{eqnarray}
 \mathbf{\Gamma}_E^L(\sigma_i)=e^{-\zeta \mathcal{Z}(\sigma_i,\sigma_{e})} \sum_{j=1}^{m_i}{\mathbf{VolumeGain}(\nu_j^i) e^{-\delta \mathcal{D}(\nu_1^i,\nu_j^i)}}
\end{eqnarray}
\normalsize
where $\zeta, \delta>0$ are tunable factors, $\mathcal{D}(\nu_1^i,\nu_j^i)$ is the cumulative Euclidean distance from vertex $\nu_j^i$ to the root $\nu_1^i$ along the path $\sigma_i$, and $\mathcal{Z}(\sigma_i,\sigma_{e})$ is a similarity distance metric between the planned path as compared to a pseudo straight path $\sigma_{e}$ with the same length along the currently estimated exploration direction. %
\textcolor{revision}{The heading of every planned vertex is generally aligned with the direction of the respective path segment. While legged systems are allowed to turn in place to ensure that the desired heading is reached, the aerial robots are subject to a maximum yaw rate constraint. Therefore, a reachable heading as close as possible to that of the direction of the path segment is assigned given the length of the path segment, the fixed reference velocity, and the yaw rate constraint.}
Importantly, this local planning stage methodology implements a specialized sub-bifurcation in its architecture to handle the case of multi-level/multi-storey environments as, for example, encountered in the SubT Challenge Urban Circuit. In this mode, the method biases its sampling of the local graph $\mathbb{G}_L$ in order to sample densely in the vertical direction. Considering that a depth sensor unveiling relevant space is available, this offers an artificial reward of exploration gain when the robot changes ``level'' and thus identifies connections from the current level to the next. This motivates the robot to take a staircase or other vertical passageway connecting two storeys of the environment. \textcolor{revision}{In the implementation of the method as deployed up to the Urban Circuit event, it is the responsibility of the Human Supervisor to select the appropriate local exploration mode.}  

In the case of legged robots, the path should not only be obstacle-free but also traversable. To that end, an elevation map \textcolor{revision}{(see Section~\ref{sec:anymal_b_subt})} is utilized to check for traversability \textcolor{revision}{along the robot's direction of motion, while following the planned path. The density of the elevation map is a trade-off between capturing the surface of the terrain accurately and keeping the computation cost to a reasonable level for onboard processing. Hence, the elevation map is calculated in a significantly smaller area than the size of the local map subset $\mathbb{M}_L$ which is used for exploration planning.} However, planning only within short distances can lead to inefficient exploration behavior. \textcolor{revision}{Therefore}, the exploration path is calculated based on the occupancy map \textcolor{revision}{(specifically $\mathbb{M}_L$)}, assuming \textcolor{revision}{that the areas outside the elevation map are traversable,} and the traversability check is done during path execution. If \textcolor{revision}{an area where the robot is walking towards} is found to be \textcolor{revision}{non-traversable}, the path execution is stopped, that area is marked \textcolor{revision}{with a} geofence, and the local planner is re-triggered. This \textcolor{revision}{process}, however, \textcolor{revision}{can lead to sub-optimalities as it may cause the robot to stop and re-plan when an initially unknown non-traversable region is detected which in turn can slow down the exploration. To better address this problem, longer range elevation mapping would be essential which in turn poses a set of computational and sensing challenges as discussed in Section~\ref{sec:anymal_b_subt}.}%

\subsection{Global Exploration Path Planning}

The global planning mode enables two critical behaviors in the exploration mission. First, upon local completion of unexplored areas, the global planning mode enables the re-positioning of the platform to unexplored areas of the map. Second, it permits autonomous homing upon reaching a critical point in the time budget of the robot. 
To execute these behaviors, the planner maintains a computationally lightweight global graph $\mathbb{G}_G$ that contains vertices which potentially lead to unexplored parts of the environment called ``frontier \textcolor{revision}{vertices'', or simply ``frontiers''}. \textcolor{revision}{A vertex is marked as a frontier if the $\mathbf{VolumeGain}$ of that vertex is found to be larger than a set threshold.} \textcolor{revision}{At each local planning step, the frontier vertices from the local graph $\mathbb{G}_L$ are extracted and the frontiers within a radius $\lambda$ of any existing frontiers in the global graph are removed. The paths from the remaining frontier vertices to the root vertex of $\mathbb{G}_L$ are extracted, clustered together according to the Dynamic Time Warping (DTW) similarity metric~\cite{bachrach2013trajectory,dang2020graph} and the principal path corresponding to the longest path from each cluster is retained. Finally, all the principal paths are added to the global graph, as well as admissible edges connecting them to nearby global vertices.}
In order to maintain a reduced list of frontiers in $\mathbb{G}_G$, vertices which no longer offer high gain are eliminated by periodically re-evaluating each frontier vertex. Additionally, \textcolor{revision}{at every local planning step, the planner needs to ensure that the robot has enough endurance to execute the path and return to the home location. Hence,} utilizing Dijkstra's algorithm on the global graph, the planner identifies a return-to-home path \textcolor{revision}{at each local planner iteration}, and commands the robot to follow this path upon reaching the time budget. \textcolor{revision}{It is noted that the homing path queries are fast due to the sparse nature of the global graph.}

Choosing the best frontier to visit upon local completion is a difficult problem due to the size of the map over which the search is executed, the uncertain knowledge of these frontier vertices, as well as their true potential for exploration. Given these challenges, \textcolor{revision}{the} selection of the frontier to visit depends on a) the time budget remaining upon reaching that frontier and b) the volumetric gain for that frontier given the time to reach it. Formally, let $\nu_{G,cur}$ be a vertex in the global graph representing the current robot configuration and $\mathcal{F} = \{\nu_{G,i}^\mathcal{F}\}, i=1...m$ be the set of updated frontiers. A Global Exploration Gain $\boldsymbol{\Gamma}_E^G(\nu_{G,i}^\mathcal{F})$ is evaluated for each frontier as follows:

\vspace{-2ex}
\small
\begin{eqnarray}
 \mathbf{\Gamma}_E^G(\nu_{G,i}^\mathcal{F}) = \mathcal{T}(\nu_{G,{cur}}, \nu_{G,i}^{\mathcal{F}}) \textbf{VolumeGain}(\nu_{G,i}^{\mathcal{F}}) e^{- \epsilon_\mathcal{D} \mathcal{D}(\nu_{G,{cur}}, \nu_{G,i}^{\mathcal{F}})}
\end{eqnarray}
\normalsize
where $\mathcal{T}(\nu_{G,{cur}}, \nu_{G,i}^{\mathcal{F}})$ is the estimated exploration time remaining \textcolor{revision}{if the planner chooses to visit} the frontier $\nu_{G,i}^{\mathcal{F}}$. \textcolor{revision}{This term is used to favor frontiers which lead to higher exploration times once they are reached. It is approximately} calculated using the \textcolor{revision}{remaining exploration time $\textrm{T}_e(t)$ (based on the robot's endurance or user-imposed mission time limit) at current instance $t$} and the \textcolor{revision}{estimated time required to travel} from the current vertex to that frontier\textcolor{revision}{, $\Upsilon(\nu_{G,{cur}}, \nu_{G,i}^{\mathcal{F}})$,} and from there to the home \textcolor{revision}{location, $\Upsilon(\nu_{G,i}^{\mathcal{F}}, \nu_{G,{home}})$,} as:

\vspace{-2ex}
\small
\begin{equation}
\textcolor{revision}
{
    \mathcal{T}(\nu_{G,{cur}}, \nu_{G,i}^{\mathcal{F}}) = \textrm{T}_e(t) - \Upsilon(\nu_{G,{cur}}, \nu_{G,i}^{\mathcal{F}}) - \Upsilon(\nu_{G,i}^{\mathcal{F}}, \nu_{G,{home}})
}
\end{equation}

\normalsize
The term $\mathcal{D}(\nu_{G,{cur}}, \nu_{G,i}^{\mathcal{F}})$ is the shortest path length from the current vertex to the frontier, which along with the tunable parameter $\epsilon_\mathcal{D}$ penalizes longer paths.

Finally, the exploration path, local or global, is further improved by modifying all vertices in the path to be further away from obstacles generating a safer path $\sigma_{r}$. The planner paths are provided to the robot-specific local planners and controllers which are responsible for their execution.

%% file: 08_distributed_multi_modal_perception.tex
\begin{figure}[h!]
\centering
    \includegraphics[width=1.0\columnwidth]{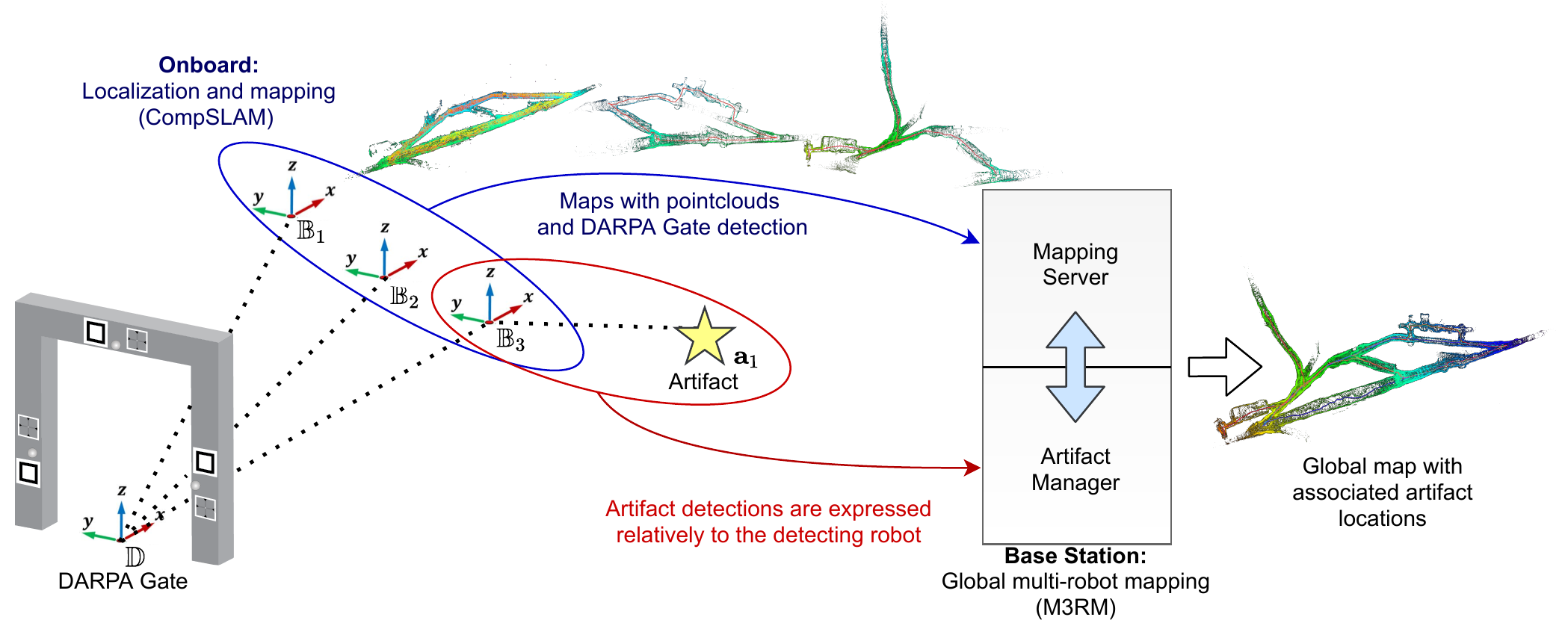}
    \caption{Illustration of the overall localization and mapping setup with coordinate frames for each robot $\mathbb{B}_1$, $\mathbb{B}_2$ and $\mathbb{B}_3$. All robots align themselves with respect to a common DARPA\textcolor{revision}{-}defined coordinate frame $\mathbb{D}$ using fiducial markers on the starting DARPA gate. The mapping server on the Base Station receives maps from the individual robots (blue) and combines them into a globally optimized map. Artifact detections, such as \textbf{a}$_1$, are recorded relative to the robot detecting them ($\mathbb{B}_3$) and are sent to the artifact manager (red) for storing, additional inspection, and reporting them in coordinate frame $\mathbb{D}$ using \textcolor{revision}{the} globally optimized map to the DARPA \textcolor{revision}{Command Post} for scoring.}
    \label{fig:mapping_overview}
\end{figure}

To enable robot autonomy in challenging subterranean environments, onboard perception plays a crucial role in enabling reliable robot localization and navigation. Resilient robot perception, both in terms of sensors and methods, is of particular importance for operation during the SubT Challenge as, in addition to being GPS-denied, underground environments can impose various conditions of sensor degradation due to their structural nature and operational conditions. Complete darkness, low-texture, geometric self-similarity, thermal-flatness, and the presence of obscurants (smoke, fog, dust) are some of the challenging conditions that can be encountered during the competition. Relying on any single sensing modality is unsuitable for operation in such environments, and hence, the robots of team CERBERUS utilize multi-modal sensor data from visual/thermal cameras, LiDARs, and IMUs in a complementary manner to provide reliability, robustness, and redundancy for underground robot operations. This section details the processing of the multi-modal sensor data to estimate the robot state and create a globally consistent multi-robot map. First, sensor time-synchronization and calibration are discussed. Second, the utilized complementary multi-modal fusion approach for onboard localization and mapping for individual robots is described. Finally, the deployed global multi-robot mapping approach is detailed, along with the alignment of individual robot frames $\mathbb{B}$ with the DARPA defined coordinate frame $\mathbb{D}$, and transmission of maps to the Base Station.
Figure~\ref{fig:mapping_overview} depicts the used notation with coordinate frames and provides a general overview of the deployed localization and mapping scheme. %
During the deployment, each robot builds and maintains an onboard map of its traversed environment. These maps are periodically sent to the Base Station mapping server, which combines individual robot maps and produces a globally optimized map. The self-contained onboard localization and mapping approach enables the robots to explore autonomously, i.e., a connection to the Base Station is optional. 
However, when connected, the global map allows, in particular, the Human Supervisor to have a combined and optimized overview of the explored environment.
\subsection{Time Synchronization and Calibration}
\label{sec:time-synchronization}
Precise time synchronization of the employed sensory systems is a crucial task for mapping and localization. Although the robots only move at moderate velocities, the range at which they detect visual landmarks and LiDAR points can be large. Given that small rotation changes can lead to large translations for points that lie far away from the sensor frame's origin, small timestamp inconsistencies between angular and linear measurements can lead to significant systematic errors when distant points are observed.

The Armadillo robot (during the Tunnel Circuit) and the ANYmal robots (during the Urban Circuit) employed a camera trigger board~\cite{Tschopp2020} capable of onboard synchronizing with a host computer using time translation. 
An integrated Kalman Filter estimates the clock-skew and offset between all sensors and the host computer. 
Additionally, since multi-camera systems inherently yield different exposure times for each camera, the trigger board performs exposure compensation using the mid-exposure times.

Moreover, a tightly-coupled illumination board was used to trigger a set of LEDs according to the cameras' exposure times for the flying robots and Armadillo. 
The LEDs are only turned on between the start of the first camera and the end of the last camera's exposure. This makes it possible to run the LEDs at a high brightness while minimizing heat and energy consumption.

For the purposes of artifact localization and detection, as well as robot localization and mapping, a number of calibrations are required. The color camera intrinsics and camera-to-IMU extrinsics were identified based on~\cite{furgale2013unified}. The intrinsic calibration parameters of the thermal camera were calculated using our custom designed thermal checkerboard pattern~\cite{ICUAS2018Thermal}. The camera-to-LiDAR calibration was derived based on an implementation of the work in~\cite{zhou2018automatic}.

The Aerial Scouts operate with hardware-assisted software camera-to-IMU synchronization. While the camera and the IMU are free-running, the internal functionalities of the VectorNav VN-100 IMU are used to receive a synchronization pulse from the FLIR Blackfly S camera at the start of each exposure. Contained within every IMU message is the time difference between the rising edge of the synchronization pulse and the sample time of the current IMU message. IMU messages are assumed to be received in near real-time, and therefore the image timestamps are corrected based on the time difference reported by the IMU subtracted from the IMU message receive time.

\subsection{Complementary Multi-Modal Localization and Mapping}\label{sec:localization}
\begin{figure}[h!]
\centering
    \includegraphics[width=0.6\columnwidth]{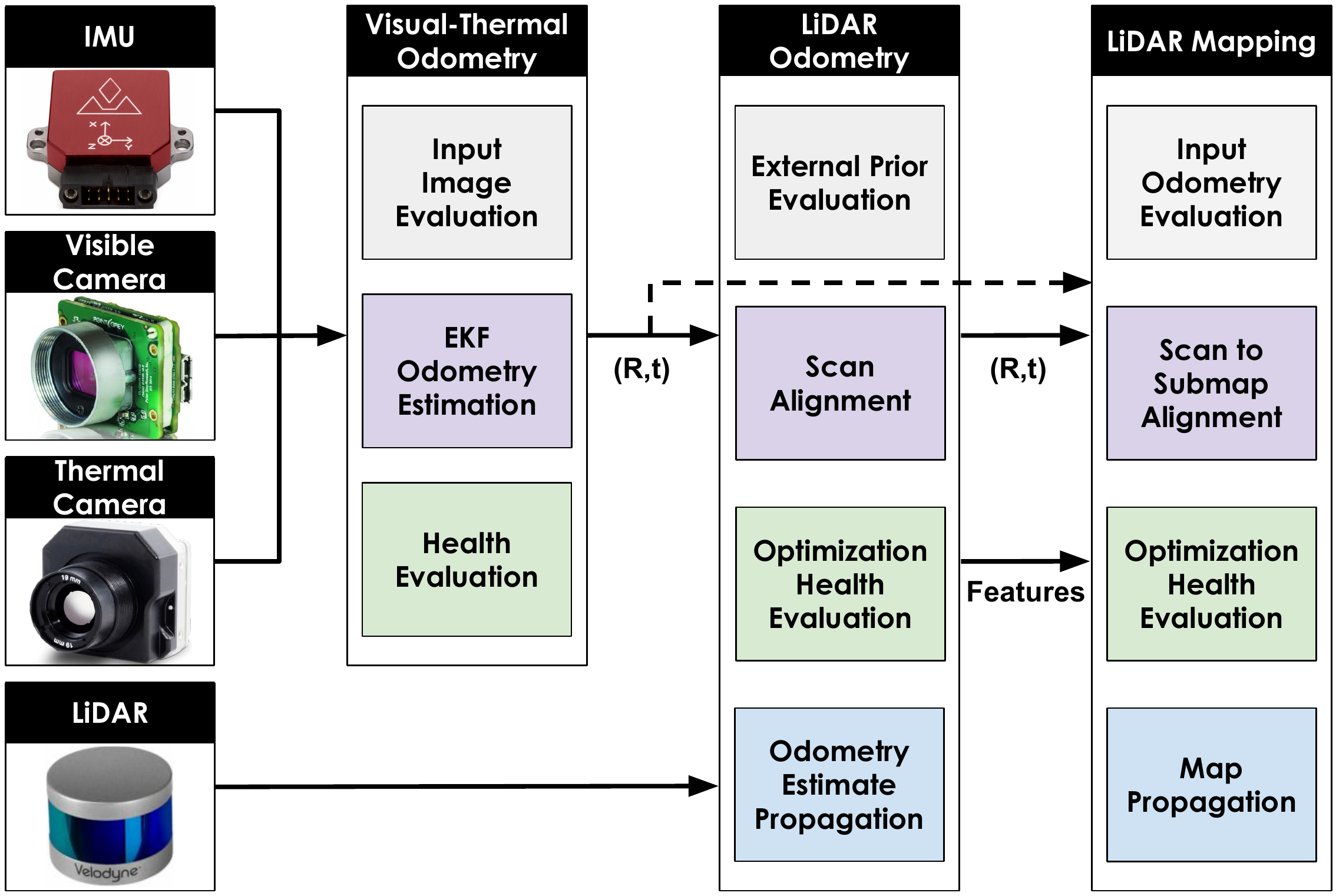}
    \caption{Overview of the deployed complementary multi-modal localization approach (CompSLAM). The Visual-Thermal-Inertial Odometry module receives camera and IMU data and provides a motion prior to both LiDAR Odometry and LiDAR Mapping modules to perform scan-to-scan and scan-to-submap alignments respectively. Motion estimates between modules are only propagated if they pass individual health checks.}
    \label{fig:comp_loc}
\end{figure}

To enable reliable and independent robot operation in challenging environments, the robots of team CERBERUS individually deploy an onboard complementary multi-modal localization and mapping (CompSLAM) approach~\cite{Khattak2020sensorfusion}. This approach fuses visual and thermal imagery, LiDAR depth data, and inertial cues in a hierarchical manner and provides redundancy against cases of sensor data degradation. A schematic overview of the proposed approach is shown in Figure~\ref{fig:comp_loc}. Camera images from both visual and thermal cameras\textcolor{revision}{,} as well as IMU measurements are fused using an Extended Kalman Filter (EKF) estimator. The deployed EKF-estimator builds on top of the work presented in~\cite{Bloesch2017rovio}, due to its low computational complexity, and extends it to concurrently operate on both visual and full-radiometric thermal images, as described in~\cite{Khattak2019rotio,ktio2020}.

To keep computational costs tractable while utilizing multi-spectral camera inputs, the Visual-Thermal-Inertial Odometry (VTIO) estimator tracks a fixed number of features in its state by balancing the share of tracked features from each image, subject to image quality using a spatial and temporal entropy evaluation~\cite{khattak2019visual}. The health of the odometry estimate is evaluated by measuring covariance growth using the D-Optimality criterion~\cite{dopt}. If the VTIO estimate is healthy, it is passed to the Laser Odometry (LO) and Laser Mapping (LM) modules. The LO module first independently ensures that the input estimate's translational and rotation components are within robot motion bounds, set by the controller, before utilizing it as a full or partial motion prior for the alignment of consecutive LiDAR scans.

Scan-to-scan alignment is performed by minimizing point-to-line and point-to-plane distances between extracted line and plane features, similar to the work of~\cite{Zhang2014loam}. The quality of scan alignment is checked by evaluating the minimum eigenvalues of the optimization Hessian matrix~\cite{jhang2016degeneracy}, and ill-conditioned dimensions are replaced with VTIO estimates before propagating the LO estimate and providing it to the LM module, along with extracted scan features. The VTIO and LO modules provide estimates at the update rates of images and point clouds, respectively; in contrast, the LM module operates at half the LO module rate to remain computationally tractable. For this reason, the LM module also independently evaluates input motion estimates from both modules before using them to estimate the refined robot pose in the local map using scan-to-submap alignment. Once a refined estimate is obtained, it is checked using similar health metrics in the LO step, and the final estimate is utilized for the propagation of the local robot map.

\subsection{Global Multi-Robot Mapping}\label{sec:multi_agent_mapping} 
In addition to the onboard CompSLAM, robots with an established connection to the Base Station send maps back to the mapping server\textcolor{revision}{,} as well as artifact detections to the artifact manager.
A centralized mapping server allows the Human Supervisor to view and inspect artifact detections along with a combined multi-robot representation of the environment.
Generally, the global multi-modal, multi-robot mapping (M3RM) approach is based on the existing framework Maplab~\cite{Schneider2017b} and divided into two parts. In the first part, each robot converts its raw sensor measurements into submaps that span consecutive time intervals. Each submap consists of a factor graph, visual descriptors, and compressed LiDAR pointclouds. These submaps are then transferred to the mapping server. Building the submaps on the robots reduces the required robot-to-mapping-server bandwidth\textcolor{revision}{,} while only using a modest amount of computing power on each robot. The second part takes place on the mapping server and consists of combining the individual submaps into an optimized, globally consistent multi-robot map.
The following subsections start by covering the submap creation and transmission approach in more detail, followed by the submap merging and global optimization strategy.

\emph{Submapping and Map Transmission over Network} - Each robot builds an on-robot pose graph comprising IMU, visual landmarks, and relative transformation constraints from the odometry. 
Furthermore, the on-robot pose graphs are divided into chunks (i.e. submaps) and are periodically sent to the Base Station to build a multi-robot map.
The composed system is generally independent of each robot's odometry estimation, i.e., the mapping system solely requires a consistent estimate of the robot's state and the IMU biases. 
Specifically, during the on-robot submap building, BRISK features are detected and tracked, and at submap completion, triangulated to construct a local map using the odometry information.
Consequently, this approach requires motions within a submap such that uninitialized landmarks can be triangulated.
In the case of motionless submaps, which commonly occur at startup, a stationary edge connects the first and the last vertex in the pose graph to avoid optimization issues.

In addition to visual landmarks, LiDAR scans are attached to the pose graph as resources using their timestamp and extrinsic calibration. 
The attached LiDAR scans are associated with a specific location in the pose graph and are not used for any additional computation on the robot itself but instead only on the Base Station.

Submaps are created by splitting the local maps at a fixed time interval and are uniquely identified using a hash of the timestamp and robot name.
To ensure consistency in the submaps when combining them at the Base Station, the first and last vertex of consecutive submaps overlap.
The server additionally ensures that the last and first vertex of consecutive submaps observe the same landmarks during the combination and merges them accordingly. Before transmission, each submap is keyframed~\cite{Dymczyk2015} and serialized to disk. A synchronization service then transfers the submaps to the Base Station in a granular fashion (file-based), such that transfers can efficiently be resumed after sudden connection losses. Whenever a submap transfer is completed, the synchronization service notifies the mapping server, which in turn processes and merges the submap into the global map.

\begin{figure}[!htb]
    \centering
     \includegraphics[scale=0.7]{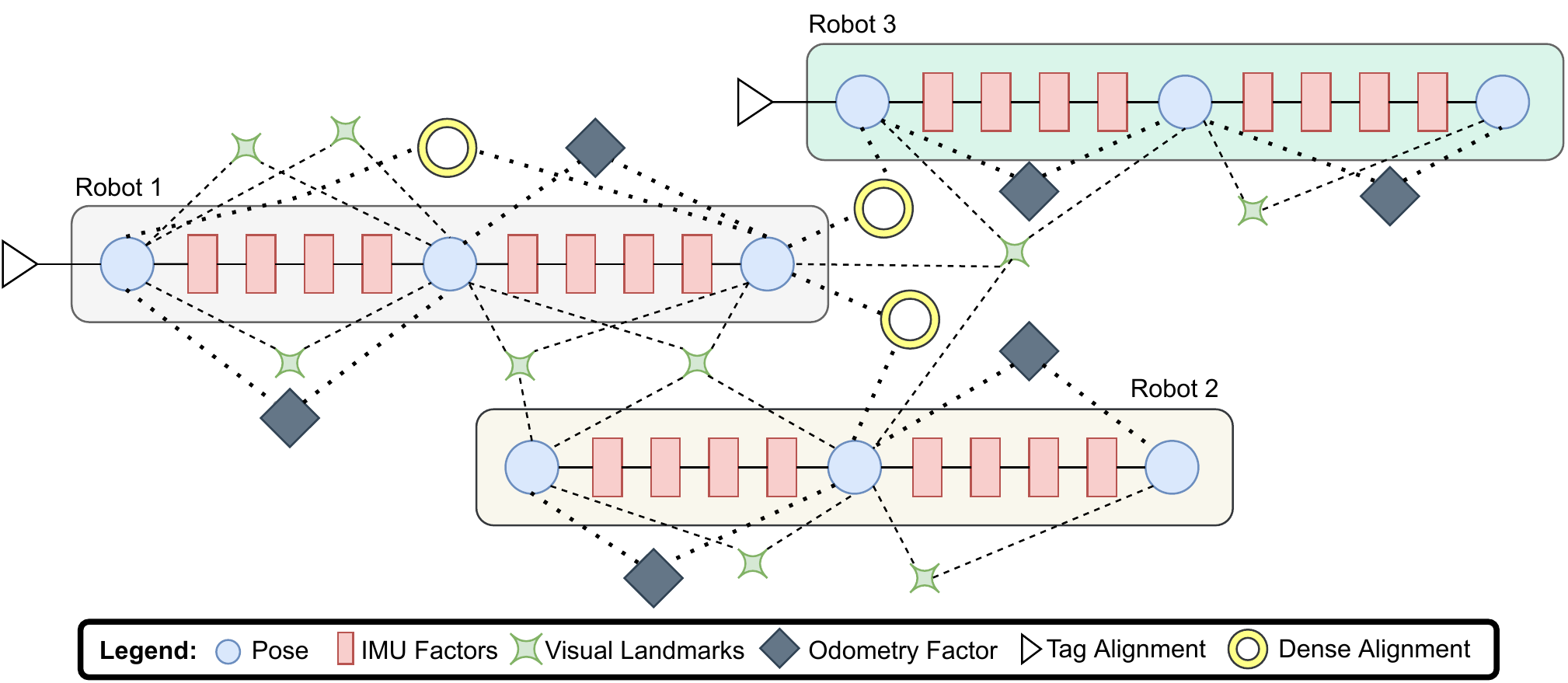}
    \caption{Illustration of three multi-modal pose graphs in the Base Station's mapping server. The graphs from individual robots are combined by co-observed landmarks or aligned pointclouds. Robot 1 and 3 detected the DARPA gate, whereas robot $2$ was merged using shared landmarks.}
    \label{fig:maplab_pose_graph}
\end{figure}

\emph{Globally Consistent Multi-Robot Mapping} - The mapping server on the Base Station periodically checks for completed submap transmissions and maintains a global map comprising all individual maps per robot. 
During the optimization of the global map, the mapping server utilizes different types of constraints as depicted in Figure~\ref{fig:maplab_pose_graph}. 
Maps from individual robots are connected by either merging visual landmarks (global visual loop closure) or performing a dense alignment between spatially close poses. 
Moreover, each robot performs an alignment to the DARPA frame $\mathbb{D}$ by detecting AprilTags on the Starting Gate with known 3D coordinates. 
The camera position is found in DARPA frame coordinates by solving the perspective-n-point problem of the $2$D tag detections and its result is added to the pose graph as a prior factor. 
Incorporating the DARPA frame alignment into the optimization has the advantage of including detection uncertainties, being more robust against outliers, and being independent of the utilized sensor system.
Additionally, this facilitates the global map merging since the base frame of the robot's map is already known. 

Before merging a robot's submap into the global map, each submap is processed by itself. 
The submap processing performs a fixed number of operations, i.e., (i) a visual landmark quality check (ii) visual loop closure (iii) LiDAR registrations, and (iv) optimization. 
Since the individual submap processing is independent of other submaps, the mapping server processes up to four submaps concurrently.
Furthermore, the optimizations are limited to a maximum of two minutes to ensure fast iterations. 

All new visual landmarks are first evaluated in terms of their distance, angle, and the number of observations. 
Landmarks that do not fall within the thresholds for these criteria are marked as disabled during the optimization.
Next, visual-based loop closures~\cite{Lynen2015} are found using the tracked visual BRISK features and an inverted multi-index.
Loop closures are handled by merging the visual landmarks in the pose graph, enabling the optimization to close loops solely using the reprojection error.

Furthermore, additional LiDAR constraints are extracted by registering consecutive scans within a submap using \ac{ICP}.
The current states of the poses in the submap graph are used to determine a transformation prior for the registration. 
Registration results that moderately contradict the prior transformation are rejected as outliers to avoid constraints across floors or through walls.
The remaining registrations are added as relative $6$DoF constraints between each scan's closest pose in the submap where the IMU is additionally utilized for interpolation between the timestamp of the scan and the pose.
The optimization assumes a fixed uncertainty for each registration constraint.

After the optimization, the submaps are merged into the current global multi-robot map. 
Similar to the submap processing, the mapping server continuously performs a set of operations on the global map, i.e., (i) global visual loop closure, (ii) global LiDAR registrations, and (iii) optimization. 
In contrast to the previous submap processing, each operation utilizes all information from each robot's currently explored areas. 
Hence, the LiDAR constraints are found based on a proximity search within the global map. 
For each pointcloud associated with the global pose graph, a radius search is performed to retrieve spatially close registration candidates, including poses to other robots.
Furthermore, all constraints resulting from LiDAR registrations are implemented as switchable constraints~\cite{Sunderhauf2012}.

The global multi-robot map provides an overview of the explored areas to the Human Supervisor and allows the operator to command robots to unexplored regions. Last but not least, artifact detections are utilized together with the reconstructed map to query their location. When only the map onboard the robot is available, the artifact locations are directly acquired, while when the M3RM-based map is also available, then it yields optimized artifact locations.

%% file: 09_artifacts.tex
In the SubT Challenge, teams are scored by reporting the class and location of a set of artifacts to the DARPA Command Post. A fixed number of report attempts are given to each team for each Competition Course. A point is earned for each report which correctly identifies the class of an artifact and its position is within $5\si{\metre}$ of the object's ground-truth location. Throughout both competition circuits, survivors, cell phones, and backpacks were placed alongside drills and fire extinguishers in the Tunnel Circuit, followed by $\text{CO}_{2}$ (gas) sources and vents in the Urban Circuit. In this section, multi-modal artifact detection and classification are described, including the initial detection onboard the robot, as well as how the detected artifacts are attached to the global map presented in Section \ref{sec:multi_agent_mapping} such that their estimated positions can be improved using the combined measurements of all robots.
\subsection{Detection and Classification}\label{sec:artifact_detection_and_classification}
Artifact detection and classification initially takes place onboard each robot. For the visible artifacts, survivors, cell phones, backpacks, drills, fire extinguishers, and vents, visual artifact detection using deep learning was employed. Specifically, we used YOLOv3~\cite{redmon2018yolov3} and its ROS porting~\cite{darknet_ros}, trained on datasets featuring all DARPA-specified artifacts except $\text{CO}_{2}$ gas. \textcolor{revision}{The training data consists of images from both the aerial and ground platforms, in laboratory settings against backgrounds of varying color, as well as in subterranean environments in varying lighting conditions. In addition, handheld versions of robot sensor configurations along with the on-board lighting solutions were re-created to facilitate rapid data collection and to improve viewpoint variation in the dataset. The number of images used to train each class varies from a minimum of $1915$ for the cell phone, to a maximum of $3519$ for the fire extinguisher. Table \ref{tab:artifact_count} details the number of training images used for each artifact class.}

\begin{table}[h]
\centering
\begin{tabular}{c|c|c|c|c|c|c}  
\toprule

\textcolor{revision}{Artifact} & 
\textcolor{revision}{Survivor} & \begin{tabular}[c]{@{}c@{}}\textcolor{revision}{Fire} \\ \textcolor{revision}{Extinguisher}\end{tabular} &
\begin{tabular}[c]{@{}c@{}}\textcolor{revision}{Cell} \\ \textcolor{revision}{Phone}\end{tabular} & \textcolor{revision}{Drill} & \textcolor{revision}{Backpack} & \textcolor{revision}{Vent} \\
\midrule
\textcolor{revision}{No. Images} & \textcolor{revision}{2660} & \textcolor{revision}{3519} & \textcolor{revision}{1915} & \textcolor{revision}{3084} & \textcolor{revision}{3388} & \textcolor{revision}{2301} \\

\bottomrule
\end{tabular}
\caption{\textcolor{revision}{Number of images used to train YOLOv3 for each class of visible artifact.}} \label{tab:artifact_count}
\end{table}

In addition to automatic detections, compressed image streams are sent back from the robots to the Base Station, where the Human Supervisor is located. Using the Operator Interface described in Section \ref{sec:operator}, the supervisor can pause and rewind these streams and manually select artifacts by clicking on the images. Given the bounding box of an automatic detection or a click from the supervisor, the interface further allows to manually add, translate, relabel and remove artifacts attached to the global map that can be reported to the \textcolor{revision}{DARPA Command Post}.

Due to the low precision-recall for visual cell phone detection, each robot could also detect the phone's Bluetooth signal, and report its presence to the Human Supervisor. In the Urban Circuit, all ground systems additionally integrated a Sensirion SCD30 $\text{CO}_{2}$ sensor to detect the gas artifact, reporting the position above which the $\text{CO}_{2}$ concentration had passed some threshold. %
\subsection{Map Attachment}\label{sec:artifact_map_attachment}
The attachment of artifacts to the global map is performed in two steps. The first step takes place onboard and consists of filtering the artifact detections to determine the artifact's position in the robot's local coordinate frame. Referring to Figure \ref{fig:mapping_overview}, we denote the estimated position of the $j\text{th}$ artifact in the body frame of the robot which detected it with $^{\mathbb{B}}\textbf{a}_j$. In the second step, an artifact report is transferred to the mapping server, where it is attached to the pose graph and its optimized position in \textcolor{revision}{the} DARPA frame $^\mathbb{D}\textbf{a}_{j,opt}$ can be computed.
\begin{figure}[!b]
    \centering
         \includegraphics[width=0.95\columnwidth]{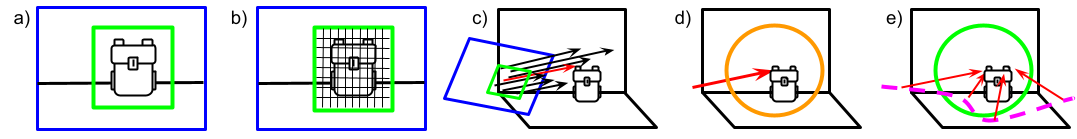}
    \caption{Illustration of the process of casting automatic detections into the map. a) Shows the detection of a backpack artifact with \textcolor{revision}{the} image frame in blue, and the bounding box in green. b) The bounding box is divided into a fixed grid. c) A ray from each grid cell is cast from the image plane until it intersects a surface. d) The median ray, in red, is selected and a sphere of radius $R_a$, in orange, is placed in the map. e) Subsequent rays from additional detections along the path shown in pink, which land within the sphere, increase the confidence of the artifact category until reaching a threshold.}
    \label{fig:bayesian_ray_casting}
\end{figure}

Due to the possibility of false positives and to prevent continuous artifact detections from saturating the robot's network connection to the Base Station, detections are first cast into the robot's map, and then filtered before being sent to the Base Station, providing the Human Supervisor with only high confidence detections. Specifically, the bounding box containing the detected object in the image plane is divided into a grid of pixels. These pixels are converted into rays using the camera model and the robot's extrinsic calibration parameters. Casting these rays into $\mathbb{M}$ results in a set of points that include points on the object, as well as other surfaces contained within the bounding box. To approximate the object's true position, the median point is selected as the measured artifact position. To filter both multiple class detections and refine the estimate of the artifact's position $\textbf{a}_j$, a sphere with radius $R_a$ is drawn at the initial estimate of $\textbf{a}_j$. Additional detections whose projected positions lie within this sphere are considered to be detections of the same object. For each sphere, separate binary Bayes filters for each artifact class are recursively updated to estimate the probability that the sphere contains an object of a given class, and the sphere's center $\mathbf{a}_j$ is further updated as the average position of all detections projected into the sphere. Once the probability of any one class exceeds a predefined threshold, the artifact's detection is considered ``confirmed''. A summarized detection report is sent to the Human Supervisor alongside a representative image of the artifact, and the corresponding sphere is frozen to prevent any additional detections and subsequent duplicate reports. This process is depicted in Figure \ref{fig:bayesian_ray_casting}.

Despite effort for automated localization of the cell phone's position via a sequence of pose-annotated Bluetooth-based detections and tracking of their RSSI values, it was determined that the most reliable method would be to notify the Human Supervisor of the robot pose at which RSSI values of a cell phone were detected. The supervisor could then improve the cell phone's position by moving a marker on the map or by drawing a bounding box around the cell phone in the camera images. In case the latter option was used, the bounding box would be sent back to the robot, which would then convert it to a 3D artifact position $\mathbf{a}_j$ using the method described in the previous paragraph. In the Urban Circuit, this same approach was used to determine the gas artifact positions.

The artifact reports sent from the robot to the mapping server consist of the artifact's type, the timestamp $t_i$ at which it was ``confirmed'' and the artifact's filtered position in the robot's local frame $^{\mathbb{B}_{t_i}}\mathbf{a}_j$. The annotated camera image in which the artifact was detected at time $t_i$ is also included to provide additional context to the Human Supervisor. Once received by the mapping server, the artifact report is attached to the pose graph node $T_{\mathbb{D}\mathbb{B}_{t_i}}$ that corresponds to the pose of \textcolor{revision}{the} robot's body in \textcolor{revision}{the} DARPA frame $\mathbb{D}$ at timestamp $t_i$. The artifact's most recent optimized position can then be computed as $^{\mathbb{D}}\mathbf{a}_{j,opt}=T_{\mathbb{D}\mathbb{B}_{t_i},opt}\ ^{\mathbb{B}_{t_i}}\mathbf{a}_j$. When desired, the artifact's unoptimized position can also be recovered with $^{\mathbb{D}}\mathbf{a}_{j,unopt}=T_{\mathbb{D}\mathbb{B}_{t_i},unopt}\ ^{\mathbb{B}_{t_i}}\mathbf{a}_j$.

%% file: 10_networking.tex
\begin{figure}[h]
    \centering
     \includegraphics[scale=0.2]{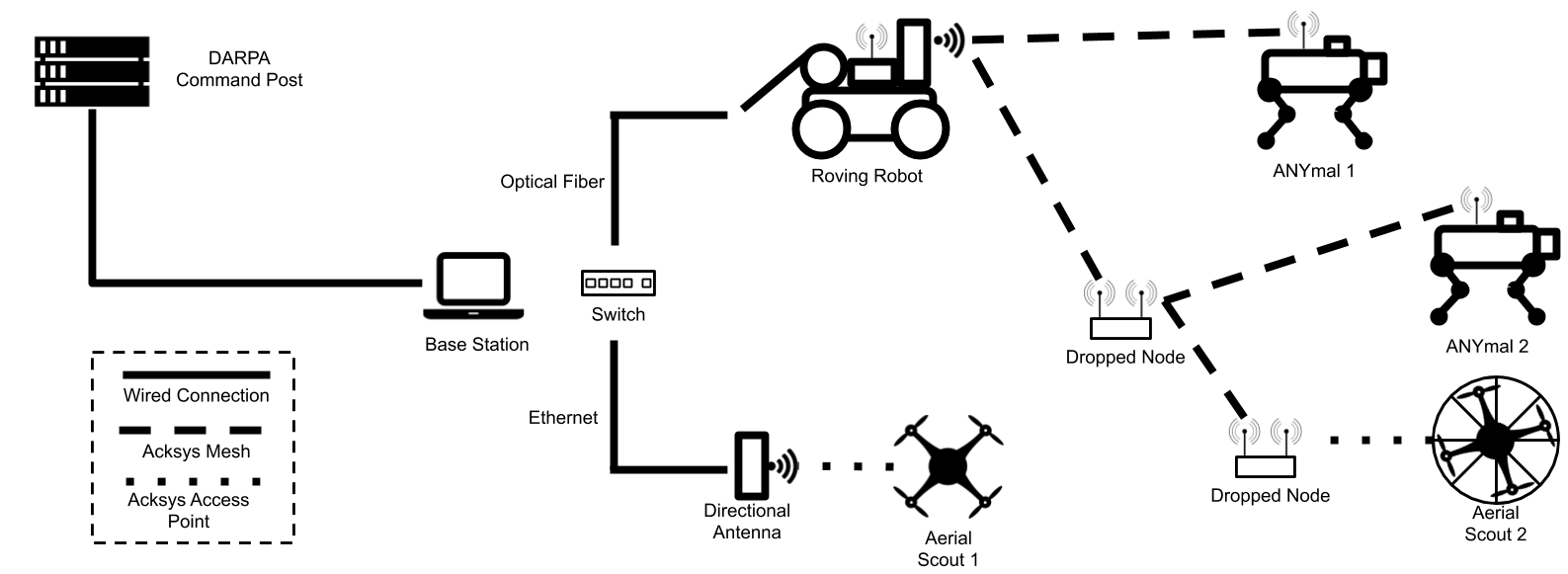}
    \caption{Block-diagram view of the networking infrastructure used during  the Tunnel and Urban Circuits. Roving and quadrupedal robots, as well as dropped communication nodes, create an ad hoc mesh network, which establishes the backbone communication infrastructure. The roving robot was also equipped with a fiber optical cable to maintain connection to the Base Station. Aerial scouts used a standard ($802.11$) WiFi network, created either by a directional antenna at the Base Station or by a dropped breadcrumb.}
    \label{fig:network_diagram}
\end{figure}

During mission deployments, all robots communicate with the Base Station via an ad hoc wireless network to share mapping data, artifact locations, and other information. This network operates in the $5.8\textrm{GHz}$ range and is progressively built by the robots that can communicate with each other and extend the network by dropping additional WiFi nodes. This section describes the hardware and software configuration that enables remote networking of the agents in the underground. Figure~\ref{fig:network_diagram} shows an overview of the communication architecture used by team CERBERUS.
\subsection{Mobile Agent Nodes}
Agents employ two different hardware methods. Ground robots are equipped with a commercial off-the-shelf device (ACKSYS Communications \& Systems - EmbedAir1000) that support\textcolor{revision}{s} WiFi access point, client, repeater, and MESH point modes. In mesh mode, these devices create a mesh network, that is a local network topology in which the infrastructure nodes connect directly, dynamically and non-hierarchically to as many other nodes as possible and cooperate with one another to efficiently route data from/to clients. In this way, ground robots establish a backbone wireless network shared among the other agents.

Due to their limited payload capability and size constraints, the aerial robots did not carry mesh node-capable devices and acted solely as clients. They connect to the closest existing wireless access point, broadcast either by the Base Station or by nodes deployed by ground robots.

\subsection{Base Station Network}
In preparation for a wide variety of competition scenarios, the team prepared four solutions to provide communications from the Base Station into the subterranean environment. First, the Base Station uses an ACKSYS EmbedAir1000 WiFi radio and connects to the mesh network using a high gain, directional panel antenna from L-Comm; supplying a high bandwidth connection both within the Staging Area as well as deep into the competition area. Second, Armadillo was deployed with a $300\si{\metre}$\textcolor{revision}{-long} optical fiber cable, a panel antenna, as well as a multi-modal perception system that further allowed it to be remotely controlled by the Human Supervisor while extending the WiFi network around corners and down corridors with ease in relative proximity to the Base Station. Third, a panel antenna was attached to a PVC pole and angled to provide a network down a stairway very near the competition gate. The pole was manually extended and oriented down the stairway by a team member. Finally, an EmbedAir1000 WiFi radio and its battery were placed in a foam ball which was manually thrown by a team member down a stairway to provide a network connection to the lower level of the circuit. The Gagarin Aerial Scout utilized both the ball and pole antennas while exploring the staircase.

\subsection{Deployable Mesh Nodes}
\label{sec:mesh_node}
ANYmal robots were able to carry additional WiFi nodes and deploy them on the ground to expand the reach of the wireless mesh network incrementally. Each robot was able to store four modules inside its belly plate and release them by lowering the torso height and activating the payload release mechanism described in Section ~\ref{sec:anymal_b_subt}. Each node is a water and dustproof self contained unit, consisting of a battery pack (allowing approximately two hours of operation), a battery management system, a magnetic switch (to switch the module on after releasing it), an LED to aid the recovery of the module after the mission, two WiFi patch antennas, and an EmbedAir1000 module (see Figure~\ref{fig:wifi_beacon}). \textcolor{revision}{In as ideal as possible underground conditions, such as a straight tunnel, the modules can reliably communicate within a range of \SI{100}{\meter} when dropped on the ground. However, due to obstructed lines of sight from solid material (e.g., in a mine corner), multipath and reflections in underground settings, performance may vary considerably. As evaluated via experimental observations, close to the first communication module a maximum available bandwidth of approximately $200\textrm{Mbps}$ was observed. In team deployments, the legged robots and the droppable communication nodes had both radios of the EmbedAir1000 module enabled to maximize the communication performance and maintain as much of the initial bandwidth as possible across the deployed communication topology. }

The Human Supervisor was left in charge of deciding the deployment location, as wireless signal propagation is often unpredictable in subterranean scenarios. \cite{rizzo2013signal} show that a deviation in robot trajectory of a few centimeters can cause a significant difference in terms of received power on its onboard mesh receiver. In a more recent analysis, \cite{tardioli2019ground} show that due to destructive self-interference of signals reflecting off tunnel walls, there is a highly varying fading effect which leads to \ac{RSSI} having a variance of up to $30-40$\si{\percent}. As a result, \ac{RSSI} can be a misleading indicator to check the quality of a wireless connection inside a tunnel. Moreover, this variance can result in a link between two nearby mesh nodes reporting a worse signal strength than between nodes further apart. These peculiar characteristics, therefore, make an automatic decision of where to deploy mesh nodes a non-trivial problem. As a result, the deployment location was decided by the Human Supervisor. This decision was based on information about signal quality (such as the output of \textit{ping} utility) between the Base Station and the agent, as well as considerations about the geometry of the environment surrounding the robot.

\subsection{Nimbro - Multi-Master Network} \label{sec:nimbro_network}
The \ac{ROS} was used as middleware for message passing to enable communication between processes running on each robot, as well as across robots on the same network. However, \ac{ROS} requires a \textit{rosmaster} process which monitors all communication connections. There can only be one \textit{rosmaster} on a given network and typically, this process is run onboard a robot's computer \textcolor{revision}{in case} there is communication loss with the Base Station. This presents a challenge in scenarios with multiple robots, each with their own \textit{rosmaster}, communicating on the same network to the same Base Station.

The Nimbro Network\footnote{\url{https://github.com/AIS-Bonn/nimbro_network}} software package allows the generic transport of \ac{ROS} topics and services over unreliable network connections and has been deployed in several robotic challenges~\cite{stuckler2016nimbro,schwarz2017nimbro}. Since it avoids any configuration/discovery handshake, it is suited for situations like the SubT Challenge where wireless connections can drop and recover unexpectedly. 
Its primary purposes are to decouple multiple \textit{rosmasters} and select which data are transmitted between agents and the Base Station. %
A hardcoded list of data to be exchanged may be cumbersome to maintain and adapt, especially when tools such as \textit{multimaster\_fkie}~\cite{Tiderko2016} offer the possibility to automatically establish and manage a multi-master network with little or no configuration. However, these automatic tools require extra customization to be adapted to the SubT Challenge, as done by team CoSTAR~\cite{queralta2020collaborative} and may be error-prone, transmitting unnecessary data and using limited bandwidth. Therefore, given that an ANYmal robot had more than $800$ topics and $400$ services in its \ac{ROS} system, the alternative to clearly identify which data streams should be transmitted and not rely on any automatic decision was preferred. 
A key factor in a system with multiple \textit{rosmasters} is time synchronization between the agents and the Base Station. This was achieved through the usage of the software tool \textit{chrony}\footnote{\url{https://chrony.tuxfamily.org/}}. \textit{chrony} is an implementation of the \ac{NTP} that can synchronize the system clock with a predefined server. In this use case scenario, each robot acted as a client of the Base Station, which was the main \ac{NTP} server.

%% file: 11_operator_interface.tex
A crucial aspect of the SubT Challenge is the single Human Supervisor at the Base Station. This person is the only individual allowed to coordinate, manage, and monitor the deployed systems. The single operator requirement is intended to minimize the amount of human intervention for this type of autonomous mission and forces the robots to have a high level of autonomy\textcolor{revision}{,} while remotely mapping and/or navigating complex environments. %

\begin{figure}[!h]
    \begin{subfigure}{.3\textwidth}
      \centering
      \includegraphics[scale=0.15]{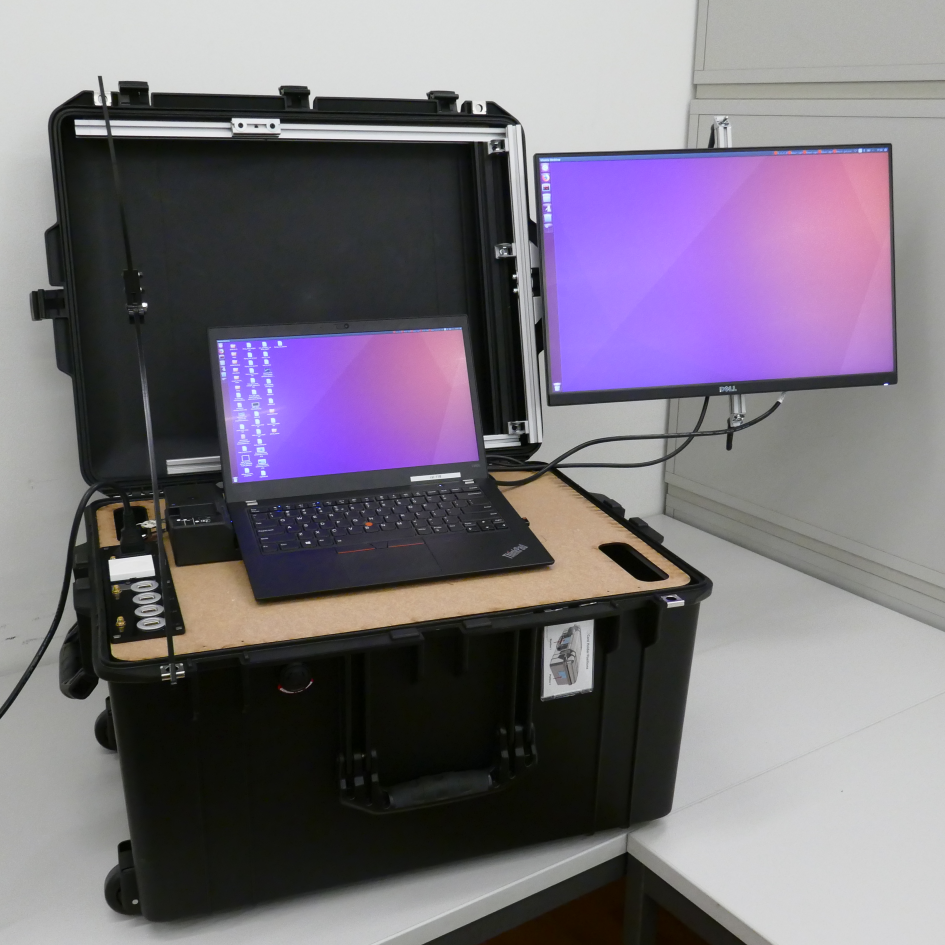}
      \caption{Base Station.}
      \label{fig:opc_base_station_closeup}
    \end{subfigure}%
    \begin{subfigure}{.3\textwidth}
      \centering
      \includegraphics[scale=0.15]{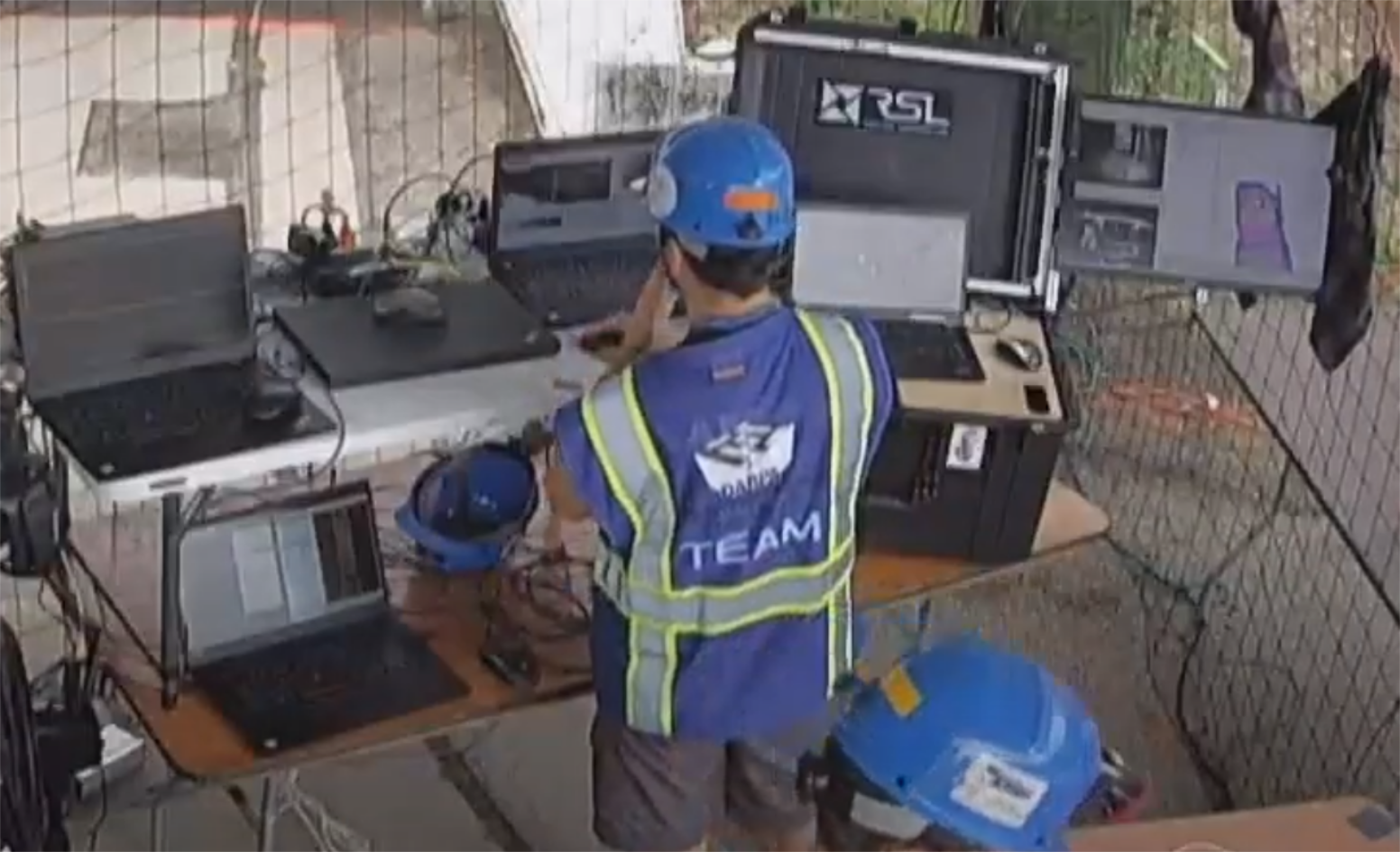}
      \caption{Tunnel Circuit setup.}
      \label{fig:opc_base_station_tunnel}
    \end{subfigure}
    \begin{subfigure}{.3\textwidth}
      \centering
      \includegraphics[scale=0.26]{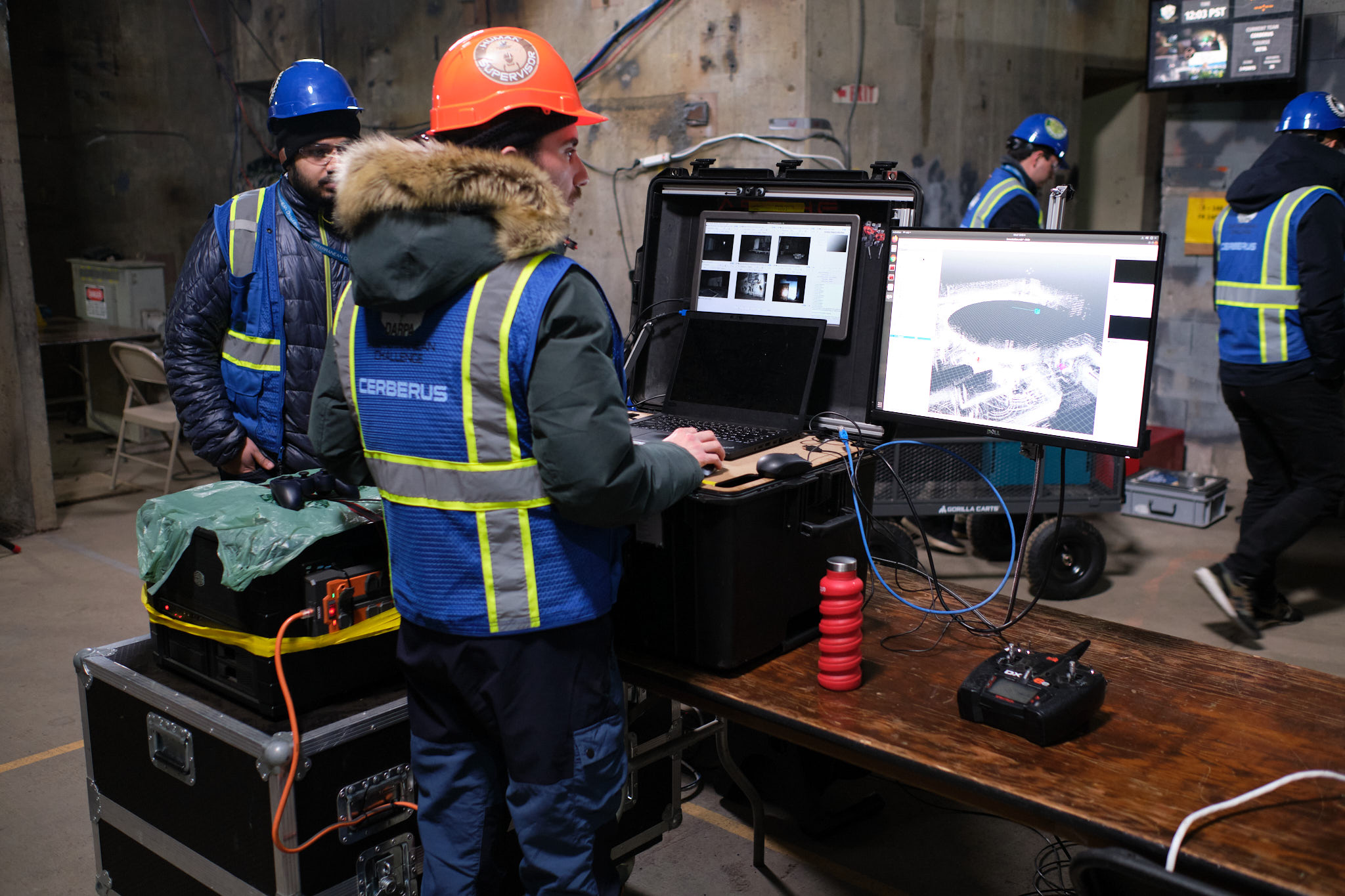}
      \caption{Urban Circuit setup.}
      \label{fig:opc_base_station_urban}
    \end{subfigure}
    \caption{Setup of the Base Station for the Human Supervisor.  Operator's box with external screen and docking station (\subref{fig:opc_base_station_closeup}). Setup for Tunnel Circuit: one PC for aerial robots, two PCs to control two ANYmal robots, one PC for \textcolor{revision}{the} Armadillo rover, and the main \ac{ARI} running on the screen of another PC (\subref{fig:opc_base_station_tunnel}). Setup for Urban Circuit: one PC for aerial robots, one PC to control all ground robots,  and the main \ac{ARI} running on the screen of another PC (\subref{fig:opc_base_station_urban}).}
    \label{fig:opc_base_station}
\end{figure}

\begin{figure}[!h]
    \begin{subfigure}{.32\textwidth}
      \centering
      \includegraphics[scale=0.17]{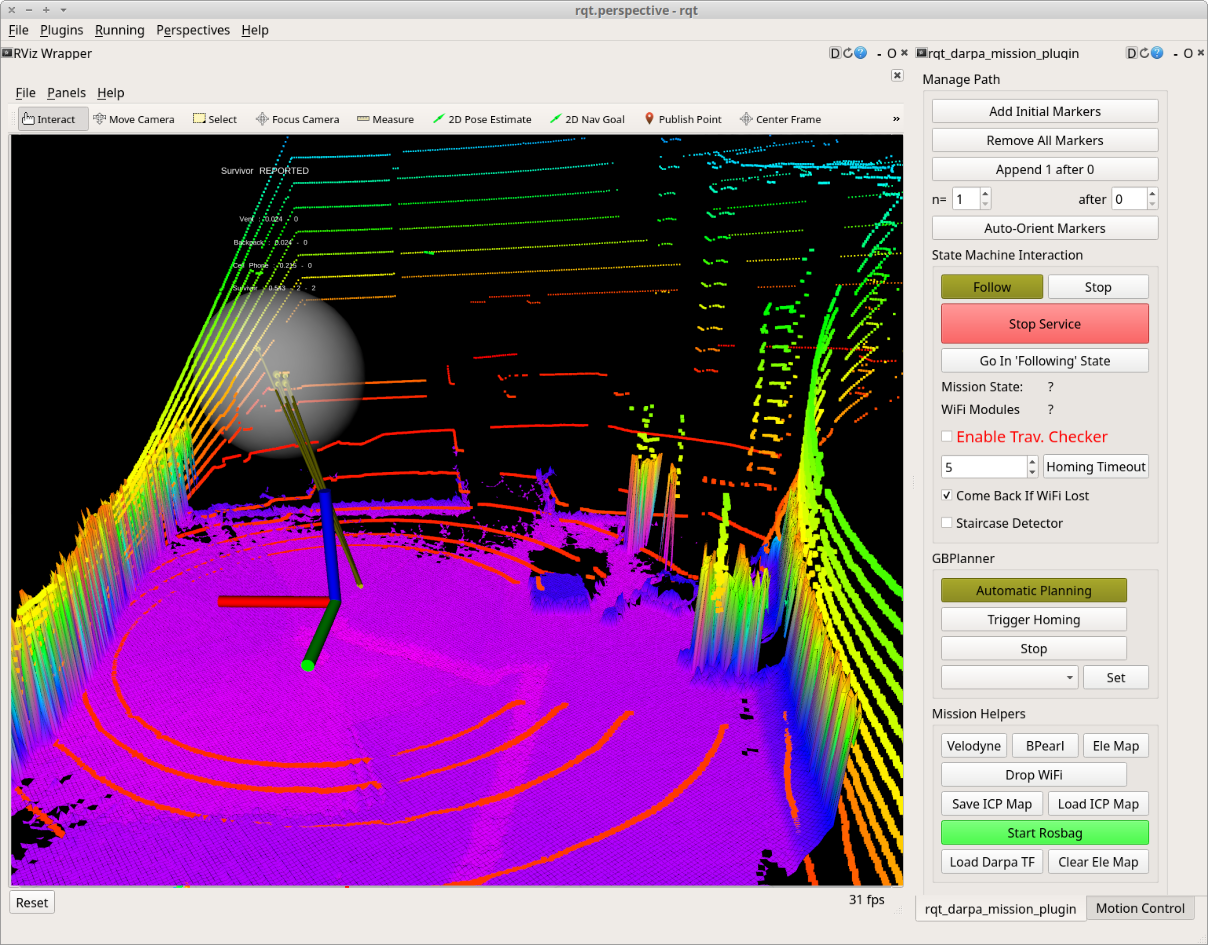}
      \caption{ANYmal \ac{MCI}.}
      \label{fig:mci_anymal}
    \end{subfigure}%
    \begin{subfigure}{.32\textwidth}
      \centering
      \includegraphics[scale=0.17]{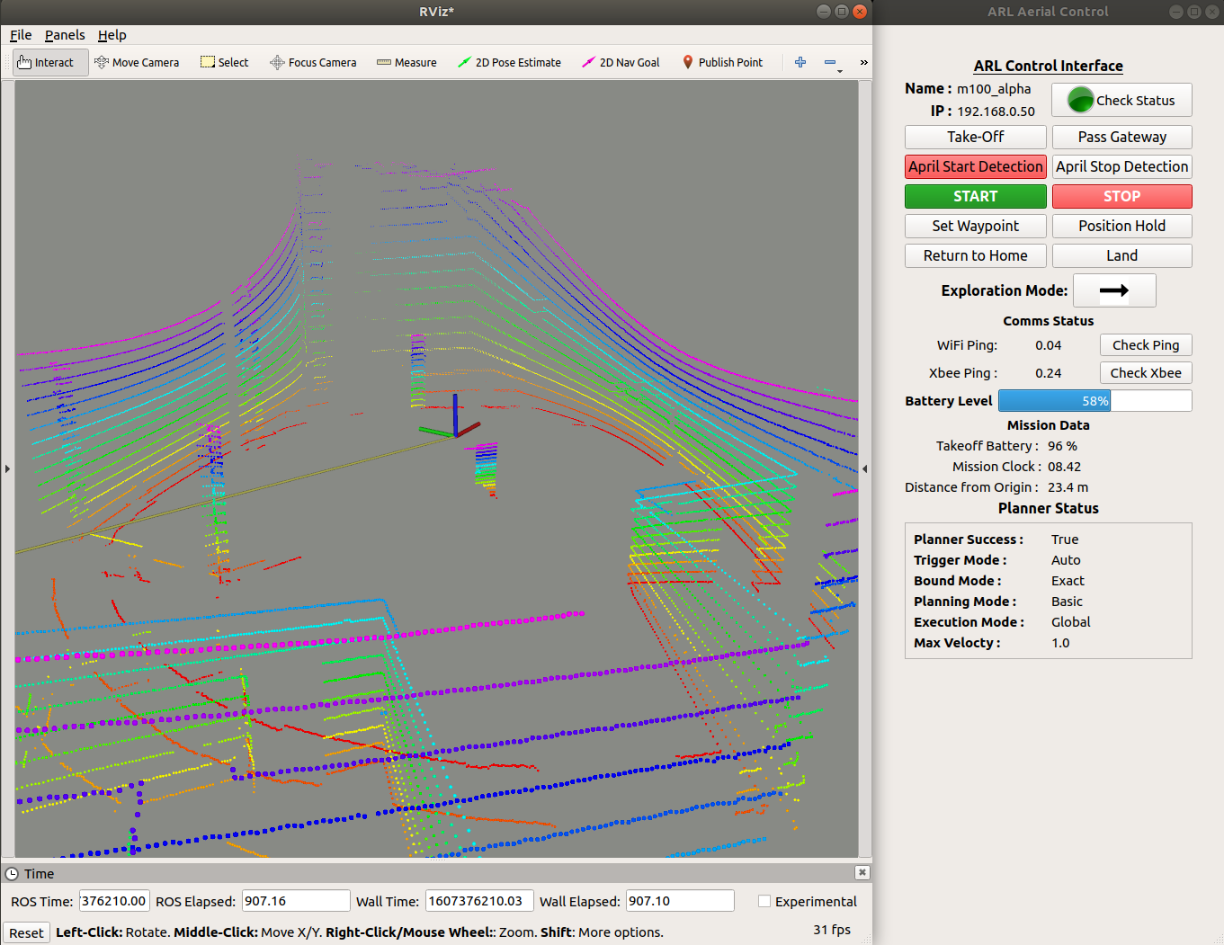}
      \caption{Aerial Scout \ac{MCI}.}
      \label{fig:mci_aerial}
    \end{subfigure}
    \begin{subfigure}{.32\textwidth}
      \centering
      \includegraphics[scale=0.17]{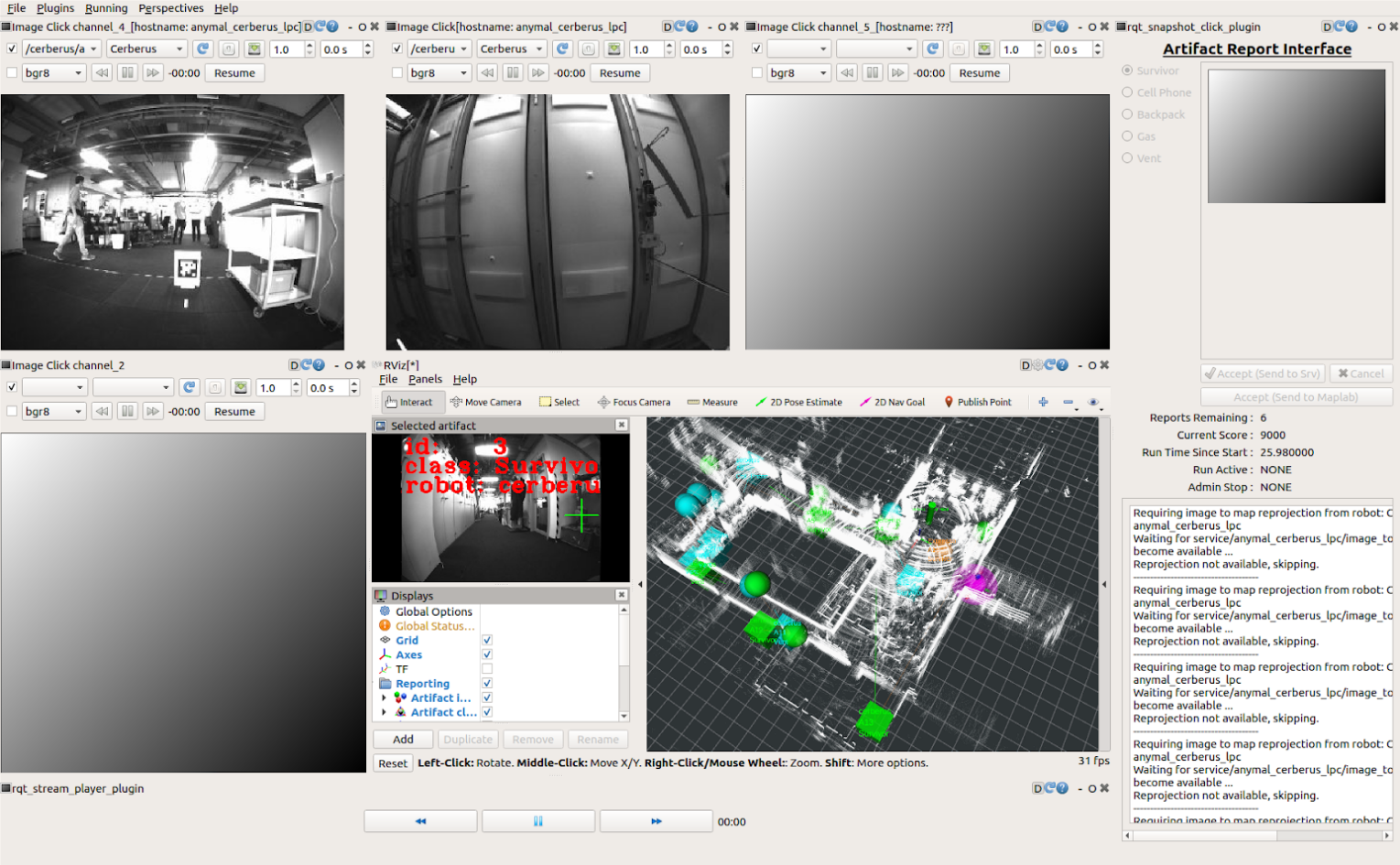}
    \caption{\ac{ARI}}
      \label{fig:ari}
    \end{subfigure}
    \caption{Mission Control Interfaces (MCI) for ANYmal (\subref{fig:mci_anymal}) and aerial scouts (\subref{fig:mci_aerial}). They have a similar layout but feature some customization based on the specific type of robot. The minimal amount of data displayed to the Human Supervisor reflects the need to reduce streamed data from each robot to the essential ones: a frame representing the robot's pose in the map, requested LiDAR scans and/or elevation map snapshot, information about automatically detected artifacts and their location. The Artifact Reporting Interface (ARI) combines video streams from all robots, a panel to accept and modify artifact detections, and an overview of the reported global dense map overlaid with the state of all detected artifacts (\subref{fig:ari}).}
    \label{fig:supervision_guis}
\end{figure}

During the Tunnel and Urban Circuits, the Human Supervisor used a custom and integrated solution for the Base Station setup (see Figure~\ref{fig:opc_base_station}). This solution was composed of a sturdy box (Peli Air Case - Pelican) containing electronic components (WiFi router, auxiliary screen, etc.). It ensured fast deployment and setup of the Base Station. The supervisor used two main software modules to interact with the deployed robots: a \ac{MCI}, and an \ac{ARI}, exposed through a \ac{GUI} (see Figure~\ref{fig:supervision_guis}).  The \ac{MCI} was used to (a) check the status of each agent, (b) give high-level commands to override the current robot's behavior and (c) request specific sensor data to evaluate the current situation better.
The \ac{ARI} was used by the Human Supervisor to (a) have a top-down view of the global map created during the mission, (b) visualize throttled and compressed image streams from the agents, to manually report artifacts, and to (c) check, and approve or discard artifacts that were automatically detected by the robots.

Team CERBERUS adapted its approach towards the single Human Supervisor requirement between the Tunnel and Urban Circuits, improving the \ac{GUI}s to control multiple agents, as well as the amount of exchanged data between robots and the Base Station. Major changes to the communication architecture were introduced. These are highlighted in the following sections.

\subsection{Robot Control} \label{sec:robot_control}
During the Tunnel Circuit, each robot received commands from the Human Supervisor via a dedicated \ac{OPC}. An \ac{OPC} was specifically set up to interact with one or more robots. It used the \textit{rosmaster} of the corresponding robot and was used to initialize the agent (launching procedure, detection of the DARPA gate, etc.). Every ground robot operated in conjunction with its own \ac{OPC}, and could stream all requested topics to the supervisor. On the other hand, all the aerial scouts could be controlled - one at a time - by another \ac{OPC}. Each of these \ac{OPC}s run its own instance of \ac{MCI}, customized based on the type of robot. Only Aerial Scouts took advantage of the multi-master architecture with the Nimbro Network software package (described in Section ~\ref{sec:nimbro_network}), whereas ground robots used it only to report artifacts and map updates. Once each robot was set up by the Pit Crew and ready to start the mission, its \ac{OPC} was passed to the Human Supervisor that could then use it to deploy the robot. The described setup resulted in an extra burden for the single Human Supervisor who had to deal with multiple PCs to execute the mission. For example, to control two ANYmals, one wheeled ground robot and aerial \textcolor{revision}{robots}, the supervisor had to deal with four different \ac{OPC}s, plus the additional PC running \ac{ARI} - see example setup in Figure~\ref{fig:opc_base_station_tunnel}. This approach was not sustainable, and therefore specific modifications were applied for the Urban Circuit.

A different setup of the Base Station and for controlling the robots was employed during the Urban Circuit. Every robot leveraged the multi-master architecture with the Nimbro Network package and exported only a subset of the \ac{ROS} topics and services to its \ac{MCI}. This prevented the wireless network from being cluttered with data that was not strictly needed for the mission execution. Moreover, the concept of ``on-demand'' topics was introduced: sensor data such as LiDAR scans and elevation maps require a considerable amount of bandwidth to be streamed over WiFi, and are most of the time not necessary to be continuously sent to the Human Supervisor. Therefore a single message of these data streams was sent only upon request of the supervisor via the \ac{MCI}.
Each robot still had its dedicated \ac{OPC}, but it was used solely for agent setup by the Pit Crew prior to the deployment. The overall startup procedure included one PC to run the instances of \ac{MCI}s for all ground robots, and one other for the \ac{MCI} for aerial robots. This meant that to control two ANYmals, the Armadillo rover and the aerial scouts, the Human Supervisor had to deal with two computers, located at the Base Station (see Figure~\ref{fig:opc_base_station_urban}).

\textcolor{revision}{In both Circuits, the Human Supervisor was able to supervise the decisions taken by each agent and, if required, decide to re-define the exploration bounding box in which each robot was allowed to explore. In every assigned area as defined by the planner bounding box, each robot was exploring the unknown space autonomously without relying on external information. If necessary, the Human Supervisor was able to overrule the autonomous planner and provide way-points to any robot.}

\subsection{Artifact Reporting}\label{sec:reporting_interface}
The artifact reporting interface (see Figure \ref{fig:ari}) is designed to reduce the strain on the Human Supervisor while improving situational awareness and reducing the chance of reporting mistakes. It is composed of panels showing video streams from the selected robots, a panel notifying the supervisor of all new automatic artifact detections, an RViz window showing all detected artifacts in the global dense map, and a text window for important system errors. The video streams can be paused and rewound, allowing the supervisor to better understand the areas explored by the robots. In case some artifacts were not detected automatically by the robots, a bounding box can be drawn on the image to attach the artifact to the map as discussed in section \ref{sec:artifact_map_attachment}. When the Human Supervisor is notified of a new automatic artifact detection, the associated camera image, annotated with the artifact's bounding box and class, is shown for context. Under normal operation, the supervisor would accept the artifact, which would then be added to the multi-robot global map. As a fallback, the supervisor is also able to bypass the automatic artifact type detection and mapping server and file artifact reports directly to the DARPA Command Post.%

Once an artifact has been added to the multi-robot global map, it appears in an RViz panel together with the optimized global dense map. For additional context, the artifact's original position, optimized position, and the robot pose from which it was observed are shown. Interactive markers
are used to allow the Human Supervisor to report, modify, and delete artifacts through a right-click menu. Whenever an artifact is selected, its associated annotated camera image is shown in a dock next to the map. This proved very helpful for the supervisor to recall and distinguish artifacts. Once an artifact is reported, its color changes from cyan (not yet reported) to green (valid) or red (invalid) depending on the response from the DARPA Command Post. As the robots explore the environment further, the global map improves due to additional measurements, including loop closures. When significant improvements occur, the Human Supervisor can submit new reports for artifacts that were initially rejected.

\subsection{Fallback Solutions}
\label{subsec:fallback}
Given the centralized nature of the reporting architecture, special care is taken to minimize the damage to the team's competition score in case of system malfunctions on the Human Supervisor's PC at the Base Station. Starting at the hardware level, a custom-built server is used as the Operator PC. To provide robustness to temporary power losses, it is powered through an Uninterruptible Power Supply. At the software level, all packages are designed such that they can be started independently and at any point in time. The packages that cannot be implemented in a stateless fashion continuously save backups of their internal state to persistent storage. As an example, all the artifact positions received at the Base Station are periodically saved to disk. All packages can, therefore, safely be relaunched at any time, either by the Human Supervisor or automatically if crashes are detected.

The primary reliability concept for the multi-robot global map is to combine measurements from complementary sensor modalities and to perform robust outlier rejection using switchable constraints (see section \ref{sec:multi_agent_mapping}). To account for cases where areas in the global map appear incorrect, or the global map is delayed, a bypass is implemented for artifact and global map reporting. As discussed in section \ref{sec:artifact_map_attachment}, the optimized position for an artifact $\textbf{a}_j$ expressed in DARPA frame $\mathbb{D}$ is computed as $^{\mathbb{D}}\mathbf{a}_j=T_{\mathbb{D}\mathbb{B}_{t_i},opt}\ ^{\mathbb{B}_{t_i}}\mathbf{a}_j$, where $T_{\mathbb{D}\mathbb{B}_{t_i},opt}$ corresponds to the optimized pose graph node of the robot that detected the artifact at timestamp $t_i$. In addition to $T_{\mathbb{D}\mathbb{B}_{t_i},opt}$, the pose $T_{\mathbb{D}\mathbb{B}_{t_i},unopt}$ given by the robot's own odometry at time $t_i$ is also stored. Using the artifact reporting GUI, the Human Supervisor can report alternative, unoptimized artifact poses computed with $^{\mathbb{D}}\mathbf{a}_j=T_{\mathbb{D}\mathbb{B}_{t_i},unopt}\ ^{\mathbb{B}_{t_i}}\mathbf{a}_j$. In a similar fashion, the global pointcloud map is typically computed by transforming the robot's pointclouds into the DARPA frame based on the robot's optimized poses $T_{\mathbb{D}\mathbb{B}_{t_i},opt}$. If the Human Supervisor flags the global map as unreliable, or if the time since the optimized map's last update exceeds a threshold, the map reporting system temporarily falls back to the unoptimized global pointcloud map computed by transforming the pointclouds into the DARPA frame using the unoptimized poses $T_{\mathbb{D}\mathbb{B}_{t_i},unopt}$.

%% file: 12_experimental_evaluation.tex
This section focuses on selected results from the deployment of team CERBERUS in the DARPA Subterranean Challenge Tunnel and Urban Circuit competition events. 

\subsection{Field Tests prior to the Circuit events}
To verify and evaluate the performance of the CERBERUS walking and flying robots system-of-systems and prepare for the participation of our team in the DARPA Subterranean Challenge Circuit events, a set of field tests took place. Reflecting the geographical organization of our team, field evaluation took place both in the USA and in Switzerland which in turn allowed us to explore different types of underground mines, subterranean urban infrastructure, and lately cave environments. 

\begin{figure}[!h]
    \begin{subfigure}{\textwidth}
      \centering
      \includegraphics[width=0.65\textwidth]{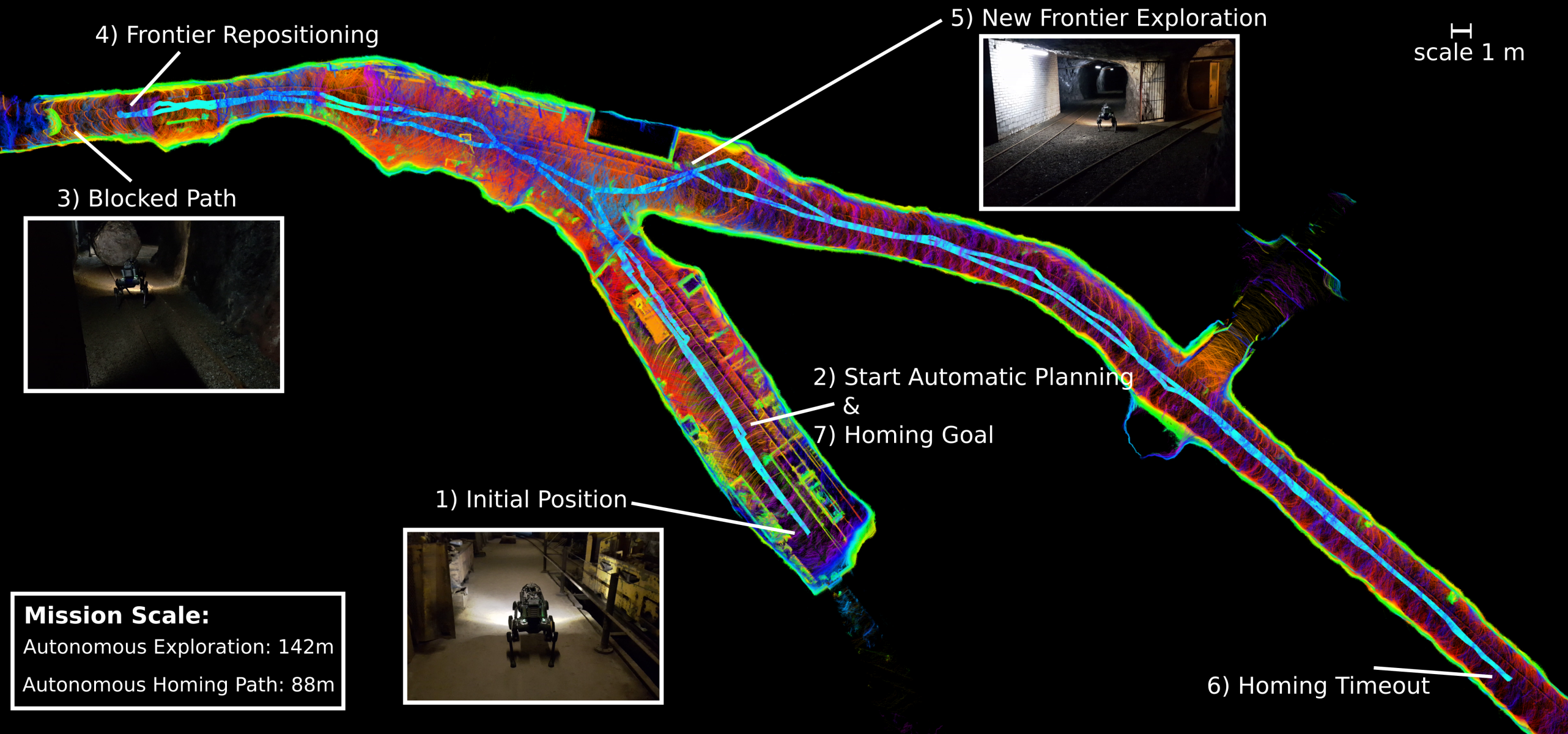}
      \caption{ANYmal B deployed in the Gonzen underground mine in Switzerland.}
      \label{fig:anymal_b_gonzen}
    \end{subfigure}%

    \begin{subfigure}{\textwidth}
      \centering
      \includegraphics[width=0.65\textwidth]{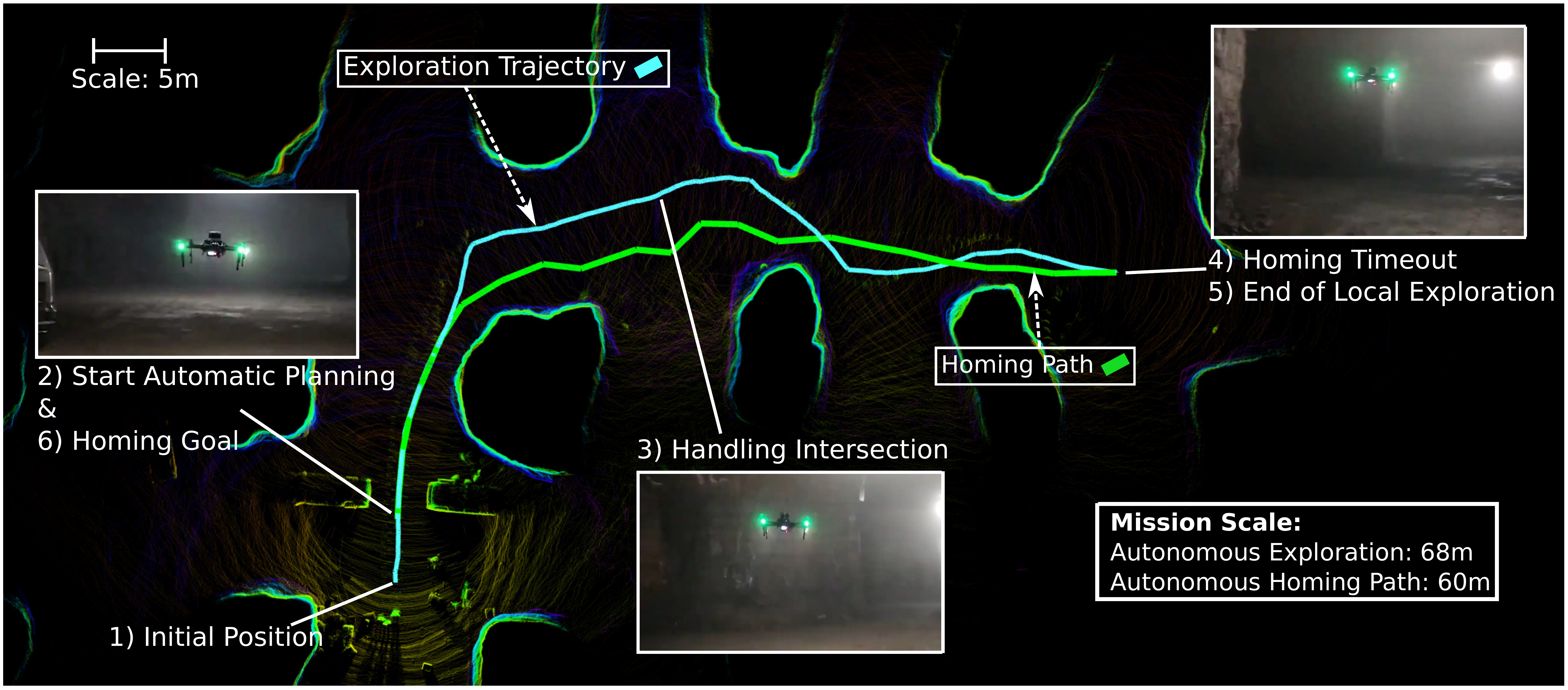}
      \caption{Alpha Aerial Scout deployed in Wampum mine. }
      \label{fig:flying_wampum}
    \end{subfigure}
    \caption{\textcolor{revision}{Indicative field} deployment\textcolor{revision}{s} of team CERBERUS' robots prior to official mission events. The shown missions were executed autonomously by the agents.}
    \label{fig:field_preparation}
\end{figure}

In further detail, our field testing activities involved the following main sites: a) the TRJV underground mine (NV, USA), b) the Comstock underground mine (NV, USA), c) the Wampum underground room-and-pillar mine (PA, USA), d) several buildings of the University of Nevada, Reno (NV, USA), f) a railroad tunnel and a city tunnel (NV, USA), g) the Gonzen underground mine (Sargans, Switzerland), h) the Menznau Bunker (Luzern, Switzerland), i) the ARCHE test facilities (Wangen an der Aare, Switzerland) and other settings.
More recently, we have tested in the Moaning Caverns (CA, USA), the Subway Cave (CA, USA), and other cave environments. Among those, only a subset have been utilized for testing with the entire team -- both in terms of members of each of our groups and of our robots -- as this required special travel organization. This included the Gonzen mine, the Wampum mine (shown in Figure~\ref{fig:field_preparation}), the Menznau Bunker, as well as the facilities of UNR. Another important testing event was the \ac{STIX}, which took place in April $2019$ at the Edgar Experimental Mine (CO, USA). STIX was organized by DARPA so that qualified teams could test in a representative environment, with the same rules as the actual runs.

During such team-wide integration activities, the main effort has been on the integration and mission-like deployment of our system-of-systems solution involving robot configurations with significant differences, all being largely ongoing research prototypes instead of acquired and rather more mature products. This in turn also implied certain challenges relating to integration largely experienced in the Tunnel Circuit. To enable smoother integration and interoperability of our robotics team, we have since emphasized on unifying components as much as possible, such as sensing, perception algorithms, planning strategies, interfaces, and more. This approach paid off significantly in our Urban Circuit deployment and has since been improved continuously.
\subsection{DARPA Subterranean Challenge -- Tunnel Circuit}
\label{sec:tunnel_results}
The Tunnel Circuit took place in August $2019$ at the National Institute for Occupational Safety and Health (NIOSH) Mine in Pittsburgh, Pennsylvania, USA. This facility has the characteristics of a ``room-and-pillar'' mine, with horizontal arrays of rooms where pillars are composed of untouched material, left to support the roof. Teams were scored on two different courses: the Safety Research Course (SR) and the Experimental Course (EC). %

Team CERBERUS deployed a total of six different types of robots across the four runs, correctly reporting six artifacts and scoring five points in total. CERBERUS finished sixth out of eleven teams in the overall ranking. The deployed robots were: two ANYmal B robots (called Bear and Badger), Armadillo rover, Alpha Aerial Scout, and Gagarin Aerial Scout. \textcolor{revision}{The ANYmal B robots, and the Alpha and Gagarin Aerial Scouts operated autonomously running the exploration planner, for which the local map subset $\mathbb{M}_L$ was set as a cuboid of dimensions $30\times30\times3.0\textrm{m}, 16\times16\times1.5\textrm{m} ~\textrm{and}~ 20\times20\times1.0\textrm{m}$ for the ANYmals, Alpha and Gagarin respectively.} The two ANYmal robots were deployed in every run, featuring different feet configurations (point feet, flat feet, wheels) but could score only one point. Armadillo scored the other four points during the last run; after both ANYmals could not proceed further and the rover was driven remotely by the Human Supervisor using a joystick. This was possible by viewing the live camera image streams, taking advantage of the extended connection provided by the fiber optic cable.

\begin{figure}[!h]
    \centering
    \includegraphics[width=0.99\textwidth]{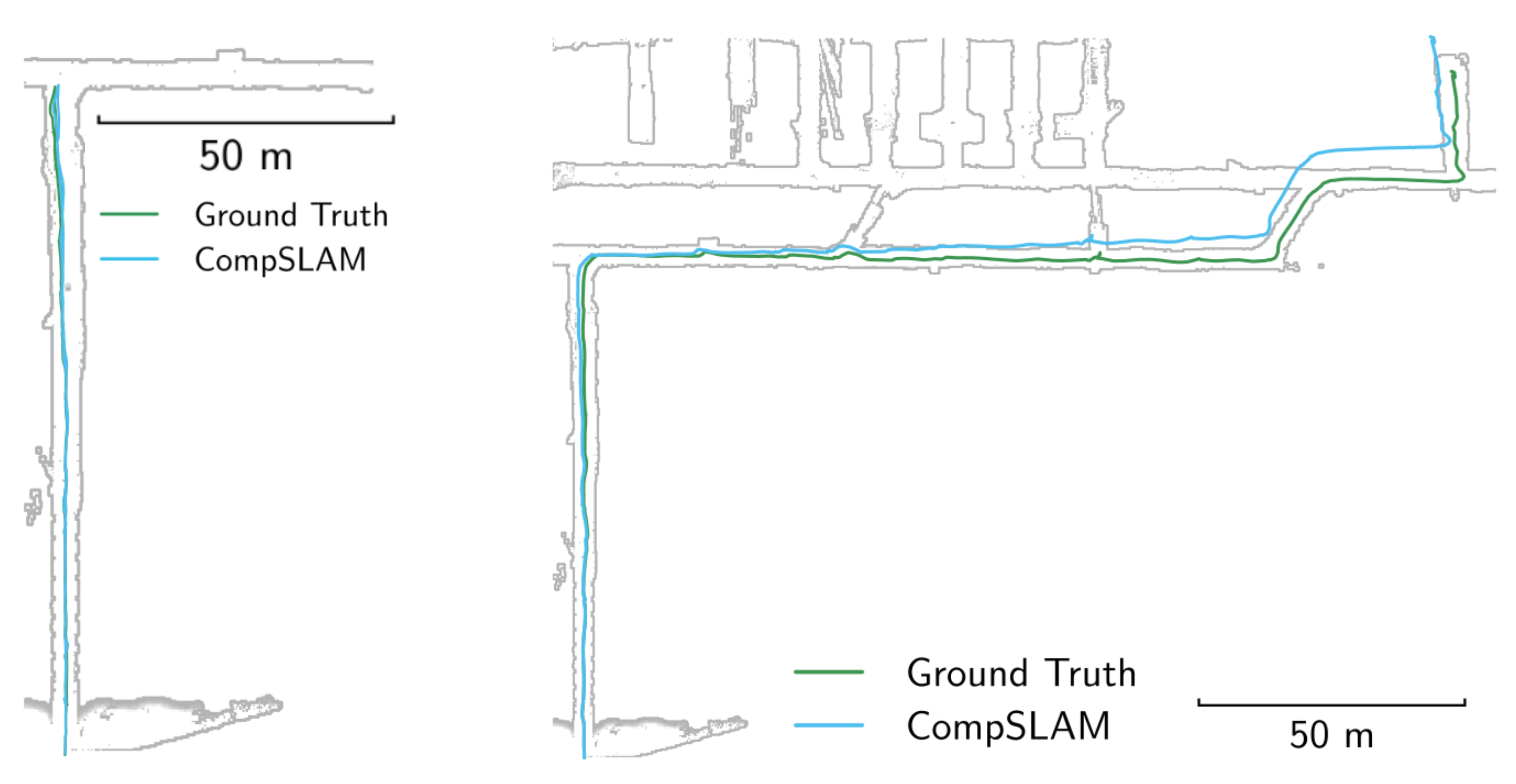}
    \caption{Comparison of onboard (CompSLAM) estimated trajectory (blue) with ground-truth trajectory (green) for the Alpha Aerial Scout (left) and the Armadillo rover (right) during competition runs of the DARPA SubT Tunnel Circuit. The aerial robot utilized visual, depth, and inertial data for its odometry estimation compared to the rover, which relied purely on depth data, hence producing better trajectory estimates when compared against ground-truth. This trajectory comparison demonstrates the importance of utilizing multi-modal sensor data for robot navigation in challenging underground environments. A video of the Aerial Scout's mission can be found at~\url{https://youtu.be/mw0qy05Fo6Q}.}
    \label{fig:alpha_smb_tunnel}
\end{figure}

To report artifacts during the Tunnel Circuit, robots relied on their onboard multi-modal odometry estimation and their initial pose with respect to \textcolor{revision}{the} DARPA global frame $\mathbb{D}$ (Figure \ref{fig:mapping_overview}). To provide a quantitative evaluation of the robot localization, ground-truth trajectories were later created by registering recorded robot LiDAR scans to the provided DARPA ground-truth maps~\cite{ramezani2020newer} using the \ac{ICP} algorithm. Table~\ref{tab:rpe_tunnel} presents the Relative Pose Errors (RPE), calculated every \SI{10}{\meter}, and Absolute Pose Errors (APE) for the Alpha Aerial Scout and the Armadillo rover for the Tunnel Circuit deployments, with Figure~\ref{fig:alpha_smb_tunnel} presenting a qualitative comparison of each robot's trajectory against the ground-truth trajectory. During the Tunnel Circuit, the Alpha Aerial Scout utilized visual, depth, and inertial data for its odometry estimation which allowed it to autonomously explore and safely returned to its take-off position while traversing a total path of ~\SI{180}{\meter}. As shown on the left in Figure~\ref{fig:alpha_smb_tunnel}, the estimated robot trajectory closely follows the ground-truth trajectory. During this deployment, the Alpha Aerial Scout would have continued its autonomous exploration mission (and would have explored the opposite direction as the Armadillo), but the exploration planner's bounding box, which defines the total area of interest for the planner, was incorrectly set by the Human Supervisor at the Base Station. The Armadillo was manually moved forward during the deployment and traversed a total path of~\SI{~285}{\meter}
before being stopped due to the optical fiber cable wrapping around its wheels.  Armadillo was initially planned to provide only extended communication during the deployment; however, after initial competition deployments, the team re-purposed the robot to detect and report artifacts by modifying the onboard sensory payload and performing complete calibrations during the break between runs. This in-field modification proved useful, and the rover successfully reported four artifacts during its deployment. However, due to technical issues, the rover only utilized LiDAR data for its odometry estimation without employing any additional sensing data to improve its estimates. As noted in Table~\ref{tab:rpe_tunnel}, even though the RPE remains low, the lack of additional sensing to improve the odometry led to an accumulation of errors, especially during rapid in-spot rotations, causing the robot trajectory to drift from the ground-truth, as seen in Figure~\ref{fig:alpha_smb_tunnel}. Nevertheless, the APE at the end of the robot's run still remained with-in the DARPA's margin of error of $\pm$~\SI{5}{\meter} for artifact reporting.

\begin{table}
\centering
\begin{tabular}{c|c|c}  
\toprule
\multicolumn{3}{c}{\textbf{DARPA Tunnel Circuit - Robot Localization Results}} \\
\midrule
\textbf{Robot / Run} & \textbf{Relative Pose Error} & \textbf{Absolute Pose Error}\\
\midrule
Alpha Aerial Scout / SR2&  0.11\si{\metre} / 0.86\si{\degree}  & 0.48\si{\metre} / 1.05\si{\degree} \\
Armadillo Rover / SR2 &  0.03\si{\metre} / 0.37\si{\degree}  & 3.39\si{\metre} / 2.87\si{\degree}\\ 
\bottomrule
\end{tabular}
\caption{Results of the onboard CompSLAM approach deployed on the Alpha Aerial Scout and Armadillo rover robots during the DARPA Tunnel Circuit.}
\label{tab:rpe_tunnel}
\end{table}

\subsection{DARPA Subterranean Challenge -- Urban Circuit}
The Urban Circuit took place in February $2020$ at the Satsop Business Park in Elma (WA, USA). The Satsop Park is an unfinished nuclear power plant, featuring an underground labyrinth to be explored, with multiple levels, narrow passages, and different obstacles to traverse. 
Teams were scored on two different courses: the Alpha Course and the Beta Course. 
Team CERBERUS deployed a total of five robots across the four sessions, correctly reported eleven artifacts and scored seven points in total, achieving fifth placement (out of ten teams) in the overall ranking. The deployed robots were: two ANYmal B robots (Bear and Badger) with point feet, the Armadillo rover, the Alpha Aerial Scout, and the Gagarin Aerial Scout. Another roving robot for networking support was deployed but was used minimally during the sessions.

\begin{figure}[!h]
    \centering
     \includegraphics[width=0.9\textwidth]{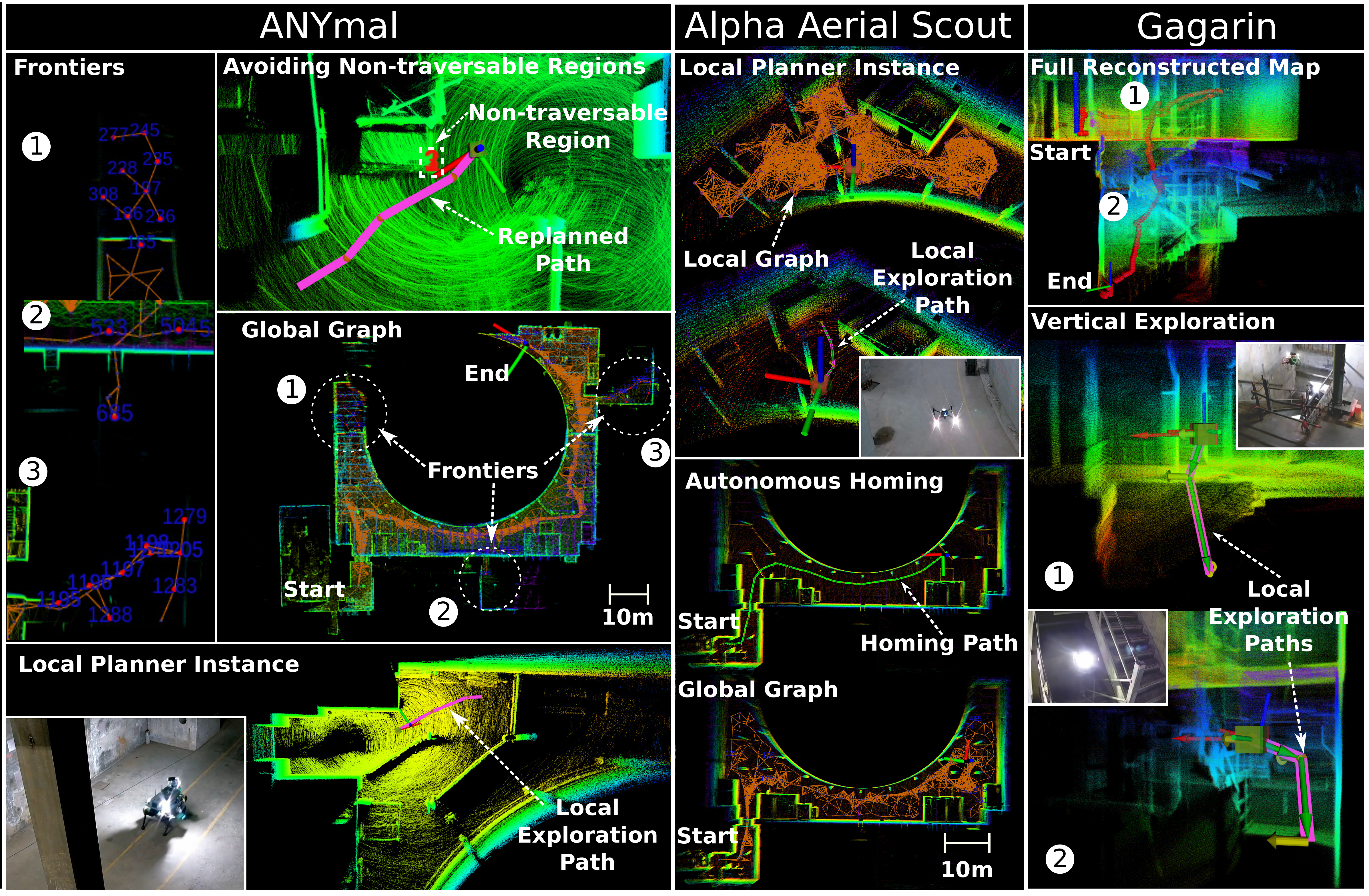}
    \caption{Details of deployment of our exploration planner in the Alpha course of the Urban Circuit. The ANYmal B robot, Alpha Aerial Scout and Gagarin Aerial Scout were deployed running the exploration planner. All robots were deployed at the entrance of the environment and after manual gate passing, the local planner was triggered. The Gagarin was commanded to perform vertical exploration of the staircase near the entrance, whereas the other robots performed horizontal exploration. The Alpha Aerial Scout explored the bottom section of the \textcolor{revision}{first floor of the} environment and triggered auto-homing after reaching the battery limit. \textcolor{revision}{The} ANYmal robot explored the area around the main reactor \textcolor{revision}{of the first floor}. When \textcolor{revision}{the robot encountered non-traversable areas}, the planner stopped, marked these areas as geofence zones and re-triggered the local planner. \textcolor{revision}{The planned paths, explored map and the global graph are shown for the Alpha Aerial Scout and ANYmal along with the frontier vertices detected by the planner in ANYmal's mission. The sub-figure for the Gagarin Aerial Scout shows the robot's path going down the staircase along with two local planner instances.} Videos of Alpha Course deployments can be found at \url{https://youtu.be/160jJqJPKdo} (ANYmal), \url{https://youtu.be/Idmq_5hhMic} (Alpha Aerial Scout), and \url{https://youtu.be/iJqPAy0_tGM} (Gagarin Aerial Scout).}
    \label{fig:urban_alpha_planner}
\end{figure}

\begin{figure}[!h]
    \centering
     \includegraphics[width=0.8\textwidth]{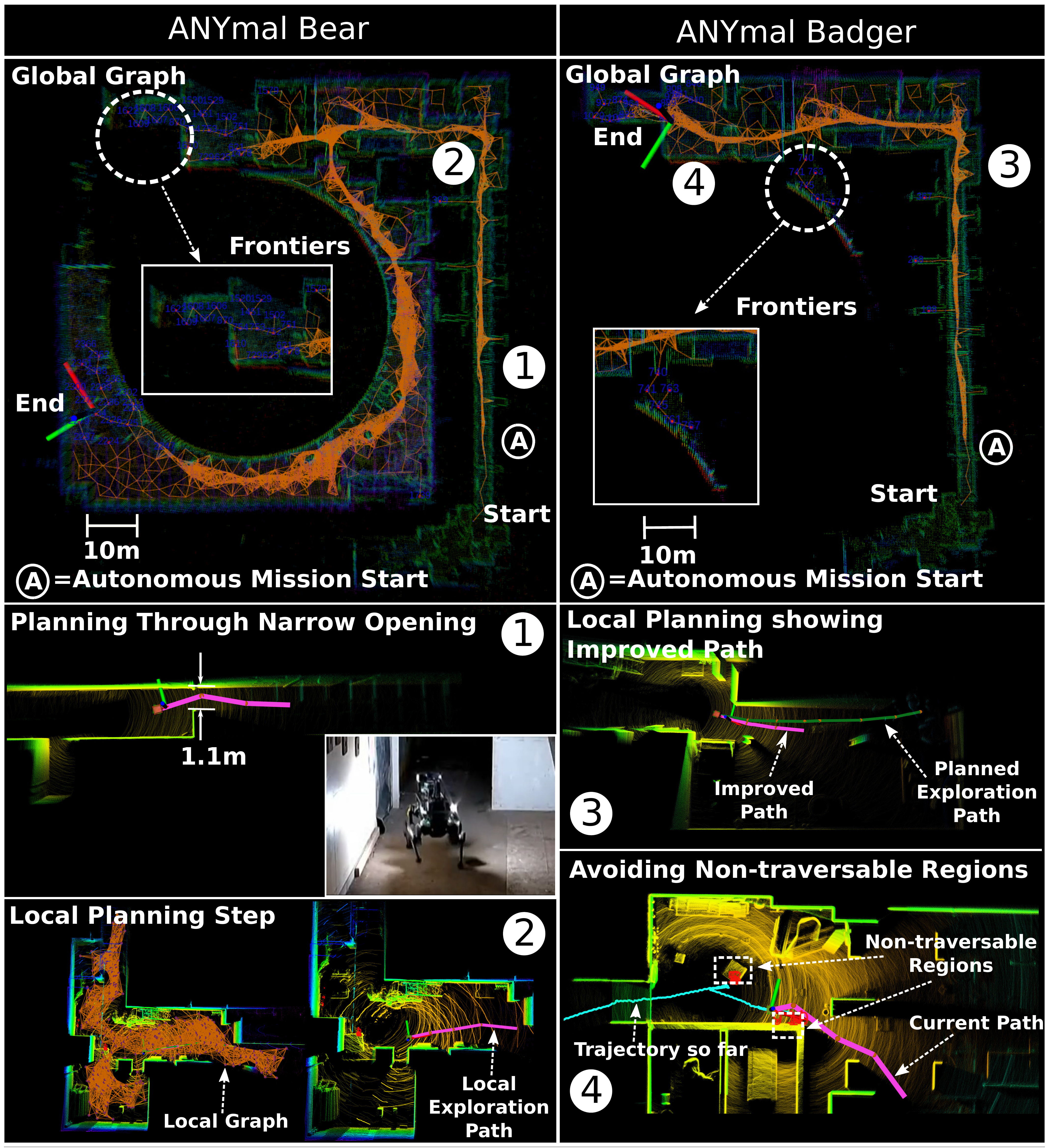}
    \caption{Details of deployment of our exploration planner in the Beta course of the Urban Circuit. Both the ANYmal B robots were deployed running the exploration planner. The figures on the top show the area explored along with the global graph \textcolor{revision}{and its frontier vertices} built during the exploration mission. In the bottom half, various local exploration planning instances are shown. The planner's performance in the presence of narrow passages, \textcolor{revision}{non-traversable} areas and an intersection are shown. The center-right sub-figure shows the planned path and path refined for safety. A video of the mission can be found at~\url{https://youtu.be/0-GbFrrWcL4}.}
    \label{fig:urban_beta_planner}
\end{figure}

To provide detail with respect to the results of the path planning autonomy of our systems, Figures~\ref{fig:urban_alpha_planner} and~\ref{fig:urban_beta_planner} present the graph-based exploration planner paths in the Alpha (second scored run) and Beta (second scored run) courses, respectively. For the Alpha Course, the paths for the Alpha \textcolor{revision}{Aerial Scout} and ANYmal robot exploring the first level are presented, alongside the result for the Gagarin collision-tolerant aerial robot traveling down the stairwell to the next level and landing afterwards. \textcolor{revision}{The frontier vertices from the global graph of ANYmal are highlighted in the leftmost sub-figure.} For the Beta Course, Figure~\ref{fig:urban_beta_planner} presents the results of the planner for the two ANYmal robots exploring the first floor of the course. The results show autonomous navigation through narrow openings, path refinement to enhance the clearance from obstacles, and instances where the traversability constraints applicable to ANYmal have led to adjustment of the exploration path. The figure further presents the dense local graphs built at different instances of the robot mission, alongside the incrementally built sparse global graph, and the associated detected frontiers of the exploration space for candidate robot re-positioning. \textcolor{revision}{In both runs, the aerial and legged robots operated in the horizontal local exploration mode with the local map $\mathbb{M}_L$ set as a cuboid of dimensions $20\times20\times3\textrm{m} ~\textrm{and}~ 30\times30\times3\textrm{m}$ respectively. The Gagarin collision-tolerant aerial robot also flew autonomously through a stairwell and for this task it was operating in the vertical exploration mode with $\mathbb{M}_L$ set to $8\times8\times5.5\textrm{m}$.}

\begin{figure}[!h]
    \centering
    \includegraphics[width=0.32\textwidth]{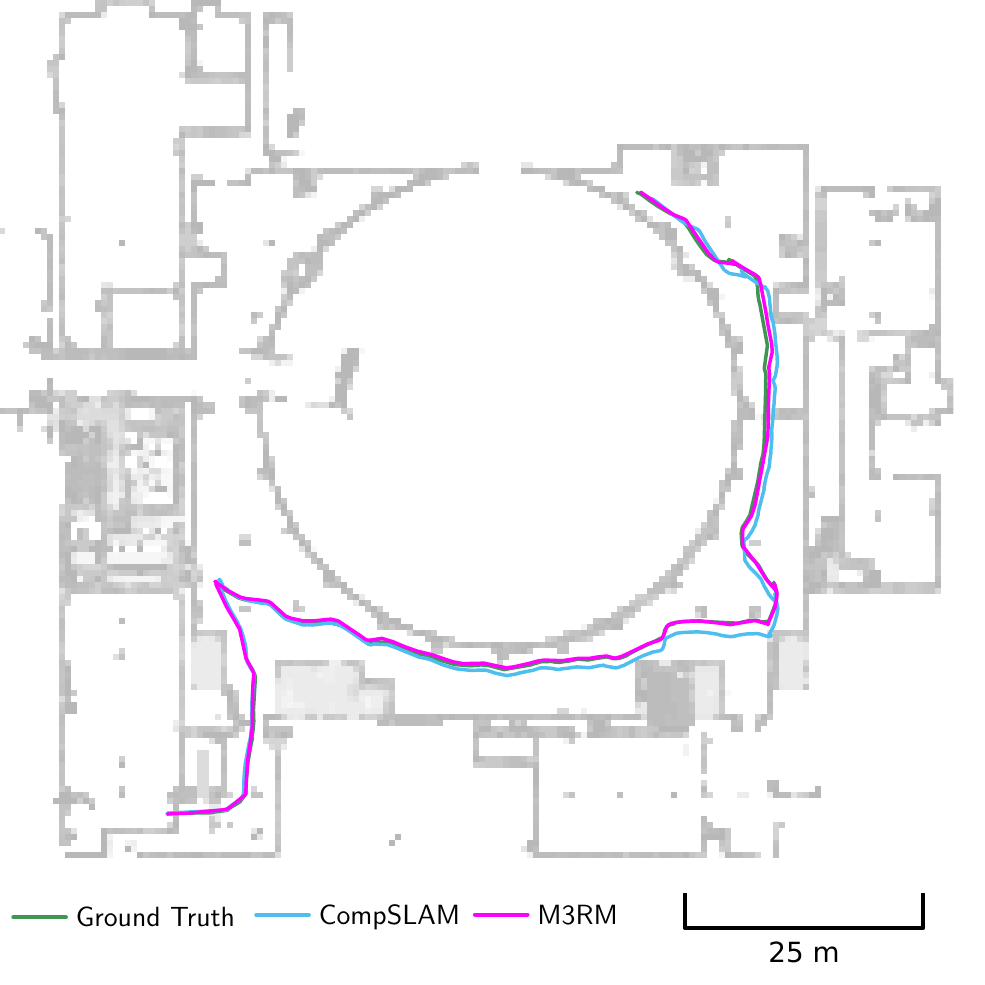}
    \includegraphics[width=0.32\textwidth]{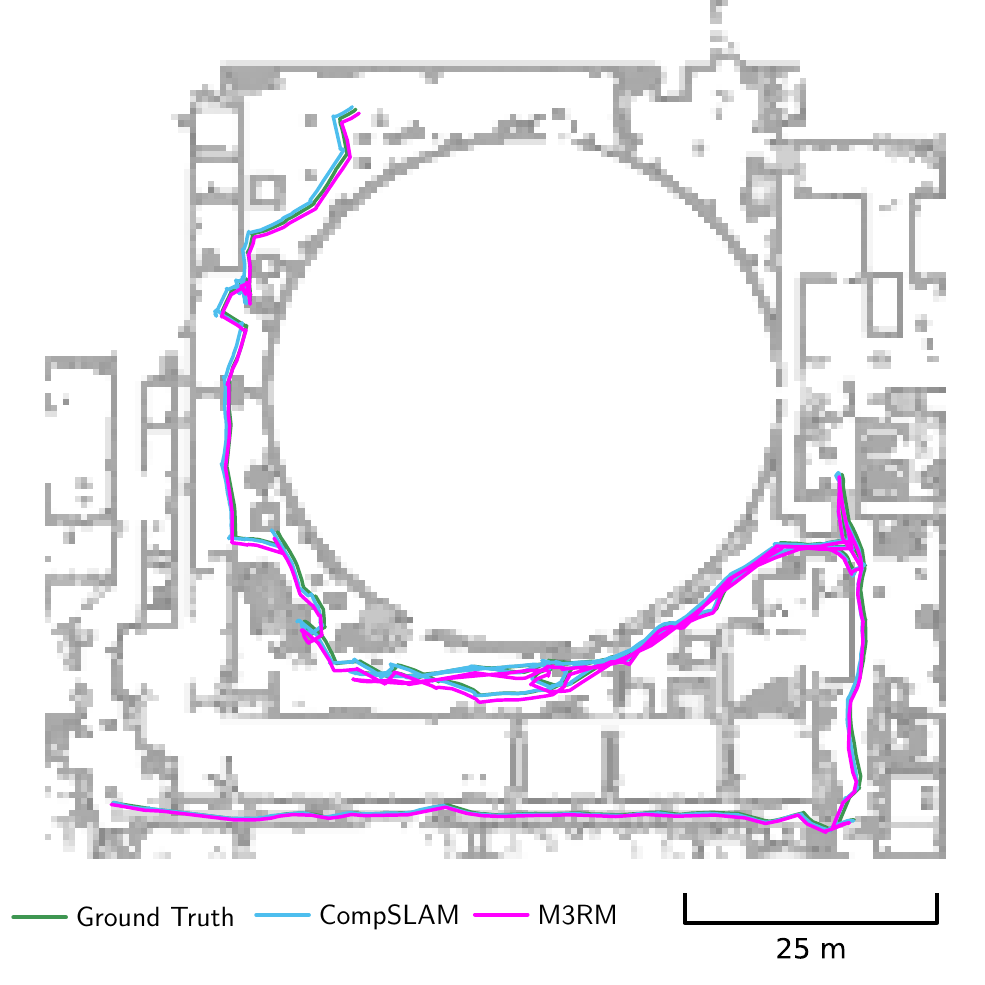}
    \includegraphics[width=0.32\textwidth]{%
    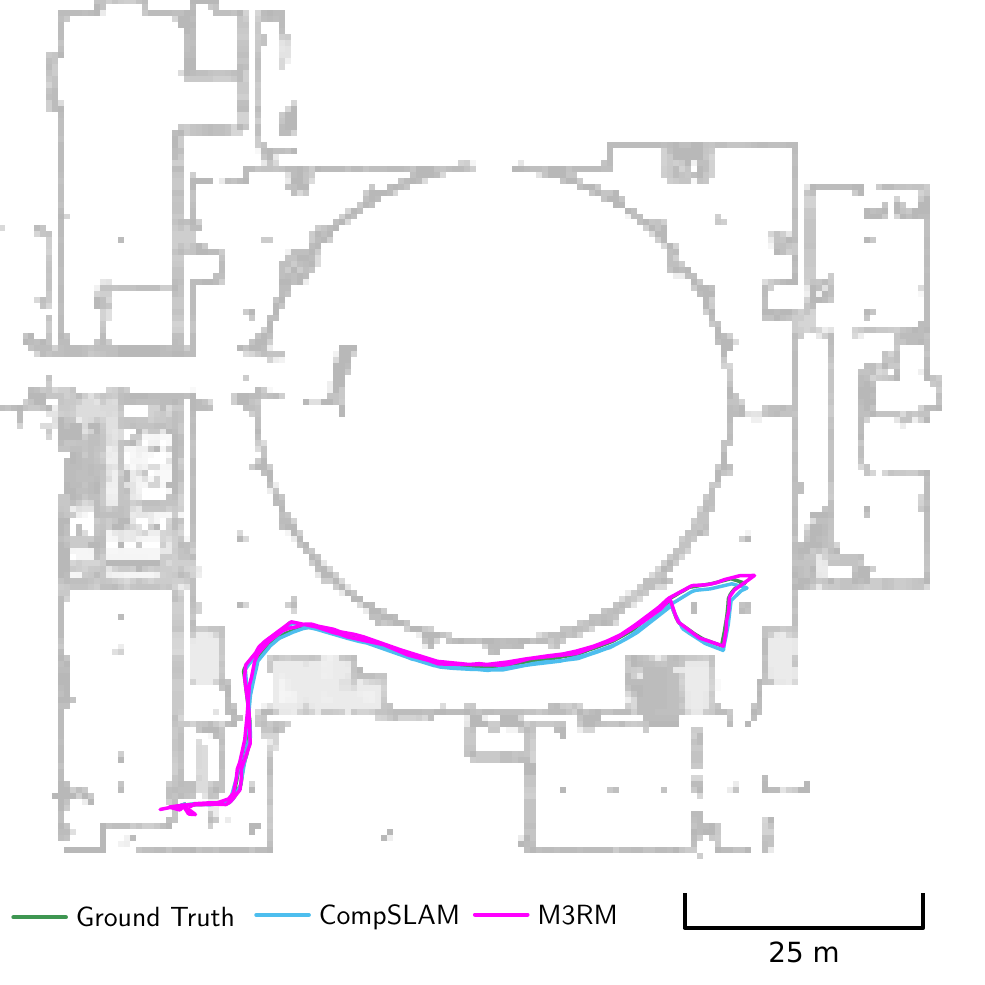}
    \caption{CompSLAM (blue) and M3RM (magenta) trajectories executed onboard the CERBERUS robots against ground truth (green). \emph{Left:} ANYmal Badger, Urban Alpha 2 mission. \emph{Middle:} ANYmal Bear, Urban Beta 2 mission. \emph{Right:} Alpha Aerial Scout, Urban Alpha 2 mission.}
    \label{fig:bear_beta2}
\end{figure}

\begin{figure}[!h]
    \centering
     \includegraphics[width=1.0\textwidth]{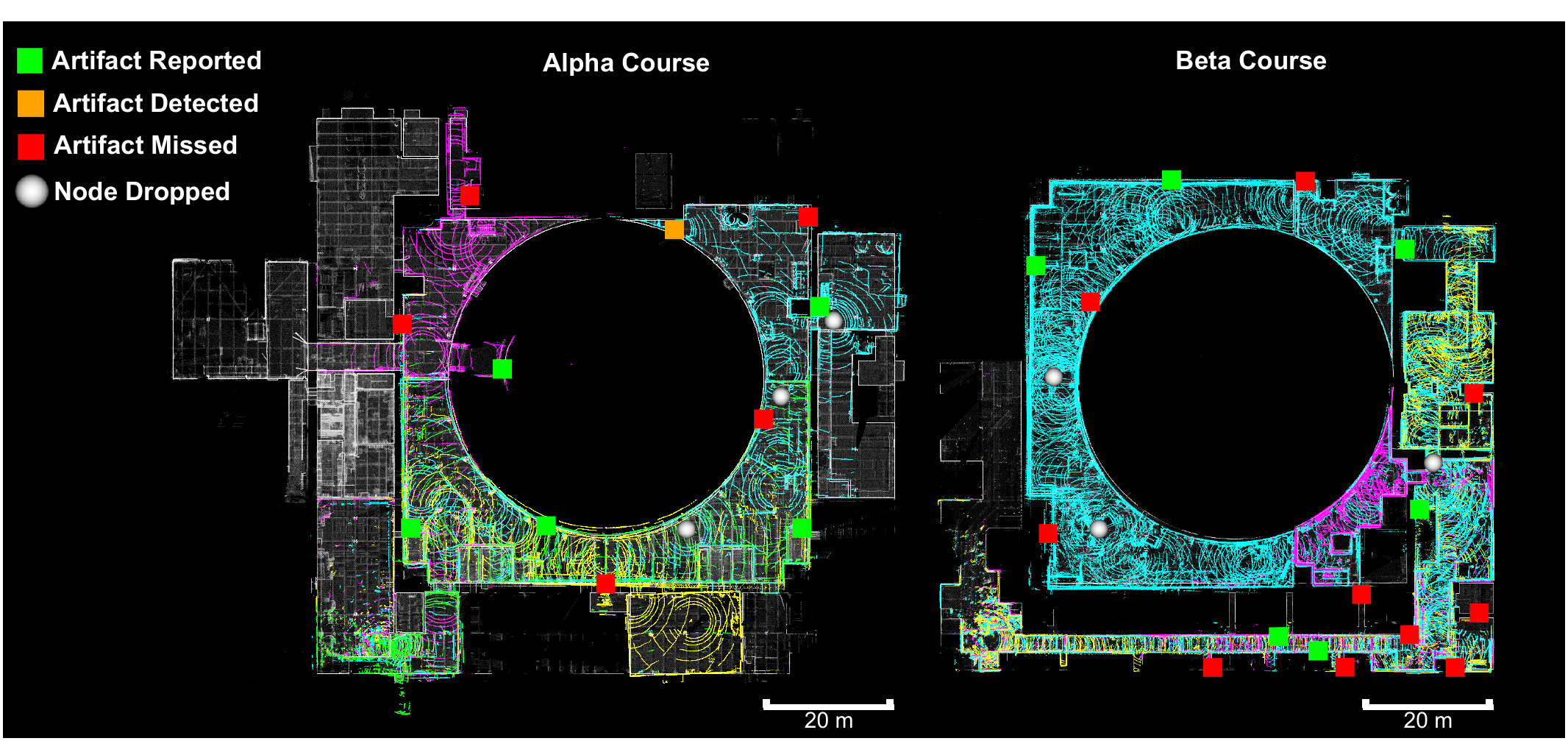}
    \caption{Global multi-robot map of all four sessions of the Urban Alpha and Beta Course using the M3RM approach. The global maps consist of the individual robot maps from ANYmal Bear (cyan), ANYmal Badger (yellow), Armadillo (magenta), and Alpha Aerial Scout (green) if available. Gagarin's map navigating to the next level is not merged. The ground truth map is denoted in white. The color squares denote successfully reported, detected, and missed artifact locations. The spheres illustrate WiFi beacons dropped by ANYmal Bear.}
    \label{fig:urban_maplab}
\end{figure}

To provide a quantitative evaluation of the onboard localization and global mapping approaches, \textcolor{revision}{ground-truth robot trajectories were created by registering each robot pointcloud to the DARPA provided ground-truth map using the ICP algorithm, following a similar approach as described in Section~\ref{sec:tunnel_results}}. Table~\ref{tab:rpe_urban} provides RPE and APE metrics for the CompSLAM (onboard) localization and the M3RM (global) mapping approaches by comparing against individual ground-truth trajectories of walking and flying robots. During the Urban Circuit deployments, all robots utilized visual, depth, and inertial data for their localization and mapping estimation. It can be noted that both onboard and global approaches report low RPE values, and the measured APE values are within the allowed error margin of $\pm$~\SI{5}{\meter} for artifact reporting. These results also demonstrate the benefit of employing a global mapping approach in addition to the onboard estimation as it notably produces lower APE values, which are of particular importance for accurate artifact reporting and scoring. A qualitative comparison of the robot trajectories with respect to ground-truth is shown in Figure~\ref{fig:bear_beta2}. Furthermore, to demonstrate global mapping consistency, point clouds from all deployed aerial, wheeled, and legged robots during all scored runs were combined into globally optimized maps of Alpha and Beta Courses. On the left in Figure~\ref{fig:urban_maplab}, a globally optimized map created by combining the individual maps of \textcolor{revision}{the} Alpha Aerial Scout, Armadillo rover, and ANYmal Bear and ANYmal Badger legged robots is shown and overlaid on top of a DARPA-provided ground-truth map of the Alpha Course for comparison. A similar comparison is provided for the Beta Course on the right in Figure~\ref{fig:urban_maplab}, with a globally optimized map created by combining individual maps of the legged robots Bear and Badger. The presented maps are further annotated with locations of reported, detected (but not reported), and missed (unobserved despite in the area covered by the robot) artifacts as well as the locations of the dropped communication nodes to provide a holistic overview of each mission.

\begin{table}[!h]
\centering
\begin{tabular}{c|c|c|c|c}  
\toprule
\multicolumn{5}{c}{\textbf{DARPA Urban Circuit - Robot Localization Results}} \\
\midrule
\multicolumn{1}{c|}{\textbf{Robot / Run}} & \multicolumn{2}{c|}{\textbf{Relative Pose Error}} & \multicolumn{2}{c}{\textbf{Absolute Pose Error}}\\
\midrule
\textbf{} & \textbf{CompSLAM} & \textbf{M3RM} & \textbf{CompSLAM} & \textbf{M3RM} \\
\midrule
ANYmal Badger / Alpha 2 &  0.19\si{\metre} / 0.94\si{\degree}  & 0.11\si{\metre} / 0.82\si{\degree}
 &  1.07\si{\metre} / 2.05\si{\degree}  & 0.26\si{\metre} / 0.89\si{\degree} \\
ANYmal Bear / Beta 2 &  0.22\si{\metre} / 0.99\si{\degree}  & 0.12\si{\metre} / 0.94\si{\degree}
 &  0.58\si{\metre} / 0.86\si{\degree}  & 0.49\si{\metre} / 0.83\si{\degree} \\
Alpha Aerial Scout / Alpha 2 &  0.07\si{\metre} / 0.66\si{\degree}  & 0.13 \si{\metre} / 1.37 \si{\degree}
 &  0.30\si{\metre} / 0.77\si{\degree}  & 0.25\si{\metre} / 1.23\si{\degree} \\
\bottomrule
\end{tabular}
\caption{Comparison of the mean Relative Pose Error and Absolute Pose Error from the DARPA Urban Circuit between for the onboard (CompSLAM) and global (M3RM) localization and mapping approaches.}
\label{tab:rpe_urban}
\end{table}

\begin{figure}[!h]
    \centering
     \includegraphics[width=\textwidth]{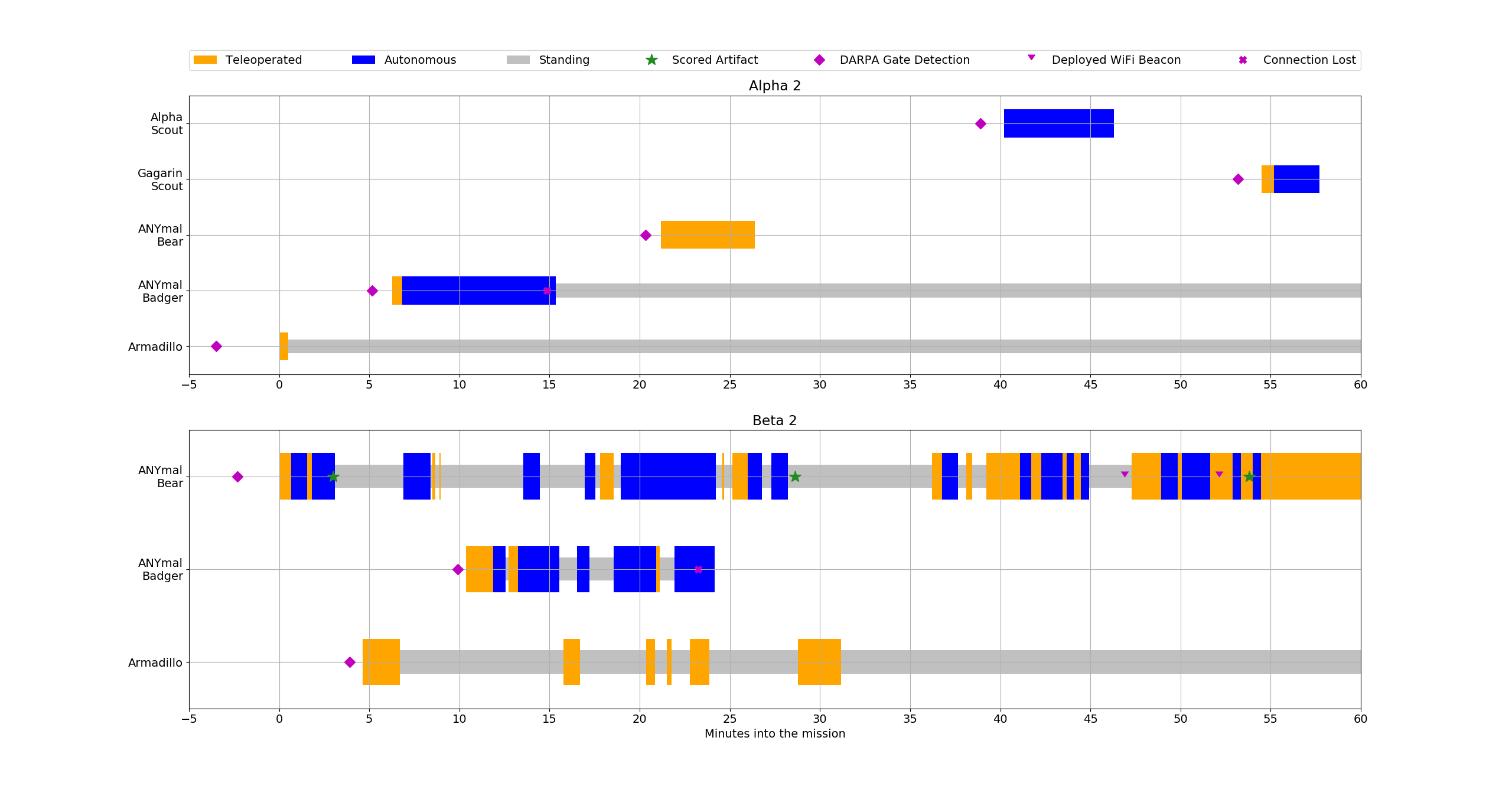}
    \caption{Mission analysis of Alpha $2$ and Beta $2$ runs. The plots show in which mode the robots were operating and when major mission events occurred. During the Alpha $2$ run three robots were deployed in autonomous mode: \textcolor{revision}{The} ANYmal Badger and \textcolor{revision}{the} Alpha Aerial Scout could explore a significant part of the upper floor of the competition area, whereas the Gagarin Scout could plan its path through a series of staircases. ANYmal Bear was teleoperated downstairs a first set of staircases but could not proceed afterwards. At the Beta $2$ run Aerial Scouts were not deployed due to narrow openings at the beginning of the course. \textcolor{revision}{The} Armadillo rover was used to support the autonomous exploration of ANYmal Bear, while ANYmal Badger was sent to explore another area but got stuck on an obstacle and could not recover anymore. The last part of ANYmal Bear's mission was supervised by the Human Supervisor due to an issue in its onboard state machine. ANYmal Bear scored three points by detecting \textcolor{revision}{and} reporting artifacts to the \textcolor{revision}{DARPA Command Post}. A video of the Beta $2$ mission can be found at~\url{https://youtu.be/0-GbFrrWcL4}.}
    \label{fig:mision_analysis}
\end{figure}

An overview of two scored runs (Alpha and Beta second run) is depicted in Figure~\ref{fig:mision_analysis}. In the Alpha $2$ run, the Armadillo rover was driven by the Human Supervisor into the mission area to provide connectivity to the other deployed robots. The rover had to stop after a short period of time because its fiber optical cable got tangled in one of the back wheels. Then ANYmal Badger was deployed and operated in fully autonomous mode. The robot explored the main area of the upper floor of the Alpha Course, but due to a bug in its onboard state machine it neither came back after it lost WiFi connection nor kept on exploring a different area once it reached a dead-end. Afterwards ANYmal Bear was deployed and teleoperated to descend a first ramp of stairs but could not proceed further. At this point, the Alpha Aerial Scout was sent in for another completely autonomous exploration of the upper floor, after which it came home successfully and was retrieved by the Pit Crew. Finally, the Gagarin Aerial Scout could plan its path through a series of staircases completely autonomously. In the Beta $2$ run ANYmal Bear was deployed first, followed at a certain distance by the Armadillo rover. Once the first legged robot arrived at a point far enough from the starting area then ANYmal Badger was deployed and started its autonomous mission. The robot could explore a significant part of the upper floor of the Beta Course until it got stuck on an obstacle and could not recover properly. Moreover, since in that spot there was no WiFi connectivity, the Human Supervisor could not intervene at all. The Aerial Scouts were not deployed in this run due to the presence of two consecutive narrow doors at the entrance of the course, that rendered their deployment less safe, while significant area forward was already explored by the ANYmal robots. After the first half of the mission the Armadillo rover was parked and ANYmal Bear continued the exploration mission. During this last part, the robot successfully deployed two WiFi beacons on the ground to extend the reach of the wireless network. It is noted that the further the robot explored the more the supervisor intervened in steering the robot (roughly after \SI{38}{\minute} inside the mission). This was due to a bug in the onboard state machine that was triggering the homing command too often, even if the robot was commanded to keep on exploring. Therefore the Human Supervisor decided to stop the autonomous exploration (\SI{54}{\minute} inside the mission) and proceed solely with manual way-points. Three artifacts were correctly detected and reported to the DARPA Command Post.

\subsection{Circuits Evaluation}

The main technology evaluations are highlighted for the Tunnel and Urban Circuits. Team CERBERUS successfully deployed a set of diverse robotic systems and finished sixth (out of eleven teams) in the Tunnel Circuit and fifth (out of ten teams) at the Urban Circuit. At the same time, a set of limitations and challenges were faced due to the complexity of the underground environments and the challenges of the competition. Importantly, we describe the relative progress of the CERBERUS system at the Urban Circuit as compared to the Tunnel. This section is followed by a summary of lessons learned and prioritized future activities which reflect how the overall experience is digested and how the plan forward is organized towards long-term resilient autonomy in subterranean environments.

\subsubsection{Tunnel Circuit}
\textbf{Legged robots:} At this Circuit the legged robots faced a set of challenges primarily relating to locomotion issues. ANYmal experienced a corrupted elevation map leading to incorrect locomotion controller actions that prevented the robot from continuing the mission. The issues in the elevation map were caused either by foot slippage or introduced by corrupted data produced by the RealSense depth sensor (for example due to the presence of puddles of water). These challenges were corrected prior to the Urban Circuit especially through the integration of a different sensor and deployment of a new locomotion neural-network based controller, to decouple terrain perception and locomotion modules. ANYmal on wheels on the other hand, executed a fixed trotting gait which did not result in optimal motions and was prone to slippage between the wheels and the mud, which becomes even more critical at a higher speed. The lessons learned at this circuit guided our work for the Urban Circuit and further helped the improvements of the next generation of ANYmal on wheels~\cite{bjelonic2021wholebody}.

\textbf{Aerial Scouts:} The Alpha and Gagarin Aerial Scouts were the flying robots of our team that demonstrated the potential for fast exploration of underground tunnel environments. Especially the Alpha robot presented full autonomous exploration capacity - with no human inputs after initialization - and accurate localization and mapping. At the same time, the Gagarin robot showcased the importance of collision-tolerance when it accidentally hit the mine ceiling but was unaffected. Human mistakes in the definition of the bounding box for the exploration mission of the Alpha flying robot did not allow the robot to go to areas that were not explored previously by ground robots. %

\textbf{Armadillo rover:} Armadillo was developed for the Tunnel Circuit to support the other robots by extending the WiFi range with its directional antenna and fiber optic cable connection to the Base Station. Given the results of the first three runs - where only one point was scored - the team decided to adapt CERBERUS' CompSLAM (a subset of its functionalities) specifically for the rover and made sure it could report artifacts with the onboard cameras. These modifications, which took place the night before the last run, made the robot tele-operable, capable of LiDAR-based localization and mapping alongside vision-based artifact detection\textcolor{revision}{/detection} and resulted in four additional scored points. As such, the Armadillo rover played a larger role in the Tunnel Circuit than originally intended. This also led to the team's decision to use it as a - fallback - perception platform in the subsequent phases of the competition. 

\subsubsection{Urban Circuit}

\textbf{Legged robots:} In the Urban Circuit our legged robots were reliably deployed in all four stages and scored the majority of the points. With the robots now presenting reliable locomotion capabilities, the robots' speed was increased from $0.2\,\textrm{m/s}$ to $0.45\,\textrm{m/s}$. The main challenges faced during this Circuit related primarily to their autonomy behavior and especially to the decoupling of the terrain-aware traversability analysis from the exploration planner. Even though the legged systems could avoid walking into obstacles, the planned paths did not take into considerations any traversability analysis for these systems. 
Without a local traversability-aware planner, the quadrupedal robots were sometimes commanded to cross \textcolor{revision}{non-traversable} regions and therefore had to stop, backup, and trigger the exploration planner again \textcolor{revision}{(with the non-traversable region now marked as geofence)}. This behavior slowed down the overall mission execution and sometimes also required the intervention of the Human Supervisor. Moreover, some corner cases, such as the robot becoming stuck on an obstacle or the concurrent triggering of safety events, were not handled correctly by the onboard state machine and therefore the robot could not recover properly from these events. These aspects are precious findings and are taken into account for the development of the new state machine for the \textcolor{revision}{Final Event}. 

\textbf{Aerial Scouts:} The team deployed two types of flying robots (Alpha Aerial Scout, Gagarin Aerial Scout) fully autonomously. The Alpha Aerial Scout performed autonomous local exploration and safely returned to the home location without a crash or need for human intervention demonstrating the reliable autonomy, path planning efficiency and accurate localization and mapping. \textcolor{revision}{It conducted its mission with an average speed of $0.75\,\textrm{m/s}$.} The Gagarin Aerial Scout demonstrated the vertical exploration capability of the planner by going down the staircase \textcolor{revision}{- presenting an average vertical velocity of $0.24\,\textrm{m/s}$ -} with the only human input being the initial selection of vertical exploration mode (the system exits this mode automatically once a floor is traversed). The Gagarin robot even collided with a metal bar of the staircase it traversed but due to its collision-tolerant design it survived and seamlessly continued its path. However, three problems persist. First, the aerial robots did not score points primarily because they started from the DARPA entrance gate and thus mainly went through areas already covered by ground robots. Second, the robots operate fully autonomously but within some general bounds provided at take-off time. If these bounds are set wrong - primarily due to the not very intuitive interface - the mission may not be productive. Third, we did not use flying robots extensively in this competition event and we deployed them after ground robots as opposed to an early scouting operation. Resolutions for these limitations are discussed in the lessons learned and future work plans.

\textbf{Evaluation of Path Planning Autonomy:} 
The exploration planner was able to provide safe and efficient local exploration paths in real-time. Being designed to work with both flying and legged robots, the planner worked as-is for all the deployed autonomous robots, with the only changes being robot-specific parameters. An observed drawback was the lack of shared information between robots, which resulted in two robots exploring the same area rendering the second robot's mission inefficient. Another observed issue was the lack of the possibility to bias the planner to change the exploration direction. If the Human Supervisor wanted to relocate the robot to a different area, a sequence of way-points towards it had to be created, forcing the supervisor to devote full attention to a single robot. %

\textbf{Evaluation of the Multi-modal and Multi-Robot Perception System:} 
The onboard complementary multi-modal localization and mapping approach deployed on the aerial and wheeled robots during the Tunnel Circuit enabled reliable and robust robot navigation. The approach was extended and unified among all robots for the Urban Circuit and demonstrated its efficacy by producing low relative pose errors, as well as absolute pose errors within the margin of error for artifact reporting. In addition, the multi-robot global mapping solution which was still under development during the Tunnel Circuit was deployed during the Urban Circuit. It improved upon the estimates of the onboard localization and mapping solution and considerably improved the absolute pose errors for accurate artifact reporting.
While exploring the Urban Circuit, the robots were sending submaps to the mapping server at the Base Station every \SI{30}{\second}.
\revisedtext{During the Urban Circuit, each submap was approximately $4$~to~\SI{7}{\MB} large.
Although exact bandwidth usages were not recorded during the competition, internal bandwidth experiments indicated a constant use of \SI{2}{\MBps} for teleoperation (e.g. camera streams) and fallback information (cf.~Section~\ref{subsec:fallback}).
Together with the submap transmission, the used bandwidth can reach $6$~to~\SI{9}{\MBps}.} 
This bandwidth demand is considered to be low enough and compatible with the capabilities of the WiFi breadcrumbs, with the caveat being the ability of the robots to deploy enough WiFi nodes across the competition area. 

Nevertheless, the onboard localization and global mapping solutions' performance was limited by a lack of standardization across the robots in terms of sensing setups. During the competition, the non-availability of sensors, such as thermal cameras, on some robots limited their suitability to operate in challenging obscurant filled sections of the environment. This did not correspond to a practical problem for our team during the deployment, but observing the complete courses after the competition it is important to report that it might become critical in the future. Similarly, robots equipped with similar sensing modalities such as visual cameras; differed in sensor types and lens setups, producing images of varying quality in poorly illuminated environments and limiting the potential for loop closure detection for multi-robot mapping. Furthermore, combining all the robots' information allowed for accurate mapping and improved pose estimation but increased the size of the underlying optimization problem, thus requiring longer processing times. During the processing of globally optimized maps, the Human Supervisor could not view a sufficiently recent global multi-robot map and relied on onboard maps of individual robots for mission assessment.

\textbf{Evaluation of the Deployed Communications Solution:}  The combination of a tethered rover and deployable WiFi modules considerably helped the team's overall performance during the mission execution by extending the wireless network's reach. The main faced challenges were due to the quality of our communication breadcrumbs, the difficulties in driving the tethered rover in an environment featuring several turns, the decision of how many WiFi modules to deploy from one ANYmal versus moving the rover further, and how many breadcrumbs ANYmal could integrate in relation to the effective range provided by each of them. %

\emph{Optical fiber solution} - The usage of an optical fiber cable offered a considerable advantage, especially in urban scenarios where thick walls and multiple turns played an important role in reducing the WiFi signal strength. On the other hand, while teleoperating the Armadillo rover, the Human Supervisor was completely focused on this task, while also making sure that the tether itself would not tangle or become tangled in the back wheels. 

\emph{WiFi Range and Quality} - Our team faced a set of challenges with respect to the WiFi modules utilized. Efforts for in-house development were not successful which led to the use of an off-the-shelf solution for which we had limited time to adjust. Observing the quality reached by other competing teams we identified this as a key limitation that we are currently prioritizing for our current field experiments and the next DARPA Circuit deployment. 

\emph{WiFi Breadcrumbing} -  The decision of where to deploy a WiFi breadcrumb is not a trivial task. Both signal strength and geometric characteristics of the surrounding environments had to be taken into account. Two commonly occurring problems were detected: wrong decisions on where to deploy, and tilted modules after deployment. In the former case, it could happen that a deployed node was released too far from the previous breadcrumb and therefore could only connect to the mesh network through the robot which deployed it. It would then lose connection as soon as the agent proceeded further. For the latter case, the modules could tilt or flip up after the deployment (for example due to releases on slopes or uneven terrains in general), causing the internal antennas to face against the ground and therefore considerably reducing the wireless range of the node.

%% file: 13_lessons_learnt_and_discussion.tex
Every field testing experience and especially every circuit event of the DARPA Subterranean Challenge provides large amount of information and insight\textcolor{revision}{s} with respect to how our research and development efforts towards efficient subterranean exploration should progress. CERBERUS represents an approach to the problem of robotic subterranean exploration that aims for one unifying solution across all types of underground environments that, once developed, can be scalable and versatile. This is the key motivation behind the robot platforms utilized, primarily legged and (collision-tolerant) flying platforms, and the methods designed. We believe that such a system-of-systems solution offers the resourcefulness required to navigate any subterranean setting. However, this approach also comes with a certain development overhead. Most of our deployed platforms are primarily research prototypes with significant technical challenges that had to be addressed during the circuit event preparations. At the same time, our focus on autonomous operation also implied that we initially had to dedicate a disproportionate amount of effort towards reaching a behavior that would have been rather easy to achieve with a more trained teleoperation process. However, observing the technological progress and the way our systems performed at every new field testing activity, we are confident that this solution is one of the most appropriate and we are thus particularly optimistic for new field tests, the subsequent circuit events, and the overall future of the technology. This section, provides overall conclusions for lessons learned that we hope can inform other teams, research groups, technology developers and system operators that share similar goals and interests.

\textbf{Multi-Modal Perception, Localization and Mapping:} The underground environments present a variety of challenges for robot perception. Proper utilization of multi-modal sensor data is critical for enabling robust and resilient, single- and multi-robot localization and mapping underground.

\emph{Unifying Onboard Sensing Solution} -  During the competition deployments, robots varied in their onboard sensing payloads. Such heterogeneous sensor setups can impose significant challenges for deploying the same multi-modal estimation and multi-robot mapping approaches between robots. In addition to utilizing different sensors, small changes such as different camera types, lenses, and placements of illumination sources on the robot, play a critical role in the overall system performance, especially for loop closure and map merging.
In the future, we aim to unify the employed sensor setup among robots. First, given robot payload capacity, we aim to equip the robots with the same set of multi-modal sensing capabilities to allow resilient operation of each robot in sensory degraded environments. Second, we aim to equip them with analogous sensors, lenses, synchronization hardware, and illumination. Such unification would help to guarantee an expected baseline performance from each robot and simplify the deployment of onboard multi-modal localization across the robots. Furthermore, it would also facilitate building globally consistent multi-robot maps.

\emph{Localization in Subterranean Environments} - Localization in underground environments can be particularly difficult due to their complex nature and perceptually-degraded operating conditions. To improve the onboard localization performance, the CompSLAM approach will be extended to utilize external odometry estimates that are unique to robots, such as legged and wheeled odometry. In addition, to improve each estimation module's performance, data augmentation across sensing modalities, such as LiDAR depth fusion for visual and thermal feature tracking, would be incorporated. This can contribute both in accuracy but also the resilience of the system in obscurant-filled settings. Furthermore, to incorporate data from multiple sensors of the same type (e.g., multiple visual camera images), data level quality evaluation~\cite{khattak2019visual} will be performed to make use of all available sensor data while keeping computational costs low. Finally, to facilitate tight constraint-based fusion across sensing modalities and external inputs, we aim to exploit factor-graph-based fusion~\cite{Wisth2020vilens} to provide robustness and resilience in dark, dusty, and degenerate underground environments.

\emph{Consistent Onboard and Multi-Robot Perception} -  The integration between the CompSLAM and the M3RM client module running on the robot was still limited during the Urban Circuit. In future missions, the computed onboard features will also be shared for tighter integration with the global map building process and to avoid redundant computations.

\emph{Map Sharing} -  Bi-directional map sharing between robots and the Base Station, when in communication range, will provide individual robots with insight into unexplored areas of the environment. Such consistent map sharing between individual robots and the Base Station will improve each robot's global planner by explicitly knowing other robots' explored areas and allowing for collaborative planning. Furthermore, maps will be shared between robots to facilitate the launching of aerial robots from ground robots deeper into the circuit, allowing them to report artifacts without requiring an independent initial DARPA frame alignment.

\emph{Artifacts} -  Accurate artifact detection and reporting turned out to be a challenging problem. Our robots experienced cases where they passed by artifacts without detecting them. One reason was the limited and inconsistent illumination provided by each robot. Another reason related to limitations in our implementation. Although we initially attempted to determine the cell phone artifact positions by trilaterating the received signal strength of their Bluetooth signal, we eventually found that just notifying the Human Supervisor of Bluetooth signal presence and letting them determine the cell phone position yielded better results. A second requirement is that of accurate location reporting. The results given by the onboard localization and mapping solutions were satisfactory, but the centralized M3RM approach with map optimization provides notable improvements. %
M3RM decreases the individual robot APE values, resulting in more precise map and artifact position estimates even in the presence of drifting onboard estimation, whose errors grow in proportion to the distance travelled in the deployed implementation.
Considering that artifact reports are only scored when the reported and true artifact positions differ by less than $5\si{\meter}$, these improved estimates become indispensable as the exploration range of the robots and the scale of the environments increases.

\textbf{Exploration and System-level Autonomy} -  Due to the scale and nature of subterranean environments and the number of robots controlled by one Human Supervisor, the deployed machines have to be fully autonomous with respect to their system functionalities and as autonomous as possible concerning their exploration and information gathering behavior.

\emph{Exploration Path Planning and Artifact Search} -  GBPlanner has proven to be an efficient planner for volumetric exploration but presents two essential limitations with respect to the needs and challenges of the SubT Challenge. First, this relates to the fact that the method only accounts for the frustum of a depth sensor and derives paths based on the maximization of an associated information gain reflecting how much new volume will be explored. However, mapping new areas with a depth sensor such as a LiDAR does not automatically imply that the artifacts in this area will be detected. Artifact detection primarily relies on multiple sensors, such as cameras with different frustums or $\text{CO}_2$ and Bluetooth detection. In fact, a multi-hypothesis/multi-objective exploration and artifact search path planning pipeline is required. The second key limitation of our exploration planner is that for legged robots in particular it only has access to a short-range traversability analysis map which in turn does not allow it to securely plan long-term exploration paths. This can be addressed in two main ways, extending the range of the traversability map or, at a minimum, re-formulate the planner to run in a receding horizon fashion, thus maximizing robustness against wrong assumptions for the future terrain traversability. The latter is a direction we already consider, while the former corresponds to a more challenging problem in robotic perception. Last but not least, as multi-robot map sharing is facilitated then the extension of our methods to autonomous multi-robot exploration will take place. 

\emph{State Machine} -  To maintain resilient autonomy and thus detect more artifacts, it is of major importance that each robot returns into communication range reliably before the end of the mission. To this end, the onboard state machine handles all the different cases. Until the Urban Circuit event, the team did not test long missions often enough which resulted in corner cases that were not handled properly during the circuit events. One of the major issues faced during the Urban Circuit by the legged robots was the inability to handle unforeseen events, such as getting stuck on an obstacle, with the predefined recovery behaviors. Whenever such events occurred, especially in areas with no WiFi connection with the Base Station, the robots could not recover and continue with the mission execution. To increase robustness in terms of autonomy, a behavior-based autonomy pipeline~\cite{colledanchise2016behavior,colledanchise2018behavior} is under consideration, in contrast to a more classical state machine framework. Furthermore, there is a clear need for a ``watchdog'' module with minimal interaction with other parts of the software that can recover the robot in case of partial software failures e.g., the planning pipeline.

\emph{Use of Simulation} -  In our developments up to the Tunnel Circuit, we were only making minimal use of simulation tools to test our path planners extensively. We had not invested in developing the world models that would allow us to test a set of planning corner cases. Instead, we prioritized field testing. On the way to the Urban Circuit, we employed simulation significantly more, and we are currently investing in its utilization even further. Based on our experience, simulation utilization is particularly important as a team progresses to investigate more complex exploration behaviors or multi-robot teaming.

\textbf{Locomotion Controllers:} A set of different controllers were deployed for the three versions of ANYmals (point feet, flat feet, wheels) during the two circuits. During the Tunnel Circuit, a perceptive model-based controller was used for the point and flat feet versions of the robot, whereas a specialized, non perceptive, model-based controller was employed for ANYmal on wheels. At the Urban Circuit, a neural-network-based controller was used for the flat feet robots.

\emph{Point Feet} -  The main driver for the change in the control strategy between the Tunnel and Urban Circuits \textcolor{revision}{lay} in the interactions between the model-based controller and the terrain perception modules. The robustness of our locomotion controller, in the end, was limited by the quality of the perception part, which turned out to be challenging in the diverse subterranean environments, making the model-based controller more prone to failure. Therefore, during the Urban Circuit, the perception part was used only for traversability analysis purposes. Moreover, we observed very resilient locomotion in diverse environments with our blind neural-network-based controller. The steps towards the following runs include the investigation of a perceptive neural-network based controller to exploit the full locomotion capabilities of the legged robots while improving the detected issues in the terrain perception modules.

\emph{Wheels} -  ANYmal on wheels and its locomotion controller were not deployed in underground environments often enough before the Tunnel Circuit, which resulted in locomotion respectively slippage issues over the muddy terrain. During the Tunnel event, the robot executed a fixed trotting gait, which did not result in optimal motions. The experience and lessons learned during this circuit contributed to the development of a new model predictive controller \cite{bjelonic2021wholebody}, where the robot dynamically chooses an optimal hybrid gait that is a fusion of powered rolling and legged stepping. In the Urban Circuit, although ANYmal on wheels could have brought competitive advantages (such as higher navigation speed), the team decided to favor a more standardized and tested approach and therefore employed two legged robots with the same point foot configuration.

\textbf{Communication:} The reliability and performance of underground communications have corresponded to a critical challenge in our \textcolor{revision}{experience so far}. On the contrary, we have a positive experience concerning the networking solution and the particular implementation via the software package ``Nimbro Network''.

\emph{WiFi Breadcrumbs} -  Due to changes in the development process our team has faced difficulty to a) deploy miniaturized communication breadcrumbs with the appropriate balance of size/weight and range such that they can correspond to a solution tailored to our ground robotic systems, and b) reliably test the deployed solution well in advance of the competition. Despite focusing on autonomy, ensuring reliable communications persists as an important technical challenge. 

\emph{Tethered Rover} -  The Armadillo rover was a valuable resource to extend the reach of the WiFi network and also act as a fallback perception platform. However, too much of the Human Supervisor's time was used to remotely control it. This solution should have been employed only for the very first part of the mission in the Urban Circuit, i.e., the rover should have been driven through the first main turns from the starting point and be parked there to support the other robots.

\textbf{Interfacing with Multiple Robots:} Supervising a multi-robot team while operating at the same time in unknown environments requires customized tools whose development was tackled only before the Urban Circuit. The main observed limitations and associated necessities with respect to this are: a) the lack of a unified and easy to use single user interface to control multiple robots; b) a user interface too complex to gather information about the updated global map, camera streams from the robots and status of detected artifacts; c) the absence of the possibility to both re-locate robots as well as bias their exploration direction without manually specifying intermediate way-points to the new zone that should be explored. These points will be addressed for the next runs.

\textbf{Software and Hardware Unification:} CERBERUS incorporates research prototypes of systems implementing novel features, either relating to legged robots or collision-tolerant flying systems. In our team - especially due to our geographic organization - we have identified the importance of unified software among as many of the robot functionalities as possible. Examples include the CompSLAM, M3RM and GBPlanner software packages which at the Urban Circuit (and subsequently) are uniformly deployed across all robots. This was not the case in the Tunnel Circuit and thus, we faced many integration challenges. The same principle holds for sensing hardware which was not fully unified in the Urban Circuit and is currently under improvement. Different hardware leads to different behaviors, the need for additional training data for artifact detection and other challenges. 

\textbf{Team Preparation:}  In the first period of the DARPA Subterranean Challenge, our team faced two important challenges specific to our configuration: a) our technological solutions existed across different groups and laboratories, and \textcolor{revision}{b) we} were established in different geographic locations \textcolor{revision}{(}USA and Switzerland). Integration activities across both sides of the Atlantic were scheduled and took place, but this was not enough for truly detailed system integration and testing. Since then we have achieved and established two major features, namely a) that our software solutions are being unified across different robotic configurations in all aspects that do not relate to robot-specific functionality, and b) we can execute missions that represent CERBERUS robots as a whole in Switzerland where the ANYmal robot is based and is rather more difficult to transfer between continents for frequent field tests. The know-how for the latter largely benefited from the exchange of research personnel among our teams and from the approach to make aerial robots operate fully autonomously after take-off.

\textbf{Research vs. Competition:}
Different tools and best practices were developed during the first two DARPA SubT events, which contributed to increasing the robustness of the overall systems. The main points of which are outlined here.

\emph{Logging} -  Extensive logging is key for robotic system development as issues can often not be detected during field-testing directly due to the complete system's complexity. Additionally, data can be used to analyze performance after field-testing in more detail. 
These findings were implemented between the Tunnel and the Urban Circuits and resulted in logging essential data in chunks (created every \SI{5}{\minute}) to preserve as much data as possible in case of a complete system failure. 

\emph{System Launch} -  To ensure a stable and robust launching of software, all ssh-shell sessions of the onboard PCs were run as \emph{screen} sessions. This allowed to detach and reattach to existing sessions and avoided issues in the event of connection drops. To enhance remote monitoring and usability of the different applications (\ac{ROS} nodes) running on the robots, \emph{rosmon}\footnote{\url{http://wiki.ros.org/rosmon}} was used during the Urban Circuit. The next step before the final event is to switch from \emph{screen} to a different terminal multiplexer with some extra built\textcolor{revision}{-}in functionalities such as \emph{tmux}\footnote{\url{https://github.com/tmux/tmux}} which also offers the possibility of saving the complete output history to file to further improve post processing analysis and debugging of every mission.

\emph{Periodic Mission Shakeouts} -  Weekly mission shakeouts were introduced after the Tunnel Circuit, to help the continuous integration tests of the deployed software. The goal was to test a mission scenario in a smaller scale environment with easy access, such as a garage or an underground hallway, and help the team spot software defects. They were carried out even if no major software changes were introduced and also helped the Pit Crew in standardizing the setup procedure of each robot. Though, these weekly shakeouts did not replace proper field deployments in underground scenarios. These were carried out regularly across the months proceeding the Urban Circuit.

\emph{Relevant and Challenging Field Testing} -  Eventually, a key component relates to real field testing our robotic systems in relevant environments and scenarios. We have learned that this process not only reveals the current status and allows for team training but also greatly helps in guiding the research required to enable resilient autonomy in such challenging robotic missions. For example, our CompSLAM and exploration planning pipelines are largely developed based on observations acquired during such deployments. Thus relevant field testing is of paramount importance both for competition-readiness and informed guidance of the research efforts.

\textbf{Open Sourcing:} Our experience in the SubT Challenge indicates that major progress will have been achieved by its completion, but corner cases will likely still exist. For the research and innovation community to proceed on top of the prior experience our team strongly considers and performs open-sourcing activities. We believe this \textcolor{revision}{is} essential as such packages lead to research reproducibility, ability to compare, continuity, community development and collaborative exploitation. A list of packages open-sourced by the team is presented in Appendix~\ref{sec:appendixA}.

%% file: 14_future_work.tex
CERBERUS has outlined a plan of essential improvements that are currently taking place. These relate primarily to the following:

\emph{Subterranean Legged and Flying robots} - One of our key priorities for the future is the introduction and utilization of the new generation of the ANYbotics ANYmal robot: ANYmal C, which will fully replace the previous ANYmal B version. The new ANYmal C robots are now in use and provide the advantages of a new stronger, larger and more dexterous robot that further implements the latest version of the learning-based locomotion controller and seamlessly integrates the complete multi-modal sensor and processing suite of CERBERUS. Furthermore, another essential priority relates to high-speed aerial robot exploration, primarily via the collision-tolerant flying robots, and the integration of the flying robots onboard the ground systems so that their critical flight time is best used by being deployed where the ground robots cannot reach. These are - according to CERBERUS' vision - prerequisites for successful utilization of flying robots for underground exploration.

\emph{Resilient Multi-modal Robot Perception} - The employed localization and mapping approaches will be improved to integrate available external estimates unique to robots, such as leg odometry for walking robots, to improve the overall estimation quality. Furthermore, to improve the robot operation's resilience and robustness in self-similar and obscurant-filled environments, detection of ill-conditioning and degradation will be performed on both method and data levels, respectively. Finally, tighter integration between multi-modal sensors will be exploited to improve the overall localization and mapping performance. 

\emph{Multi-robot Mapping} - To improve the collaborative localization, mapping, and planning between robots, the onboard and global localization and mapping approaches will be updated to communicate in a bi-directional manner with knowledge of the areas covered. Such development would facilitate the synergistic deployment of the communications breadcrumbs and detection of artifacts. 

\emph{Autonomous Exploration and Visual Search} - Fully autonomous exploration and artifact search capability at the level of every robot by co-optimizing both in terms of volumetric gain and artifacts' coverage will be developed, alongside sampling of paths in a manner that accounts for all the particularities of every platform, while remaining unified, and incorporating the recovery behaviors that facilitate resilient and unobstructed mission execution without need for human intervention other than true high-level decisions. 

\emph{Communications} - New hardware for subterranean communications based on a new baseline WiFi module (Rajant DX2) with improved range and reliability is being developed, alongside optimized breadcrumbing capabilities on the ANYmal C robot. This is combined with the Armadillo rover deploying its high-gain antenna while maintaining optical fiber connection to the ground.

\emph{Frequency and Character of Field Testing} - A key aspect to meaningfully progress towards the SubT Challenge Finals and help in enabling resilient autonomous robotic operations underground relates to the frequency and the character of our field testing. The goal is to conduct large-scale field testing with increased frequency (e.g., monthly) and with a ``Circuits-character'' (i.e., by deploying the robots as per the rules of the challenge).

These priorities with respect to future work are part of a broader plan to enhance the operational capabilities and technological novelty of the CERBERUS system-of-systems as guided by the lessons learned and renewed via the continuous research contributions of our team laboratories. 

%% file: 15_open_source_list.tex
With the goal to support the overall community around subterranean robotics, our team simultaneously focuses on open-sourcing significant components of our technological solution. A selected - but not exhaustive - list of relevant open-source packages of our team are outlined below. 

\emph{GBPlanner} - a graph-based exploration planner for subterranean environments:~\url{https://github.com/ntnu-arl/gbplanner_ros} 

\emph{MBPlanner} - a motion primitives-based agile exploration planner for subterranean environments:~\url{https://github.com/ntnu-arl/mbplanner_ros} 

\emph{Gird Map} - a universal grid map library for mobile robotic mapping:~\url{https://github.com/ANYbotics/grid_map}

\emph{Kalibr} - a visual-inertial calibration toolbox:~\url{https://github.com/ethz-asl/kalibr}

\emph{Lidar Align} - a package to calibrate LiDAR extrinsics:~\url{https://github.com/ethz-asl/lidar_align}

\emph{Maplab} - an open visual-inertial mapping framework, which includes the M3RM centralized multi-modal, multi-robot mapping extension:~\url{https://github.com/ethz-asl/maplab}

\emph{Darknet ROS} - Real-Time Object Detection for ROS:~\url{https://github.com/leggedrobotics/darknet_ros}

\emph{Elevation Mapping} - a robot-centric elevation mapping framework for rough terrain navigation:~\url{https://github.com/ANYbotics/elevation_mapping}

\emph{ROVIO} - a Robust Visual Inertial Odometry framework:~\url{https://github.com/ethz-asl/rovio}

\emph{Traversability Estimation} - a traversability mapping framework for mobile rough terrain navigation:~\url{https://github.com/leggedrobotics/traversability_estimation}

\emph{Voxblox} - a library for flexible voxel-based mapping:~\url{https://github.com/ethz-asl/voxblox}

\emph{Voxgraph} - a globally consistent volumetric mapping framework:~\url{https://github.com/ethz-asl/voxgraph}

\section*{Simulation Models}

The SubT Challenge Systems and Virtual Competitions are interconnected with the Systems' teams delivering models of their robots for the Virtual Competition, while progress achieved in the Virtual Competition can benefit the autonomy functionalities in the Systems Competition. Models of most of the robots fielded by our team in the Tunnel and Urban Circuits have already been released.

\emph{Virtual Robot Models} - CERBERUS robots have been released for use in the the official ``SubT Virtual Track Challenge'' simulation environment. Robots models released so far include the ANYmal B ``SubT'' (\url{https://github.com/osrf/subt/tree/master/submitted_models/cerberus_anymal_b_sensor_config_1}), the Alpha Aerial Scout (\url{https://github.com/osrf/subt/tree/master/submitted_models/cerberus_gagarin_sensor_config_1}), and the Gagarin Aerial Scout (\url{https://github.com/osrf/subt/tree/master/submitted_models/cerberus_m100_sensor_config_1}).

%% file: 16_videos.tex
We have assembled a set of videos presenting highlights of the deployment of the CERBERUS system-of-systems in the Tunnel and Urban Circuit events of the  DARPA Subterranean Challenge. 

\begin{itemize}
    \item CERBERUS deployment at the DARPA SubT Challenge Urban Circuit \textit{(combination from different courses)}: \url{https://youtu.be/pkgcsdAgxVc}
    \item Detailed overview of the CERBERUS deployment at the Beta Course of the DARPA SubT Challenge Urban Circuit: \url{https://youtu.be/0-GbFrrWcL4}
    \item The Alpha Aerial Scout exploring the first level of the Alpha Course of the DARPA SubT Challenge Urban Circuit \textit{(video previously released)}: \url{https://youtu.be/Idmq_5hhMic}
    \item The Gagarin Aerial Scout proceeding to the lower level of the Alpha Course of the DARPA SubT Challenge Urban Circuit \textit{(video previously released)}: \url{https://youtu.be/iJqPAy0_tGM}
    \item The ANYmal B exploring the first level of the Alpha Course of the DARPA SubT Challenge Urban Circuit \textit{(video previously released)}: \url{https://youtu.be/160jJqJPKdo}
    \item The Alpha Aerial Scout in the Safety Research Course of the DARPA SubT Challenge Tunnel Circuit \textit{(video previously released)}: \url{https://youtu.be/mw0qy05Fo6Q}
\end{itemize}